\documentclass[11pt]{article}
\usepackage[preprint]{acl}
\usepackage{xcolor}
\usepackage{colortbl}
\usepackage{booktabs}
\usepackage[linesnumbered,ruled,vlined]{algorithm2e}
\usepackage{algorithmic}
\usepackage{amsmath}
\usepackage{amssymb}
\usepackage{amsthm}
\usepackage{times}
\usepackage{latexsym}
\usepackage{multirow}
\usepackage{siunitx}
\usepackage{makecell}
\usepackage[T1]{fontenc}
\usepackage[utf8]{inputenc}

\usepackage{microtype}
\usepackage{tabularx}
\usepackage{inconsolata}

\usepackage{graphicx}
\usepackage{makecell}
\usepackage{multirow}
\usepackage{graphicx}
\usepackage{subcaption}
\usepackage{svg}
\usepackage[most]{tcolorbox}
\usepackage{enumitem}
\tcbuselibrary{listings}
\theoremstyle{definition} % 1. 设置样式为直体 (而非斜体)
\newtheorem{defn}{Definition} % 2. 定义环境名为 defn，显示名为 Definition，按章节编号
\newtcblisting[auto counter]{promptbox}[2][]{ 
    colback=green!5,
    colframe=green!40!black,
    arc=3mm,
    fonttitle=\bfseries,
    label={#1},                        % 这一行让你能引用它
    title={Prompt \thetcbcounter: #2}, % 这一行让它自动显示 "Prompt X: 标题"
    listing only,    % 保持原始内容模式
    listing options={
        basicstyle=\ttfamily\small,
        breaklines=true,
        breakindent=0pt,
        columns=fullflexible,
        keepspaces=true
    }
}

\title{Quantifying Error Tolerance in Synthetic Data: An Atomic-level \\ Operand vs. Operator Perturbation Study}

\author{
    \textbf{Jiaxiang Liu}\textsuperscript{$*$ 1,2}\textbf{,} \textbf{Chenhao Yuan \thanks{Equal Contribution: liujiaxiang2025@ia.ac.cn}}\textsuperscript{ 1,2}\textbf{,} \textbf{Shuwen Xu}\textsuperscript{$*$2}\textbf{,} \textbf{Boxuan Xing}\textsuperscript{1,2}\textbf{,} \\ \textbf{Xiusheng Huang}\textsuperscript{1}\textbf{,} \textbf{Yinhao Xu}\textsuperscript{2}\textbf{,} \textbf{Hao Liu}\textsuperscript{1,2}\textbf{,} \textbf{Wenhao Teng}\textsuperscript{3}\textbf{,} \textbf{Xiangwen Liao}\textsuperscript{4}\textbf{,} \\ \textbf{Pengfei Cao}\textsuperscript{$\dag$ 1,2}\textbf{,} \textbf{Jun Zhao}\textsuperscript{1,2}\textbf{,} \textbf{Kang Liu \thanks{Corresponding author: \{pengfei.cao, jzhao, kliu\}@nlpr. ia.ac.cn}}\textsuperscript{1,2}
    \\
    \textsuperscript{1}C$^2$DL, Institute of Automation, Chinese Academy of Sciences, Beijing, China\\
    \textsuperscript{2}School of Artificial Intelligence, University of Chinese Academy of Sciences \\
    \textsuperscript{3}Department of Gastrointestinal Surgery Fujian Provincial Cancer Hospital \\
    \textsuperscript{4} College of Computer and Data Science,  Fuzhou University \\
}

\begin{document}
\maketitle
\begin{abstract}

Synthetic data generation has become a cornerstone for advancing large language models. However, the lack of the quantitative analysis for error tolerance became a critical bottleneck. Consequently, current filtering strategies fluctuate between two extremes: they are either overly aggressive, risking the exclusion of potentially valuable samples, or overly permissive, failing to eliminate erroneous samples effectively. To bridge this gap, this paper introduces \textbf{A}tomic \textbf{T}ree \textbf{O}peration \textbf{M}odeling (\textbf{ATOM}), a framework that decomposes data into functional units ($f(x)\rightarrow y$). ATOM distinguishes \textit{benign Operand $x$ perturbations} from \textit{fatal Operator $f$ perturbations}. The former are needlessly discarded by aggressive filtering, while the latter slip through permissive filtering. Our experiments reveal a \textbf{double dissociation}: models are robust to operand perturbations but collapse under operator perturbations. By prioritizing operator over aggressive operand precision, our ATOM-synthesized data outperforms rigorous baselines (e.g., +3.1\% gain over LIMA), suggesting that operator diversity matters more than operand precision. Our code is available at \href{https://github.com/Lut-hub/ATOM}{GitHub}.

\end{abstract}

\section{Introduction}
% The scaling laws of  have shifted the field's focus from compute constraints to data constraints. As high-quality human-generated data approaches exhaustion, synthetic data generation—distilling capabilities from strong models—has become the dominant paradigm for post-training. However, this paradigm faces a fundamental bottleneck: teacher models inevitably produce hallucinations, polluting the training signal with factual errors.

The ability of Large Language Models (LLMs) heavily relies on training data \citep{ grattafiori2024llama3herdmodels, openai2024gpt4technicalreport, qwen2.5, deepseekai2025deepseekr1incentivizingreasoningcapability, qwen3, geminiteam2025geminifamilyhighlycapable}. As high-quality human-generated data becomes increasingly scarce, the field has largely adopted synthetic data generation \citep{alpaca, zhou2023limaalignment, wang2023selfinstructaligninglanguagemodels, xu2024magpiealignmentdatasynthesis, xu2025wizardlmempoweringlargepretrained, maosongcao-etal-2025-condor} to distill capabilities from stronger models. However, a major bottleneck remains: teacher models inevitably hallucinate and introduce factual errors.

\begin{figure}[t]
    \centering
    \includegraphics[width=0.95\linewidth]{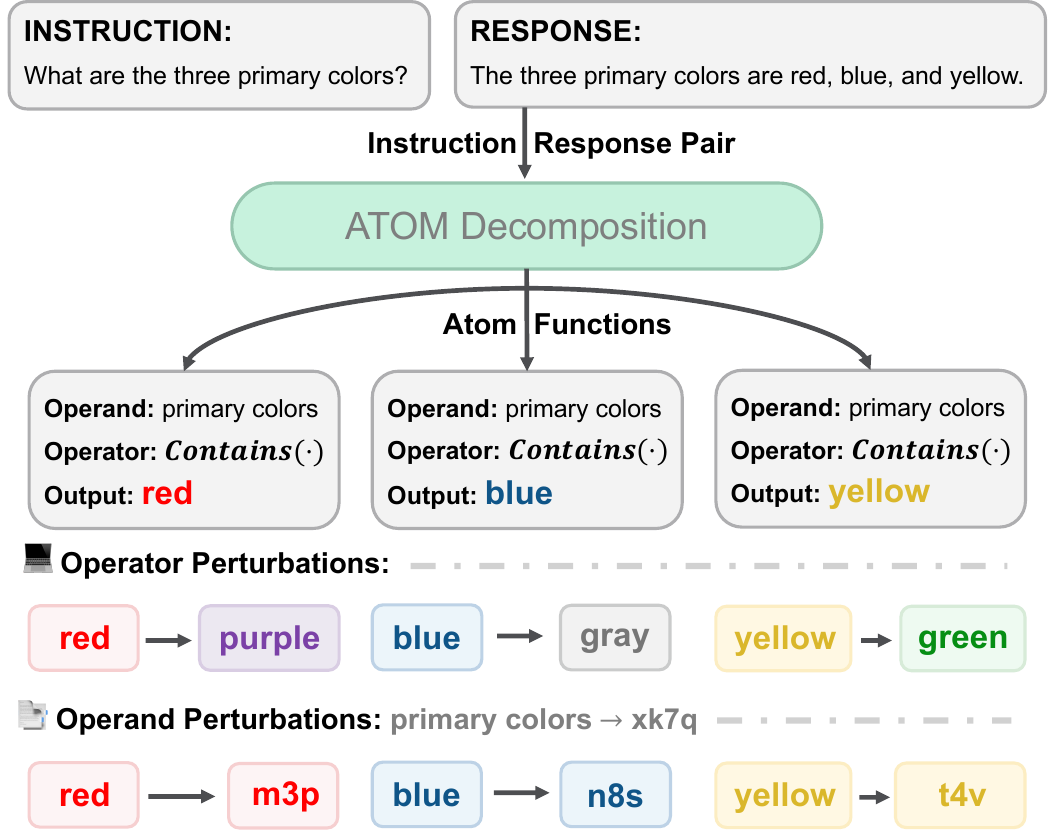}
    \caption{The illustration of extract atomic functions and two types of perturbations: \textit{Operand Perturbations} modify $x$ while updating $y$ to keep $f$ valid, e.g., $Contains(\text{xk7q})$ $ \rightarrow$ $\text{\textcolor{red}{m3p}/\textcolor{blue}{n8s}/\textcolor{orange}{t4v}}$. And \textit{Operator Perturbations} keep $x$ fixed but modify $y$ to violate $f$, e.g., $Contains(\text{primary colors}) \rightarrow  \text{\textcolor{purple}{purple}/\textcolor{gray}{gray}/\textcolor{green}{green}}$.}
    \label{fig:figure1}
    \vspace{-10pt}
\end{figure}

Studies such as LIMA \citep{zhou2023limaalignment} underscored that data quality is a decisive factor in supervised fine-tuning (SFT). However, they lack a framework to characterize which errors actually harm SFT. Lacking such an error tolerance criterion \citep{6685834}, current filtering 
methodologies are forced into two extremes: the first is \textit{aggressive filtering} \citep{zhou2023limaalignment, xu2024magpiealignmentdatasynthesis, maosongcao-etal-2025-condor}, which risks discarding valuable samples. The second is \textit{permissive filtering} \citep{li-etal-2024-superfiltering, zhao2024long, li-etal-2025-scar}, which retains harmful ones. Unfortunately, existing quantifying error tolerance methods still fall short: they are limited to either coarse, qualitative assessments that neglect error quantification \citep{cho-2024-unveiling, raghavendra2025revisiting}, or linguistic surface taxonomies that overlook essential logical properties \citep{alajrami2025finetuningnoisyinstructionseffects}. \textbf{More importantly, principled filtering hinges on quantifying which errors a model can tolerate and to what extent, which remains undefined.}

To rigorously identify the error tolerance in synthetic data, this paper proposes an \textbf{A}tomic \textbf{T}ree \textbf{O}peration \textbf{M}odeling (\textbf{ATOM}) framework, grounded in Problem Space Theory, which models problem-solving as applying operators to transform inputs into outputs. As shown in Figure~\ref{fig:figure1}, ATOM decomposes instruction-response pairs into atomic operations:
$f(x) \rightarrow y$. For example, the response ``The three primary colors are red, blue, and yellow.'' is decomposed as
$Contains(\text{primary colors}) \rightarrow \text{red/blue/yellow}$. ATOM makes error tolerance quantifiable by turning each response into countable operation units, where we can precisely control the error type and perturbation ratio. Specifically, based on whether the logical operator $f$ is preserved, ATOM defines two error types: \textit{Operand Perturbations} modify $x$ while updating $y$ to keep $f$ valid, e.g., $Contains(\text{xk7q})$ $ \rightarrow$ $\text{\textcolor{red}{m3p}/\textcolor{blue}{n8s}/\textcolor{orange}{t4v}}$. And \textit{Operator Perturbations} keep $x$ fixed but modify $y$ to violate $f$, e.g., $Contains(\text{primary colors}) \rightarrow  \text{\textcolor{purple}{purple}/\textcolor{gray}{gray}/\textcolor{green}{green}}$. %Our experiments show that models are robust to \textit{operand perturbations} but collapse under \textit{operator perturbations}. 
This taxonomy allows us to investigate and compare the model's tolerance to \textit{operand and operator perturbations}.

Specifically, this paper constructs an ATOM operator tree to generate Atomic-QA. Explicit atomic operations of Atomic-QA enable controlled injection of \textit{operand and operator perturbations}. Experiments on these two variants reveal a clear asymmetry: models are highly tolerant to \textit{operand perturbations}, with downstream performance remaining stable even when operands are replaced by random strings and yielding only a \textbf{2.0\%} relative decrease. Centered Kernel Alignment analysis \citep{pmlr-v97-kornblith19a} further shows that the resulting internal representations remain aligned with those learned from the original Atomic-QA. In contrast, \textit{operator perturbations} substantially degrade performance, causing a \textbf{18.2\%} relative decrease. These findings support an operator-centered principle: useful synthetic data should preserve valid and diverse operators rather than strict operand precision.

Guided by the above operator-centered principle, this paper further validate the effectiveness of synthetic data with Atomic-QA and Atomic-LIFD subsets that preserve broad operator coverage from the ATOM operator tree, retaining structural diversity often discarded by \textit{aggressive filtering}. Across four model backbones and 10 benchmark tasks, these ATOM-guided datasets consistently outperform standard baselines, including Alpaca \citep{alpaca}, LIMA \citep{zhou2023limaalignment}, WizardLM \citep{xu2025wizardlmempoweringlargepretrained} and Magpie\citep{xu2024magpiealignmentdatasynthesis}.

%Finally, this paper examines whether this finding generalizes to the real synthesis data setting. The Atomic-X pipeline is applied to existing instruction and math datasets, producing Atomic-Alpaca and Atomic-MetaMath with controlled perturbations. The results further confirm that increasing operator \textit{perturbations leads} to severe model collapse. Additionally, existing \textit{permissive filtering} methods struggle to detect these atomic operator faults. 
We also apply the generalizable Atomic-X pipeline to existing instruction and math datasets, producting Atomic-Alpaca and Atomic-MetaMath with the controlled perturbations. The results further confirm our findings and prove the generalization of the ATOM.
Our contributions are as follows:

\begin{itemize}[leftmargin=*, nolistsep]
\setlength{\itemsep}{1mm}
    % \item \textbf{A Quantifiable Atomic Framework:} We introduce \textbf{ATOM} to decompose data into atomic operations ($f(x) \rightarrow y$). This formalism provides the first precise quantification between \textit{Operand Perturbations} and \textit{Operator Perturbations}.
    
    %\item \textbf{Controllable Error Injection:} ATOM's decomposition enables targeted perturbation at the atomic level. We construct Atomic-QA and apply the Atomic-X pipeline to Alpaca and MetaMath, providing controlled experimental platforms for the error tolerance analysis.
    \item \textbf{ATOM Framework:} We construct \textbf{ATOM}, a unified framework for defining, quantifying, and controlling error tolerance in synthetic data.
    
    \item \textbf{Operator–Operand Asymmetry:} We discover that models are sensitive to \textit{operator perturbations} but robust to \textit{operand perturbations}.

    \item \textbf{Operator-Centered Synthesis:} Guided by this principle, we construct Atomic-QA, which achieves strong performance. And this paper further generalize the error tolerance findings to existing datasets via the Atomic-X pipeline.
\end{itemize}

\section{Related Work}
\paragraph{Construction and Filtering of Synthetic Data}
Prior research establishes that SFT prioritizes data quality over sheer scale, with works like LIMA \citep{zhou2023limaalignment} and Textbooks Are All You Need \citep{gunasekar2023textbooksneed} proving that effective alignment can emerge from small, high-quality corpora. Consequently, pipelines such as Self-Instruct \citep{wang2023selfinstructaligninglanguagemodels} and WizardLM \citep{xu2025wizardlmempoweringlargepretrained} employ generate-then-filter strategies to synthesize data, while other approaches focus on selection via heuristic-based strategies \citep{zhao2024long, li-etal-2024-superfiltering, li-etal-2025-scar} or contribution estimation \citep{dai2025improvinginfluencebasedinstructiontuning, chen2025migautomaticdataselection, jiang2025importanceawaredataselectionefficient}. However, current methods lack a quantitative framework for defining data quality, leading to a trade-off between overly aggressive filtering and overly permissive retention.

\paragraph{Learning Mechanisms in Noisy Data}
Prior work on noisy SFT investigates various settings. FTNI \citep{alajrami2025finetuningnoisyinstructionseffects} found that perturbed instructions can surprisingly improve performance. TInt \citep{havrilla2024understandingeffectnoisellm} and FACO \citep{cho2024unveilingimitationlearningexploring} explored model robustness to erroneous reasoning chains. Recent evidence challenges the superficial alignment hypothesis \citep{zhou2023limaalignment}, showing that SFT goes beyond style adaptation to genuinely enhance reasoning and follow pre-training scaling laws \citep{chen2025extractingunderstandingsuperficialknowledge,raghavendra2025revisiting,Zhang_2026}. However, current research lacks fine-grained analysis, failing to determine noise thresholds. Details are listed in Appendix \ref{sec:exrelated}.

\section{Methodology}
\begin{figure*}[!ht]
    \centering
    \includegraphics[width=0.94\textwidth]{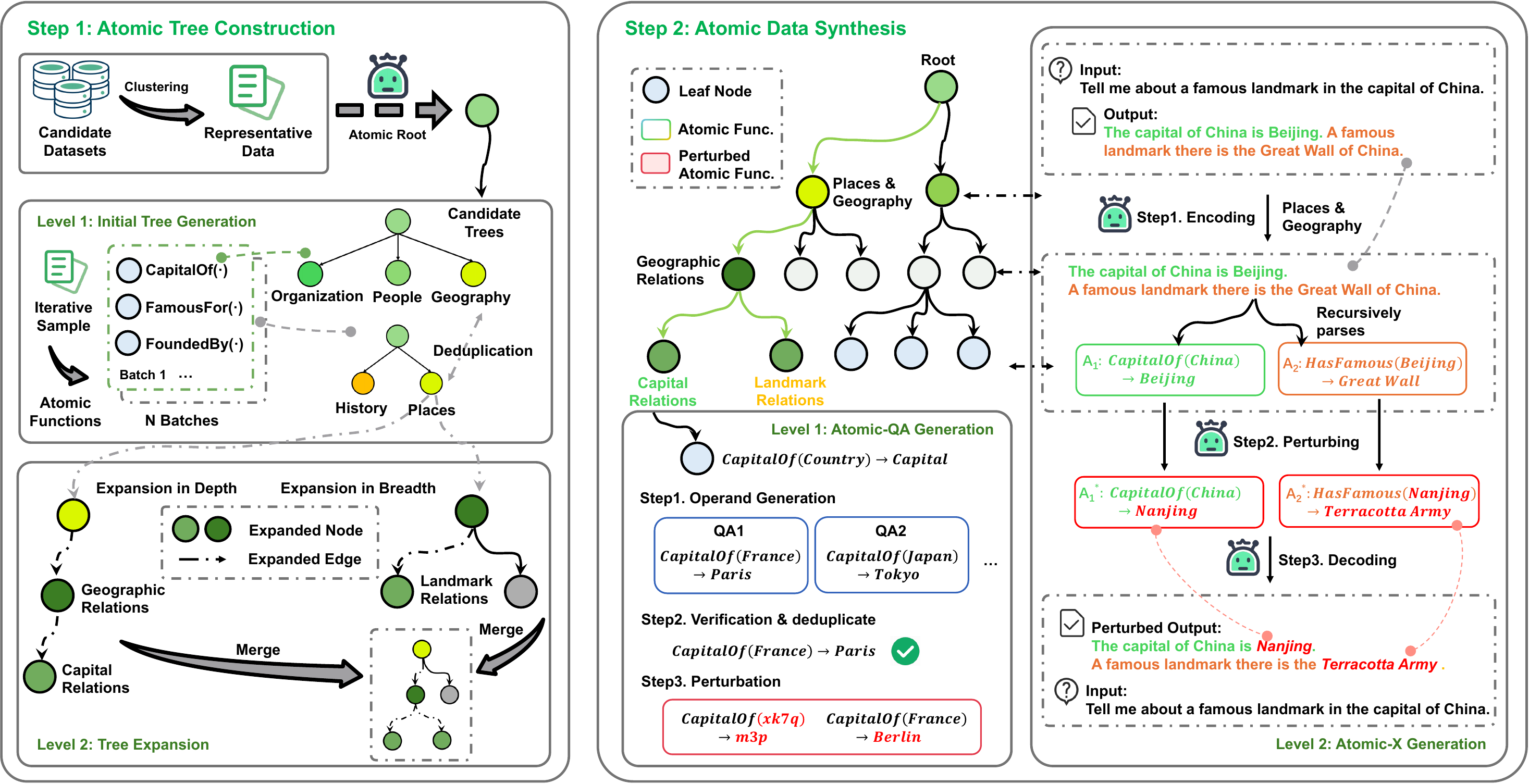}
    \caption{The pipeline of our \textbf{A}tomic \textbf{T}ree \textbf{O}peration \textbf{M}odeling (\textbf{ATOM}) framework. ATOM consists of two phases: 1) Atomic Tree Construction, which builds an operator hierarchy where leaf nodes define concrete atomic operators, and 2) Atomic Data Synthesis, which supports both Atomic-QA and Atomic-X.}
    \label{fig:pipeline}
    \vspace{-10pt}
\end{figure*}

This section details the \textbf{A}tomic \textbf{T}ree \textbf{O}peration \textbf{M}odeling (\textbf{ATOM}) framework. As depicted in Figure \ref{fig:pipeline}, the proposed framework operates in two distinct phases: 1) Atomic Tree Construction ($\S$\ref{ATC}) and 2) Atomic Data Synthesis ($\S$\ref{ADS}).

\subsection{Atomic Tree Construction} \label{ATC}
To formalize the construction process, we first introduce the concept of the Atomic Function.

\subsubsection{Atomic Function}
\begin{defn}[Atomic Function]
    An Atomic Function is defined as the minimal, logically indivisible unit of execution, denoted as a triplet $\mathcal{A} = \langle x, f, y \rangle$, where we have $f(x) \rightarrow y$.
\end{defn}
Formally, it represents a deterministic mapping:

\begin{itemize}[leftmargin=*, nolistsep]
\setlength{\itemsep}{1mm}
    \item $f$ (\textbf{The Operator}): The abstract logical rule or relational predicate that governs the transformation. Operators are reasoning patterns that define what transformation to apply, e.g., $CapitalOf(\cdot)$, $FamousFor(\cdot)$, or $FoundedBy(\cdot)$ in Figure~\ref{fig:pipeline}.
    \item $x$ (\textbf{The Operand}):  The concrete, substitutable input value upon which the operator acts. Operands are domain-specific factual content: they specify which particular instance the operator processes, and can be replaced without altering the underlying logic, e.g., \textit{Paris}, $100$, \textit{primary colors}.
    \item $y$ (\textbf{The Output}): The deterministic result of applying $f$ to $x$: $CapitalOf(\text{France}) \rightarrow \text{Paris}$.
\end{itemize}
An Atomic Function is defined such that any further decomposition of the operator $f$ would result in a loss of semantic intent or invalidate the executable logic within the task context. As shown in Figure~\ref{fig:figure1}, the instruction \textit{List the primary colors} with response \textit{Red, blue, and yellow} is decomposed into atomic function: $Contains(\text{primary colors})$ $\rightarrow$ $\text{\{red, blue, yellow\}}$, where $Contains$ is the \textbf{operator} and primary colors is the \textbf{operand}. An \textit{operand perturbation} jointly modifies $(x, y)$ so that $f(x')=y'$ still holds (e.g., xk7q) while preserving the operator, yielding $Contains(\text{xk7q})$ $\rightarrow$ $\text{\textcolor{red}{m3p}/\textcolor{blue}{n8s}/\textcolor{orange}{t4v}}$. This result is factually different but logically coherent. In contrast, an \textit{operator perturbation} would modify y alone so that $f(x)\neq y'$, for example $Contains(\text{primary colors})$ $\rightarrow$ $\text{\textcolor{purple}{purple}/\textcolor{gray}{gray}/\textcolor{green}{green}}$, producing an incoherent result that violates the operator's logic. \textbf{To further clarify the atomic function, more examples are in Appendix \ref{sec:appendix-atomicexample}.}

%This definition holds across various domains. For example, in the knowledge domain, $CapitalOf(\text{France}) \rightarrow \text{Paris}$ retrieves the target entity based on a given subject and relation.

\subsubsection{Pipeline of Atomic Tree Construction}
We organize atomic operators into a tree-structured hierarchy. For instance, the path from tree: \texttt{Root → Geography \& Places → Geographic Relations → Capital Relations} reaches a leaf operator $CapitalOf(\cdot)$. This hierarchical structure naturally aligns with the recursive composition of complex tasks: a multi-step reasoning problem decomposes through intermediate categories until reaching semantically irreducible leaf operators. Consequently, our goal is to construct a complete operator hierarchy whose leaf nodes cover the space of atomic operators required by a given domain.

% As illustrated in Step I of Figure \ref{fig:pipeline}, the phase commences with \textbf{Representative Data Selection}. To mitigate computational overhead while ensuring coverage, we encode candidate datasets using bge-m3 \citep{bge-m3} and apply KCenterGreedy clustering to extract high-quality samples that embody diverse operational logic.

%In Step 1 of Figure \ref{fig:pipeline}, the phase commences with \textbf{Representative Data Selection}.

As shown in Figure~\ref{fig:pipeline}, Step 1 starts with Representative Data Selection. To mitigate computational overhead while ensuring coverage, we encode candidate seed datasets using bge-m3 \citep{bge-m3}. We then apply KCenterGreedy \citep{sener2018active} clustering to select samples that maximize the coverage of the semantic space and ensure the extracted subset diversity.

After representative data selection, \textbf{Level 1: Initial Tree Generation} employs an iterative strategy in which o3-mini processes batches of atomic functions sampled from representative data and assigns each function to a category. Within each batch, the categories are organized into candidate sub-trees, where upper nodes represent coarser domains and finer nodes represent more specific operator categories. For example, atomic functions such as $CapitalOf(\text{France}) \rightarrow \text{Paris}$ are grouped under the category path \texttt{Root $\rightarrow$ Geography \& Places $\rightarrow$ $CapitalOf(\cdot)$}, where the leaf category anchors the concrete operator. As shown in the left-middle of Figure~\ref{fig:pipeline}, the candidate sub-trees produced from successive batches are iteratively merged into a global tree. During each merge, an LLM-based semantic assessment module identifies category nodes in the incoming sub-tree whose semantics overlap with existing ones (e.g., \texttt{Places \& Geography} v.s. \texttt{Geography \& Places}). \textbf{The initial trees are ultimately verified by 4 human experts' votes, detailed in Appendix \ref{sec:appendix-ConstructionPrompt}.}

After initialization, in \textbf{Level 2: Tree Expansion}, the generated tree is expanded with the strict atomic constraints defined in our  definition. This phase iteratively refines category nodes through a dual-path expansion strategy inspired by EpiCoder \citep{wang2025epicoder}: Expansion in Depth for granularity and Expansion in Breadth for coverage. Following the example above, the initial tree contains a coarse category path \texttt{Root $\rightarrow$ Geography \& Places}. \textit{Expansion in Depth} recursively subdivides category nodes by inserting finer-grained relation classes, yielding \texttt{ $\cdots \rightarrow$ Geographic Relations $\rightarrow$ Capital Relations}, progressively narrowing the scope until each leaf is irreducible as judged by the LLM or capped at a maximum depth. \textit{Expansion in Breadth} supplements sibling category nodes at the same level, e.g., adding sibling nodes like \texttt{Landmark Relations} alongside \texttt{Capital Relations}. As shown in the left-bottom panel of Figure \ref{fig:pipeline}, the resulting expanded nodes are consolidated back into the main hierarchy through a final merge operation. \textbf{Detailed examples and prompts are shown in Appendix \ref{sec:appendix-ConstructionPrompt}. Leaf nodes are sampled and verified by expert voting.}

\subsection{Atomic-QA Synthesis}
\label{ADS}
After obtaining the atomic tree, we synthesize training data and its controlled perturbation variants through the following pipeline.
\label{sec:Atomic-QA}
Atomic-QA Generation synthesizes training data starting from the leaf nodes of the atomic tree. Each leaf node defines a concrete atomic operator $f$ like $CapitalOf(\cdot)$, and the generation proceeds through:

\textbf{Step 1 (QA Generation):} Given a leaf node and its associated operator $f$, we first generate diverse operands $x$ in its domain, and then compute the corresponding output $y = f(x)$ to form atomic triplets $\mathcal{A} = \langle x, f, y \rangle$. For example, for $f = CapitalOf$, we generate operands such as France, Japan, Brazil and compute their outputs Paris, Tokyo, Bras\'{\i}lia.

\textbf{Step 2 (Verification):} To ensure data quality and diversity, we deduplicate overlapping operands $x$ across neighboring nodes and rigorously verify the logical validity of the generated triplets, producing the \textbf{Atomic-QA} corpus.

\textbf{Step 3 (Perturbation):} We inject noise to create variants. Specifically, \textit{Operand perturbation} replaces $x$ with a random string and adjusts $y$ while preserving $f$, e.g.,~$CapitalOf(\text{sa9v}) \rightarrow \text{m3p}$, giving structurally coherent but factually meaningless samples. \textit{Operator perturbation} alters y while keeping $x$ fixed, such that the resulting pair $(x, y')$ encodes a mapping that contradicts $f$, e.g.,~$CapitalOf(\text{France}) \rightarrow \text{Berlin}$. At last, two variants, \textit{Operand-Perturbed Atomic-QA} and \textit{Operator-Perturbed Atomic-QA} are generated.

\section{Experimental Setup}
\label{sec:expsetup}
In the following sections, we exploit experiments to answer the following research questions (\textbf{RQs}): 
\begin{itemize}[leftmargin=*, nolistsep]
\setlength{\itemsep}{1mm}
\item \textbf{RQ1:} Which plays the dominant role in training: the \textbf{Operand} ($x$) or the \textbf{Operator} ($f$)? ($\S$\ref{sec:Measuring})
\item \textbf{RQ2:} Does ATOM-synthesized data outperform aggressively-filtered baselines? ($\S$\ref{sec:Quantifying})
\item \textbf{RQ3:} Whether the error tolerance extends to the real synthesis data setting. ($\S$\ref{sec:Atomic-X})
\end{itemize}

%i) Does aggressive filtering for strict factual correctness come at the cost of structural diversity, thereby hindering model performance? ii) Which plays the dominant role in training: the specific \textbf{Operand} ($x$) or the abstract \textbf{Operator} ($f$)? iii) What is the quantitative threshold for noise tolerance in real synthesis data setting, and at what point does operator corruption trigger model collapse?

\begin{figure*}[!ht]
    \centering
    \includegraphics[width=0.9\textwidth]{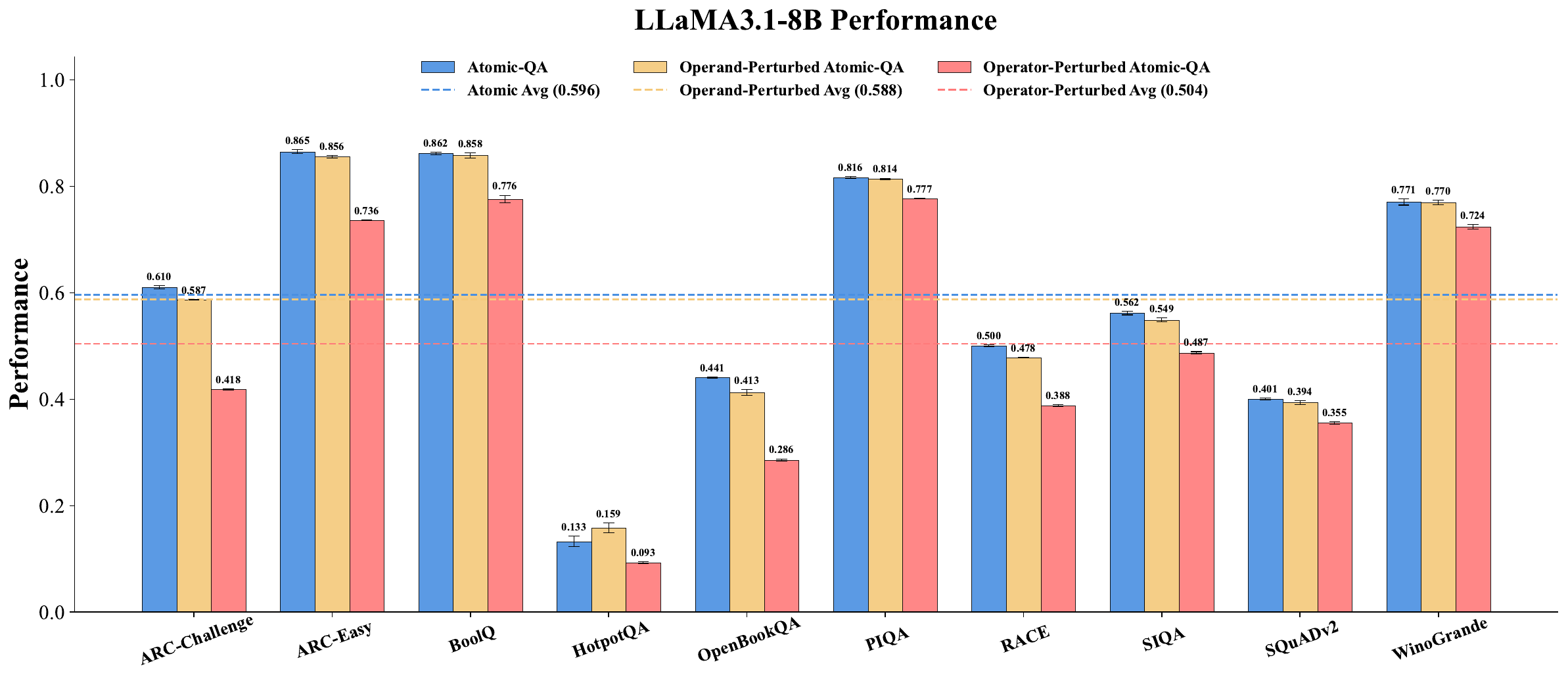}
    \caption{
    Performance comparison of LLaMA3.1 across Atomic-QA, Operand-Perturbed Atomic-QA, and Operator-Perturbed Atomic-QA. Dashed lines denote the average performance for each setting.
    }
    \label{fig:factqa_llama31}
    \vspace{-10pt}
\end{figure*}

\textbf{Dataset:}
To address RQ1 and RQ2, we construct an operator tree following Section \ref{sec:Atomic-QA} using FB15k \citep{fb15k}, ZsRE \citep{zsre}, and wiki\_recent \citep{wang2023easyedit} as seeds, and synthesize the Atomic-QA dataset (1.37M instances) along with its operand-perturbed and operator-perturbed variants. For RQ3, we construct Atomic-Alpaca and Atomic-MetaMath via the Atomic-X pipeline ($\S$\ref{sec:Atomic-X}), injecting operator errors at controlled rates into real-world complex data. See Appendix \ref{sec:DatasetGeneration} for details.

\textbf{Model:}
To ensure generalizability, we evaluate LLaMA3.1 (8B) \citep{grattafiori2024llama3herdmodels}, Qwen2.5 (7B \& 14B) \citep{qwen2.5}, and Qwen3 (8B) \citep{qwen3}. Training is conducted via Llama-Factory \citep{zheng2024llamafactory} and evaluation via the LM Evaluation Harness \citep{eval-harness}. Configuration details are in Appendix \ref{sec:ModelTraining}.


\section{Operator–Operand Asymmetry}
\label{sec:Measuring}
To evaluate operator–operand tolerance during training, we construct two controlled variants of Atomic-QA using the perturbation strategies defined in $\S$\ref{sec:Atomic-QA}. \textit{Operand-Perturbed Atomic-QA} applies operand perturbation: each operand $x$ is replaced with a random string of the same character length while preserving the operator $f$, producing data that is factually meaningless but structurally coherent. \textit{Operator-Perturbed Atomic-QA} applies operator perturbation: following \citet{xie2024controlledautomatictaskspecificsynthetic}, we use an LLM to rewrite the answers and verify hallucinations, thereby corrupting the operator logic $f$ while keeping operands intact. This data is superficially fluent but logically incoherent.

\subsection{Performance Comparison}
Figure \ref{fig:factqa_llama31} presents the comparative results on LLaMA3.1-8B, revealing a double dissociation: models are robust to operand corruption but collapse under operator corruption. \textbf{Results of other models are listed in Figure \ref{fig:qwen2_5_7b}-\ref{fig:qwen3_8b} of Appendix.}

\textbf{Findings 1: Robustness to Operand Noise:}
The resilience to operand noise is not limited to simple tasks but generalizes consistently across downstream evaluations of varying complexity. As shown in Figure \ref{fig:factqa_llama31}, the performance of Operand-Perturbed Atomic-QA closely tracks the baseline Atomic-QA across all datasets. On reasoning-heavy tasks like ARC-Challenge, the performance gap is minimal ($0.587$ v.s. $0.610$), and on the complex multi-hop reasoning task HotpotQA, the model trained on random strings even slightly outperforms the baseline ($0.159$ v.s. $0.133$). This demonstrates that even in complex scenarios, the model's performance relies primarily on acquiring the operator logic $f$ rather than memorizing the specific atomic operands $x$. The results shown in Figures \ref{fig:qwen2_5_7b}, \ref{fig:qwen2_5_14b}, and \ref{fig:qwen3_8b} indicate that our findings generalize across different model structures.

\begin{figure*}[!ht]
    \centering
    \includegraphics[width=0.9\textwidth]{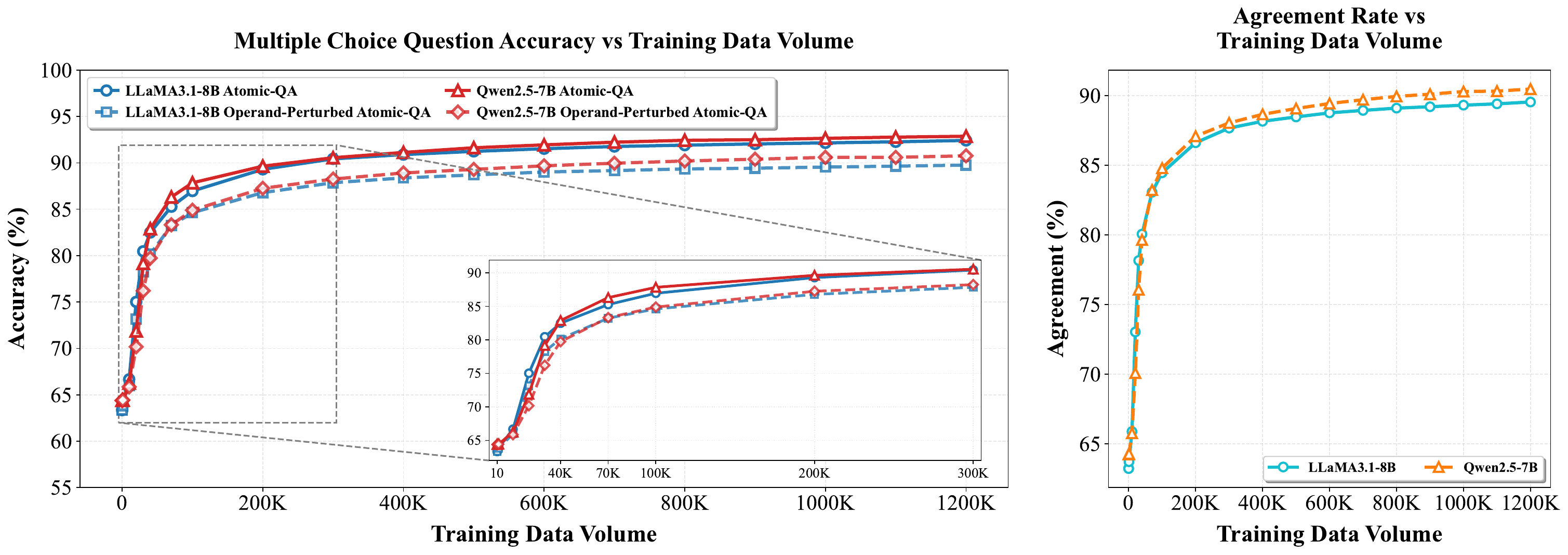}
    \caption{
       MCQ accuracy vs.\ training data scale (main plot), with model agreement under replaced vs.\ original entity settings. Agreement Rate is the proportion of questions both models answer correctly.}
    \label{fig:mcq_main_acc}
    \vspace{-10pt}
\end{figure*}
\textbf{Findings 2: Sensitivity to Operator Noise:}
In stark contrast, when trained on Operator-Perturbed Atomic-QA, where the structural logic is disrupted (corrupting the operator $f$), performance plummets to \textbf{50.4\%}. This represents a substantial absolute decline of \textbf{9.2\%} (a relative drop of $\sim$18.2\%), which highlights the model's acute sensitivity to the operation perturbation which may corrupt the structural integrity in the training data.

\subsection{Mechanism Validation}

One might concern that Operand-Perturbed Atomic-QA's strong performance comes from metric hacking, surface coherence (Appendix~\ref{sec:appendix-ppl-analysis}), or superficial heuristics rather than structural integrity. To rule this out, we analyze the training dynamics and latent representations, showing that the model learns consistently with the unperturbed setting.

\textbf{Behavioral Alignment:}
We constructed a held-out diagnostic test set by sampling subsets from Atomic-QA and rewriting them into multiple-choice questions (MCQs). We then monitored the validation metrics throughout the training process. As shown in Figure \ref{fig:mcq_main_acc} (Left), the accuracy curves for models trained on both Atomic-QA and Operand-Perturbed Atomic-QA rise in unison, converging to nearly identical performance levels. Complementing this, Figure \ref{fig:mcq_main_acc} (Right) illustrates the Agreement Rate—the proportion of test instances where both models predict correctly. The steady increase in agreement demonstrates that as training progresses, the models do not just achieve similar scores. They converge on the same behavioral patterns. This synchrony indicates that the optimization landscape is dominated by the structural operator $f$, while the precise values of atomic operands $x$ play a comparatively limited role in learning.

\begin{figure}[!ht]
    \centering
    \includegraphics[width=0.95\linewidth]{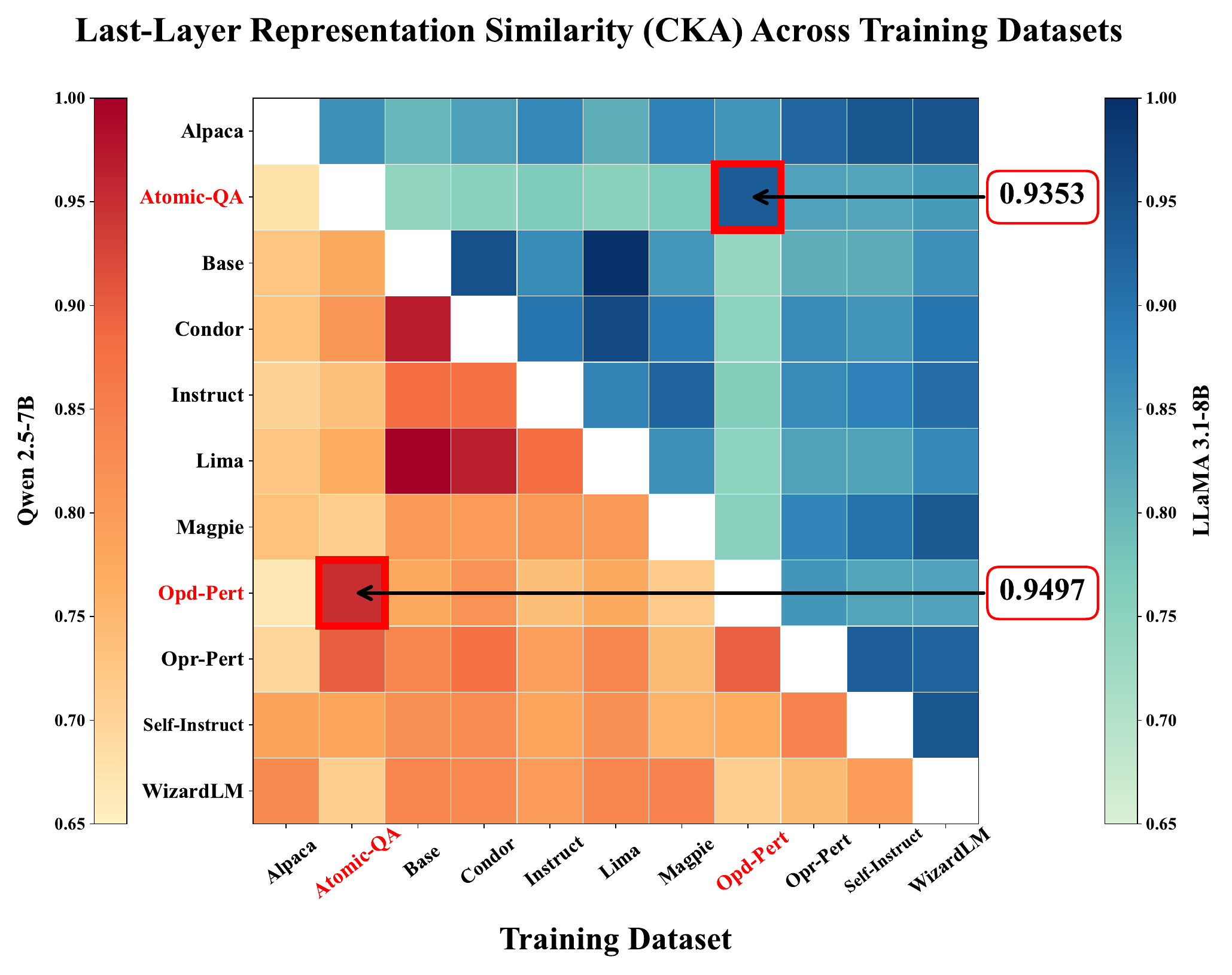}
    \caption{
        CKA  similarity of the output hidden states between different trained models. The lower part represents the CKA similarity of Qwen 2.5-7B, while the upper part represents LLaMA 3.1-8B.
    }
    \label{fig:cka_heatmap}
    \vspace{-10pt}
\end{figure}

\textbf{Representational Alignment:}
Do these models merely behave similarly, or do they actually learn the same internal representations? We utilize Centered Kernel Alignment (CKA) \citep{pmlr-v97-kornblith19a} to measure the similarity of the last-layer hidden states between different models. Figure \ref{fig:cka_heatmap} presents the CKA heatmap. Strikingly, the model trained on Operand-Perturbed Atomic-QA exhibits an exceptionally high similarity score (e.g., \textbf{0.9497} on Qwen2.5-7B) with the model trained on the standard \textit{Atomic-QA}. This alignment suggests that the models have learned equivalent internal representations, despite their differing training data. %In other words, the model effectively filters out the atomic perturbations as "local noise" while robustly capturing the "global structure," treating the random strings and real entities as functionally equivalent operands within the learned reasoning schema. 
Training dynamics of PPL and CKA are in Appendix~\ref{ref:addresult} (Figures \ref{fig:ppl_trend}, \ref{fig:variance_aggregate}). Together, these findings reveal an asymmetry and motivate an operator-centered principle: \textbf{synthetic data should prioritize operator coverage over strict operand filtering.}

\section{Effectiveness of Atomic-QA}
\label{sec:Quantifying}
% The double dissociation in $\S$\ref{sec:Measuring} motivates an operator-centered design principle: synthetic data should prioritize broad operator coverage over strict operand-level filtering. We now validate this principle by comparing ATOM-synthesized data against aggressively filtered baselines, from two angles: degradation in model performance across 10 benchmarks and four backbones, and exclusion of valuable samples as measured by diversity metrics.

We validate the operator-centered principle through extensive comparisons against existing baselines.

% Lighter color scheme for heatmap visualization
% Ranges from light red (poor) through white (average) to light green (excellent)

% Negative performance (red tones - lighter)
\definecolor{verypoor}{RGB}{239,138,98}      % Light coral red - Very Poor (-4)
\definecolor{poor}{RGB}{252,174,145}         % Lighter coral - Poor (-3)
\definecolor{belowavg}{RGB}{253,204,192}     % Very light coral - Below Avg (-2)
\definecolor{slightbelow}{RGB}{254,229,217}  % Pale coral - Slightly Below (-1)

% Neutral performance (white)
\definecolor{average}{RGB}{255,255,255}      % White - Average (0)

% Positive performance (green tones - lighter)
\definecolor{slightabove}{RGB}{229,245,224}  % Pale green - Slightly Above (+1)
\definecolor{aboveavg}{RGB}{199,233,192}     % Light green - Above Avg (+2)
\definecolor{good}{RGB}{161,217,155}         % Medium light green - Good (+3)
\definecolor{excellent}{RGB}{116,196,118}    % Lighter green - Excellent (+4)
\begin{table*}[!ht]
  \centering
  \scriptsize
  \caption{Overall performance across all training datasets on LLaMA3.1-8B and Qwen3-8B. Color-coded cells (\textcolor{excellent}{green}/\textcolor{verypoor}{red}) are used to indicate performance above or below the mean across methods (All Average), with darker colors representing better or worse performance. \textbf{The results for Qwen2.5-7B and Qwen2.5-14B are in Table \ref{tab:qwen3_performance}.}}
  \resizebox{0.93\linewidth}{!}{
  \begin{tabular}{c|c*{11}{c}}
    \toprule
    \multicolumn{2}{c|}{\textbf{Model Family}} & \makecell{\textbf{ARC-}\\\textbf{Challenge}} & \makecell{\textbf{ARC-}\\\textbf{Easy}} & \textbf{BoolQ} & \makecell{\textbf{Hotpot-}\\\textbf{QA}} & \makecell{\textbf{Open-}\\\textbf{BookQA}} & \textbf{PIQA} & \textbf{RACE} & \textbf{SIQA} & \textbf{SQuADv2} & \makecell{\textbf{Wino-}\\\textbf{Grande}} & \textbf{Avg} \\
    \hline
    
    % LlaMA3.1-8B 部分
   \multirow{13}{*}{\rotatebox{90}{\textbf{LLaMA3.1-8B}}}

& Base
& \cellcolor{average}0.549
& \cellcolor{average}0.847
& \cellcolor{average}0.820
& \cellcolor{average}0.106
& \cellcolor{average}0.366
& \cellcolor{average}0.807
& \cellcolor{average}0.433
& \cellcolor{average}0.529
& \cellcolor{average}0.375
& \cellcolor{average}0.765
& \cellcolor{average}0.560 \\

& All Average & \cellcolor{average}0.564 & \cellcolor{average}0.850 & \cellcolor{average}0.855 & \cellcolor{average}0.127
& \cellcolor{average}0.387 & \cellcolor{average}0.812 & \cellcolor{average}0.464 & \cellcolor{average}0.548 & \cellcolor{average}0.381 & \cellcolor{average}0.768 & \cellcolor{average}0.576 \\
\cline{2-13}

& Instruct & \cellcolor{slightbelow}0.558 & \cellcolor{aboveavg}0.854 & \cellcolor{excellent}0.870 & \cellcolor{excellent}0.209 & \cellcolor{poor}0.362 & \cellcolor{slightbelow}0.811 & \cellcolor{average}0.464 & \cellcolor{average}0.548 & \cellcolor{excellent}0.420 & \cellcolor{slightbelow}0.764 & \cellcolor{aboveavg}0.586 \\

& LIMA & \cellcolor{slightbelow}0.552 & \cellcolor{slightbelow}0.848 & \cellcolor{verypoor}0.828 & \cellcolor{slightabove}0.134 & \cellcolor{poor}0.368 & \cellcolor{belowavg}0.808 & \cellcolor{verypoor}0.432 & \cellcolor{poor}0.529 & \cellcolor{slightbelow}0.377 & \cellcolor{good}0.773 & \cellcolor{belowavg}0.565 \\
& Condor & \cellcolor{slightbelow}0.563 & \cellcolor{slightabove}0.854 & \cellcolor{average}0.855 & \cellcolor{belowavg}0.117 & \cellcolor{slightbelow}0.382 & \cellcolor{slightbelow}0.811 & \cellcolor{slightabove}0.470 & \cellcolor{slightbelow}0.546 & \cellcolor{slightbelow}0.378 & \cellcolor{slightabove}0.770 & \cellcolor{slightbelow}0.575 \\
& Magpie & \cellcolor{slightabove}0.575 & \cellcolor{slightbelow}0.845 & \cellcolor{slightabove}0.859 & \cellcolor{slightbelow}0.122 & \cellcolor{slightbelow}0.380 & \cellcolor{belowavg}0.808 & \cellcolor{average}0.464 & \cellcolor{good}0.563 & \cellcolor{slightbelow}0.374 & \cellcolor{slightbelow}0.761 & \cellcolor{slightbelow}0.575 \\

& WizardLM & \cellcolor{aboveavg}0.580 & \cellcolor{excellent}0.862 & \cellcolor{slightabove}0.858 & \cellcolor{poor}0.107 & \cellcolor{slightabove}0.396 & \cellcolor{excellent}0.823 & \cellcolor{aboveavg}0.478 & \cellcolor{excellent}0.573 & \cellcolor{slightabove}0.385 & \cellcolor{slightbelow}0.766 & \cellcolor{aboveavg}0.583 \\
& Alpaca & \cellcolor{slightbelow}0.562 & \cellcolor{slightbelow}0.846 & \cellcolor{good}0.867 & \cellcolor{belowavg}0.115 & \cellcolor{aboveavg}0.406 & \cellcolor{good}0.819 & \cellcolor{slightabove}0.469 & \cellcolor{excellent}0.576 & \cellcolor{aboveavg}0.393 & \cellcolor{excellent}0.776 & \cellcolor{aboveavg}0.583 \\
& Self-Instruct & \cellcolor{verypoor}0.469 & \cellcolor{verypoor}0.811 & \cellcolor{poor}0.839 & \cellcolor{verypoor}0.097 & \cellcolor{verypoor}0.352 & \cellcolor{verypoor}0.803 & \cellcolor{verypoor}0.430 & \cellcolor{verypoor}0.511 & \cellcolor{verypoor}0.321 & \cellcolor{verypoor}0.739 & \cellcolor{verypoor}0.537 \\
& Wiki & \cellcolor{slightbelow}0.562 & \cellcolor{aboveavg}0.856 & \cellcolor{belowavg}0.845 & \cellcolor{slightabove}0.132 & \cellcolor{belowavg}0.370 & \cellcolor{slightbelow}0.810 & \cellcolor{belowavg}0.453 & \cellcolor{belowavg}0.530 & \cellcolor{slightabove}0.386 & \cellcolor{excellent}0.778 & \cellcolor{slightbelow}0.572 \\
& Wiki-Rewrite & \cellcolor{slightbelow}0.560 & \cellcolor{aboveavg}0.855 & \cellcolor{belowavg}0.847 & \cellcolor{verypoor}0.100 & \cellcolor{poor}0.364 & \cellcolor{slightbelow}0.811 & \cellcolor{slightbelow}0.456 & \cellcolor{poor}0.528 & \cellcolor{slightabove}0.388 & \cellcolor{excellent}0.777 & \cellcolor{slightbelow}0.569 \\
& Atomic-LIFD & \cellcolor{excellent}0.607 & \cellcolor{good}0.858 & \cellcolor{excellent}0.875 & \cellcolor{slightabove}0.133 & \cellcolor{excellent}0.440 & \cellcolor{average}0.812 & \cellcolor{excellent}0.494 & \cellcolor{good}0.568 & \cellcolor{belowavg}0.364 & \cellcolor{slightabove}0.770 & \cellcolor{good}0.592 \\
& Atomic-QA & \cellcolor{excellent}0.610 & \cellcolor{excellent}0.865 & \cellcolor{aboveavg}0.862
& \cellcolor{slightabove}0.133 & \cellcolor{excellent}0.441 & \cellcolor{aboveavg}0.816 & \cellcolor{excellent}0.500 & \cellcolor{good}0.562 & \cellcolor{aboveavg}0.401 & \cellcolor{good}0.771 & \cellcolor{excellent}0.596 \\

\hline

    % Qwen3-8B 部分
    \multirow{13}{*}{\rotatebox{90}{\textbf{Qwen3-8B}}} 
    & Base & \cellcolor{average}0.631 & \cellcolor{average}0.871 & \cellcolor{average}0.871 & \cellcolor{average}0.233 & \cellcolor{average}0.370 & \cellcolor{average}0.799 & \cellcolor{average}0.473 & \cellcolor{average}0.566 & \cellcolor{average}0.444 & \cellcolor{average}0.743 & \cellcolor{average}0.600 \\
    
& All Average & \cellcolor{average}0.631 & \cellcolor{average}0.870 & \cellcolor{average}0.875 & \cellcolor{average}0.183 & \cellcolor{average}0.375 & \cellcolor{average}0.798 & \cellcolor{average}0.489 & \cellcolor{average}0.562 & \cellcolor{average}0.420 & \cellcolor{average}0.746 & \cellcolor{average}0.595 \\
\cline{2-13}

& Instruct & \cellcolor{slightbelow}0.629 & \cellcolor{slightabove}0.873 & \cellcolor{belowavg}0.871 & \cellcolor{belowavg}0.114 & \cellcolor{slightabove}0.378 & \cellcolor{verypoor}0.782 & \cellcolor{slightabove}0.491 & \cellcolor{aboveavg}0.567 & \cellcolor{aboveavg}0.430 & \cellcolor{verypoor}0.707 & \cellcolor{slightbelow}0.584 \\

& LIMA & \cellcolor{slightabove}0.632 & \cellcolor{average}0.870 & \cellcolor{slightbelow}0.873 & \cellcolor{aboveavg}0.237 & \cellcolor{slightbelow}0.374 & \cellcolor{good}0.803 & \cellcolor{slightbelow}0.478 & \cellcolor{slightbelow}0.558 & \cellcolor{excellent}0.445 & \cellcolor{good}0.755 & \cellcolor{slightabove}0.602 \\
& Condor & \cellcolor{slightabove}0.637 & \cellcolor{aboveavg}0.875 & \cellcolor{average}0.875 & \cellcolor{aboveavg}0.230 & \cellcolor{slightbelow}0.374 & \cellcolor{slightabove}0.800 & \cellcolor{slightbelow}0.486 & \cellcolor{aboveavg}0.567 & \cellcolor{good}0.435 & \cellcolor{aboveavg}0.751 & \cellcolor{aboveavg}0.603 \\
& Magpie & \cellcolor{slightbelow}0.629 & \cellcolor{slightbelow}0.869 & \cellcolor{average}0.875 & \cellcolor{belowavg}0.125 & \cellcolor{poor}0.352 & \cellcolor{average}0.798 & \cellcolor{slightbelow}0.482 & \cellcolor{slightbelow}0.553 & \cellcolor{good}0.436 & \cellcolor{slightbelow}0.742 & \cellcolor{slightbelow}0.586 \\

& WizardLM & \cellcolor{slightabove}0.639 & \cellcolor{average}0.870 & \cellcolor{aboveavg}0.877 & \cellcolor{verypoor}0.053 & \cellcolor{belowavg}0.360 & \cellcolor{slightabove}0.799 & \cellcolor{slightabove}0.491 & \cellcolor{good}0.572 & \cellcolor{aboveavg}0.428 & \cellcolor{excellent}0.756 & \cellcolor{slightbelow}0.585 \\
& Alpaca & \cellcolor{slightabove}0.634 & \cellcolor{slightabove}0.871 & \cellcolor{good}0.880 & \cellcolor{verypoor}0.042 & \cellcolor{belowavg}0.362 & \cellcolor{aboveavg}0.803 & \cellcolor{slightabove}0.495 & \cellcolor{excellent}0.580 & \cellcolor{excellent}0.445 & \cellcolor{good}0.753 & \cellcolor{slightbelow}0.586 \\
& Self-Instruct & \cellcolor{verypoor}0.515 & \cellcolor{verypoor}0.820 & \cellcolor{verypoor}0.866 & \cellcolor{poor}0.103 & \cellcolor{verypoor}0.338 & \cellcolor{poor}0.787 & \cellcolor{verypoor}0.439 & \cellcolor{verypoor}0.513 & \cellcolor{verypoor}0.341 & \cellcolor{poor}0.721 & \cellcolor{verypoor}0.544 \\

& Wiki & \cellcolor{slightabove}0.638 & \cellcolor{aboveavg}0.876 & \cellcolor{poor}0.869 & \cellcolor{slightabove}0.191 & \cellcolor{slightbelow}0.366 & \cellcolor{aboveavg}0.801 & \cellcolor{average}0.489 & \cellcolor{aboveavg}0.570 & \cellcolor{slightabove}0.424 & \cellcolor{excellent}0.759 & \cellcolor{slightabove}0.598 \\
& Wiki-Rewrite & \cellcolor{slightabove}0.641 & \cellcolor{aboveavg}0.877 & \cellcolor{belowavg}0.871 & \cellcolor{slightabove}0.212 & \cellcolor{slightbelow}0.370 & \cellcolor{slightabove}0.800 & \cellcolor{slightabove}0.497 & \cellcolor{aboveavg}0.568 & \cellcolor{slightbelow}0.417 & \cellcolor{excellent}0.758 & \cellcolor{slightabove}0.601 \\
& Atomic-LIFD & \cellcolor{excellent}0.671 & \cellcolor{excellent}0.886 & \cellcolor{excellent}0.884 & \cellcolor{excellent}0.364 & \cellcolor{excellent}0.420 & \cellcolor{slightabove}0.800 & \cellcolor{excellent}0.523 & \cellcolor{aboveavg}0.568 & \cellcolor{belowavg}0.398 & \cellcolor{aboveavg}0.751 & \cellcolor{excellent}0.626 \\
& Atomic-QA & \cellcolor{excellent}0.675 & \cellcolor{excellent}0.886 & \cellcolor{excellent}0.883 & \cellcolor{excellent}0.334 & \cellcolor{excellent}0.435 & \cellcolor{excellent}0.804 & \cellcolor{good}0.515 & \cellcolor{aboveavg}0.567& \cellcolor{slightabove}0.423 & \cellcolor{good}0.753 & \cellcolor{excellent}0.627 \\
\bottomrule

  \end{tabular}
  }
  \label{tab:overall_performance}
  \vspace{-5pt}
\end{table*}
\subsection{Atomic-QA Outperforms Baselines}
Tables \ref{tab:overall_performance} and \ref{tab:qwen3_performance} report evaluation results across 10 benchmarks and four backbones.

\textbf{Surpassing Aggressive Filtering Baselines:} We verify the operator-centered principle by comparing Atomic-QA against LIMA, a canonical aggressive-filtering baseline. On LLaMA3.1-8B, Atomic-QA achieves an absolute gain of \textbf{+3.1\%} over LIMA (56.5\% average). This trend persists across model scales, with Atomic-QA outperforming LIMA by 1.7\% on Qwen2.5-7B and 2.5\% on Qwen3-8B. This gap underscores that aggressively filtering methods discard valuable training signals, capping the model's potential. 
To further isolate the effect of operator coverage, we also evaluate Atomic-LIFD, a compact subset of Atomic-QA selected to retain diverse operator structures. 
\textbf{Despite using minimal training tokens (Appendix \ref{sec:trainingtoken}), Atomic-LIFD achieves competitive results. This confirms that operator-level structural diversity is the driver of performance gains.}

\textbf{Superiority over Standard Synthetic Baselines:}
Compared to widely used synthetic datasets like WizardLM, Atomic-QA demonstrates consistent superiority. For instance, on the Qwen2.5-7B, Atomic-QA (61.1\%) surpasses WizardLM \citep{xu2025wizardlmempoweringlargepretrained} (56.8\%) by over \textbf{4\%}. Notably, Atomic-QA excels in reasoning-intensive tasks like OpenBookQA \citep{mihaylov2018suitarmorconductelectricity} and ARC-Challenge \citep{clark2018thinksolvedquestionanswering}, suggesting that preserving operator diversity is more critical.

\begin{figure}[t]
    \centering
    \includegraphics[width=0.85\linewidth]{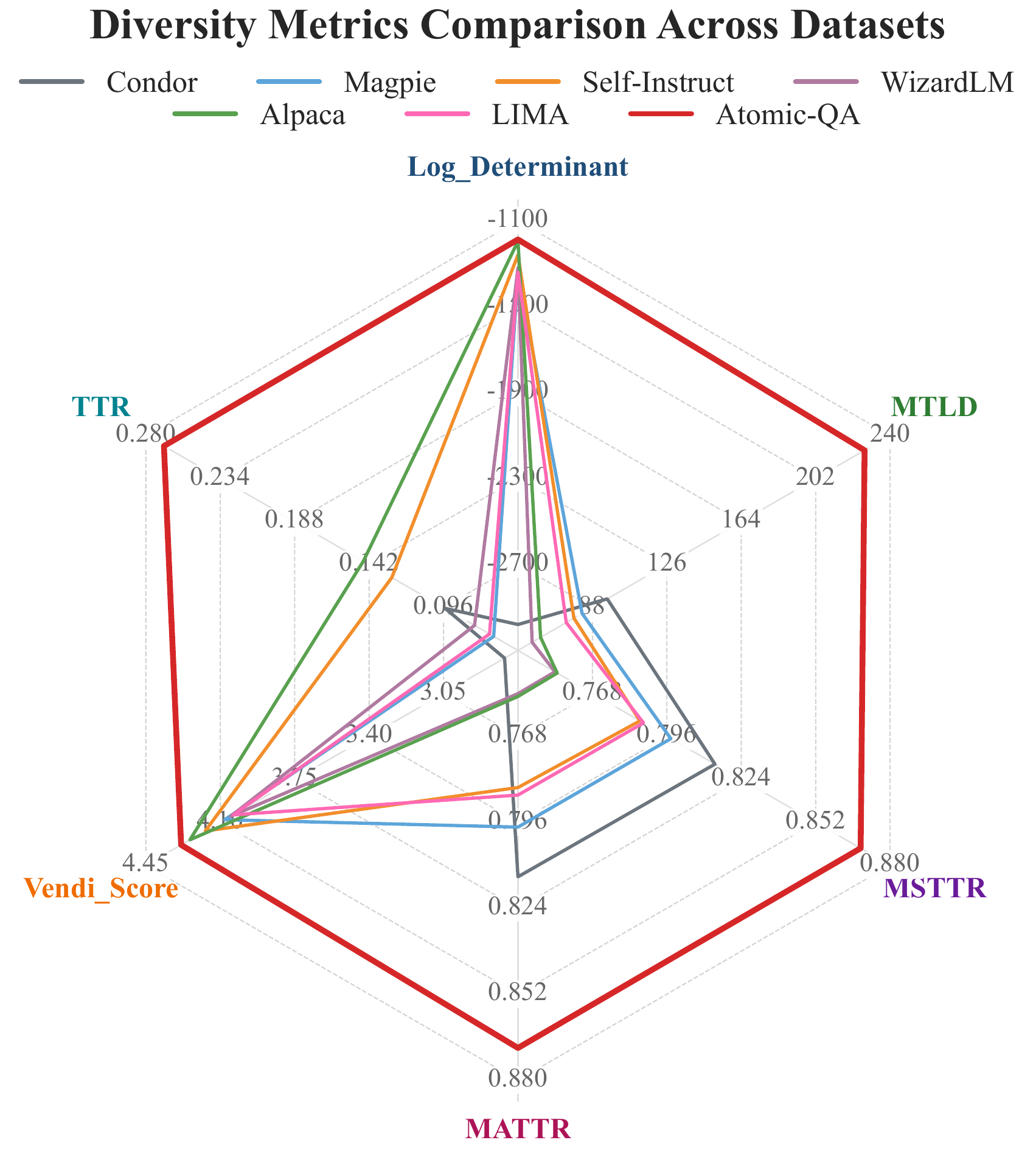}
    \caption{Radar plot of corpus diversity.}
    %Radar-plot diversity comparison across datasets, illustrating differences in Log-Determinant, MTLD, MSTTR, MATTR, TTR, and Vendi\_Score among training corpora.}
    \label{fig:final_diverse}
    \vspace{-10pt}
\end{figure}

\subsection{Exclusion of Potentially Valuable Samples}

Figure \ref{fig:final_diverse} visualizes corpus diversity using Vendi Score \citep{friedman2023the}, Log Determinant \citep{yang-etal-2025-measuring}, TTR-variants \citep{Covington2010CuttingTG}, and MTLD \citep{mccarthy2010mtld}. Atomic-QA achieves the highest scores across all six metrics and encompasses all baselines. While aggressively filtered datasets show contracted diversity. This confirms that tree-based operator enumeration preserves the structural samples that filtering discards. However, a practical question remains: what is the quantitative noise threshold in real-world synthetic data?

\section{Generalization to Existing Datasets}
\label{sec:Atomic-X}
To answer RQ3, we introduce the \textbf{Atomic-X} pipeline, which decomposes existing datasets into atomic operators and injects controlled perturbations. In right panel of Figure \ref{fig:pipeline}, Atomic-X adopts an Encoder–Perturber–Decoder pipeline:

\textbf{Step 1 (Encoding):} A encoder takes an instruction-response pair from an existing dataset and performs a top-down decomposition of the response into atomic operators, guided by the operator tree. The model recursively parses the content into distinct logical branches, where each leaf corresponds to an atomic operator in the tree, yielding an ordered sequence $\mathbb{A} = \{\mathcal{A}_1, \mathcal{A}_2, \dots, \mathcal{A}_n\}$.

\textbf{Step 2 (Perturbing):} A perturbation function is applied to the extracted atomic functions. As shown in the formula $\text{Perturbing}(\mathcal{A}_i) = \mathcal{A}'_i$, we inject operator-level noise (e.g., modifying operations from $1+1=2$ to $1+1=3$) to generate corrupted atomic functions.

\textbf{Step 3 (Decoding):} The decoder reconstructs the output: First, the perturbed atomic functions are reranked based on the magnitude of their logical impact as $\mathbb{A}^* = \{\mathcal{A}_1^*, \mathcal{A}_2^*, \dots, \mathcal{A}_n^*\}$. Next, the model generates specific decoding instructions for modification. Finally, a deterministic replace tool utilizes these instructions to edit the original text, producing the final \textit{Perturbed Output}. \textbf{Further details are in Appendix \ref{sec:appendix-SynthesisPrompt}.} We apply Atomic-X to Alpaca \citep{alpaca} and MetaMath \citep{yu2023metamath}, producing \textbf{Atomic-Alpaca} and \textbf{Atomic-MetaMath}. This enables us to analyze noise tolerance on multi-step reasoning task.

\begin{figure}[!ht]
    \centering
    \includegraphics[width=0.9\linewidth]{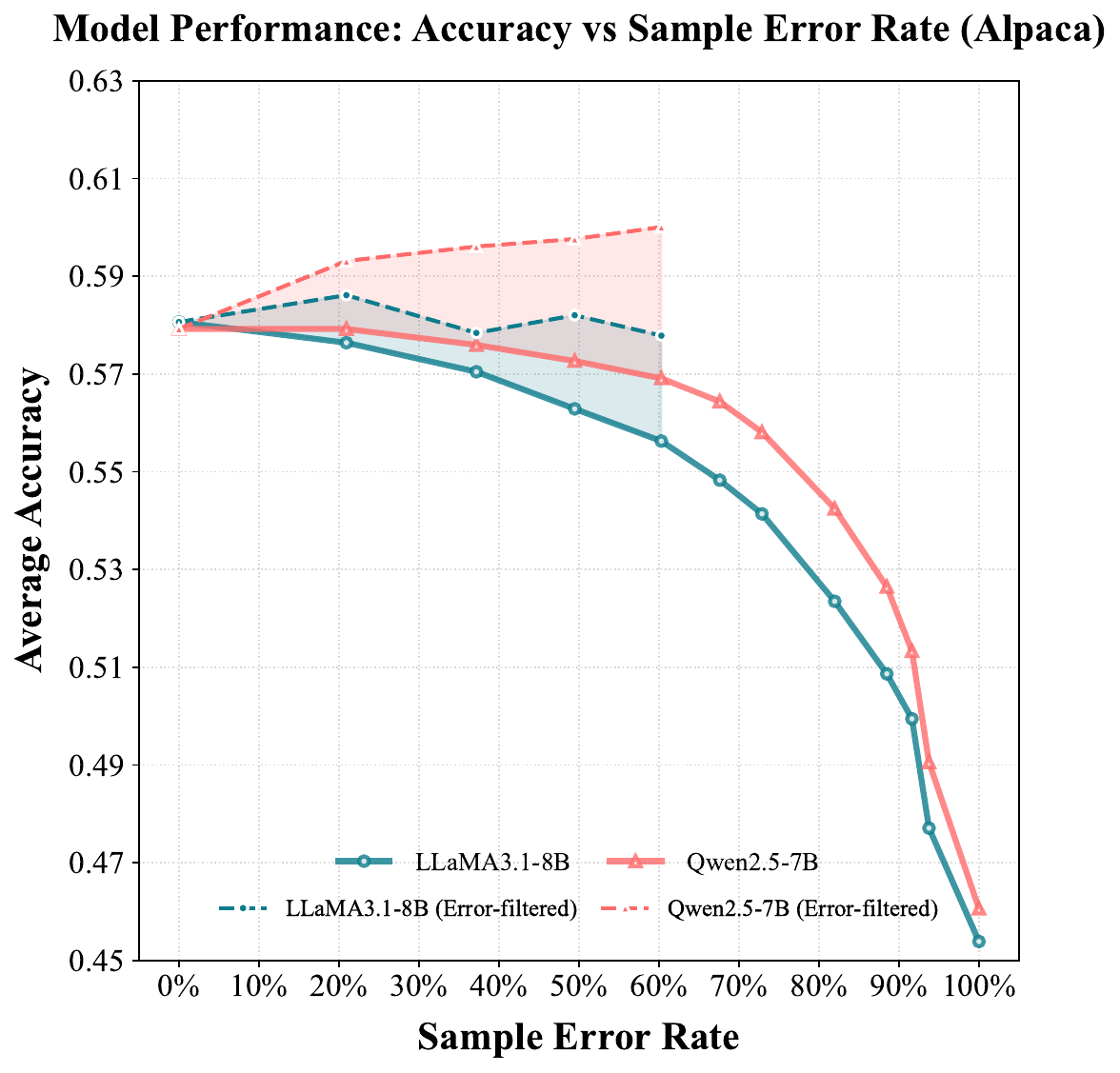}
    \caption{
     %Average accuracy for LLaMA 3.1-8B and Qwen 2.5-7B on Atomic-Alpaca with operator perturbations at controlled rates (§\ref{sec:Atomic-X}). Dashed lines: upper bound from samples. See Figure \ref{fig:real_error_sensitivity-math} for Atomic-MetaMath.
     Average accuracy for LLaMA 3.1-8B and Qwen 2.5-7B on Atomic-Alpaca with operator perturbations at controlled rates (\S\ref{sec:Atomic-X}). Dashed lines: upper bound from training only on correct samples, truncated at 60\% to approximate the solid lines' data scale. See Figure \ref{fig:real_error_sensitivity-math} for Atomic-MetaMath.
    }
    \label{fig:real_error_sensitivity-alpaca}
    \vspace{-10pt}
\end{figure}

\subsection{Operator Noise Tolerance Limits}
To verify that operator sensitivity (§\ref{sec:Measuring}) generalizes to real data, we adopt Beta-distribution-based sampling as described in Appendix \ref{sec:Beta-Distribution} and inject operator erros through varying rates of atomic-operation perturbations into Alpaca and MetaMath. We analyze the impact of increasing atomic perturbations rates (from 0\% to 100\%) on model performance, visualized in Figure \ref{fig:real_error_sensitivity-alpaca}. Results reveal a distinct two-stage non-linear degradation pattern.

\textbf{Stage 1: Gradual Logic Degradation:}
As shown in Figure \ref{fig:real_error_sensitivity-alpaca}, injecting perturbations into atomic operations results in a slow but observable performance decay during the initial stage ($0\% \rightarrow 40\%$ noise). Unlike operand, operator perturbations disrupt models' training process more severely, causing proportional capability loss.

\textbf{Stage 2: Structural Collapse:}
As the error rate exceeds a critical threshold (roughly $>60\%$ in Figure \ref{fig:real_error_sensitivity-alpaca}), the degradation pattern shifts distinctly, characterized by a significantly steepened negative slope. Visually, this is represented by the rapid plunge in accuracy towards the tail end of the curve ($60\% \rightarrow 100\%$). This sharp performance drop suggests that accumulated operator perturbations severely impair the model’s ability to learn reliable patterns beyond this threshold.
\begin{figure}[!ht]
    \centering
    \includegraphics[width=0.91\linewidth]{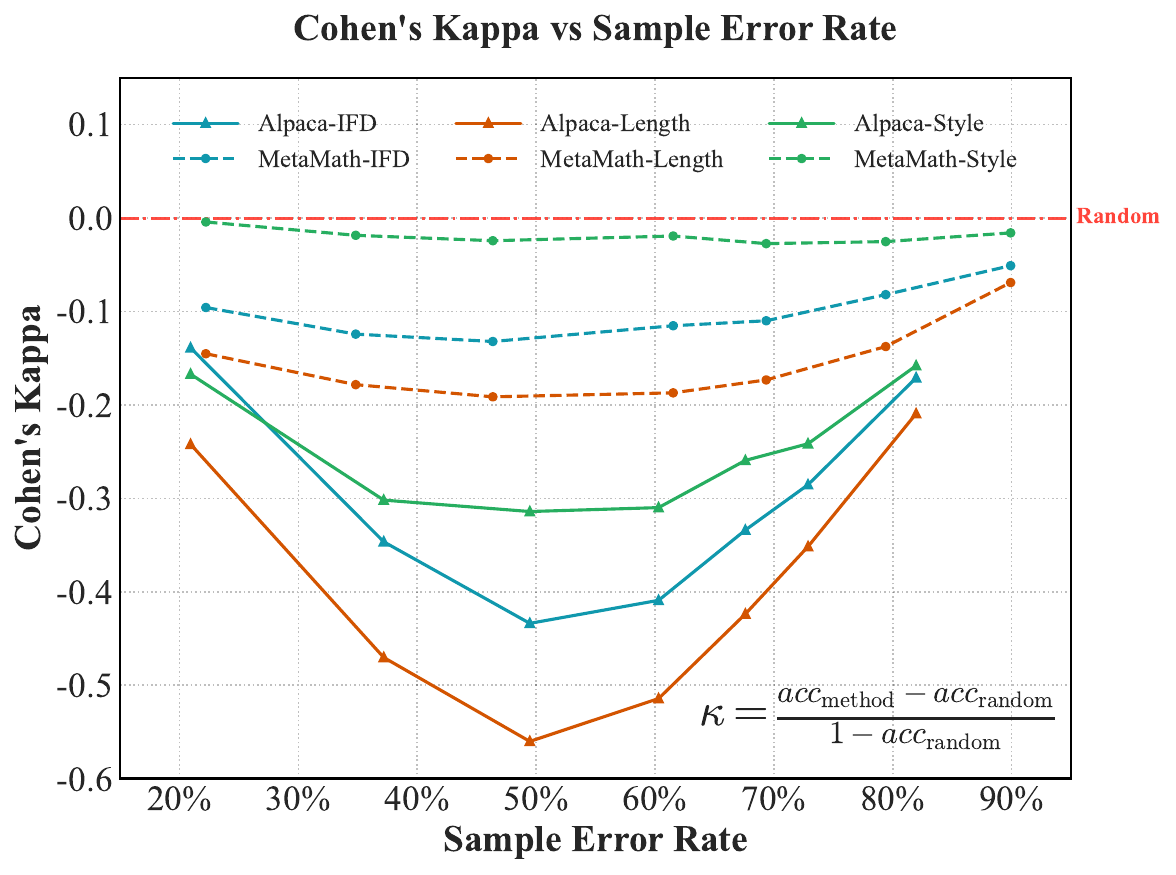}
    \caption{
      Cohen's Kappa between filtering methods (IFD, Length, Style) and ground-truth error labels as a function of sample error rate, evaluated on \textbf{Atomic-Alpaca} and \textbf{Atomic-MetaMath}.
    }
    \label{fig:kappa}
    \vspace{-10pt}
\end{figure}

\subsection{Upper Bound vs. Filtering Reality}
% To assess filtering potential, we established an upper bound by retraining on error-filtered subsets with sample error rates less than 60\% (dashed lines in Figure \ref{fig:real_error_sensitivity-alpaca}). These models matched the noise-free baseline, confirming that degradation drives from operator errors. However, as shown in Figure \ref{fig:kappa}, current filtering strategies based on metrics such as IFD \citep{li-etal-2024-superfiltering}, sequence length \citep{zhao2024long}, and style-consistency \citep{li-etal-2025-scar} fail to identify these atomic faults. The resulting negative Cohen's Kappa coefficients indicate that these algorithms are unable to differentiate between atomic logic faults and correct samples.

To assess filtering potential, we construct an oracle upper bound by discarding all samples containing operator errors based on ground-truth labels, and retraining on the remaining clean subset at each error rate (dashed lines in Figure \ref{fig:real_error_sensitivity-alpaca}). These oracle-filtered models fully recover the noise-free baseline, showing that operator-induced degradation is recoverable if erroneous samples can be identified. This sets the ceiling for any practical filter. Yet, as shown in Figure \ref{fig:kappa}, existing strategies cannot approach this ceiling: their negative Cohen's Kappa scores indicate they are unable to differentiate between fault and correct samples.

\section{Conclusion}

This study quantifies error tolerance in synthesis data, moving beyond the binary high-low quality definitions. We demonstrate that SFT is fundamentally a logic-learning process in atomic view where models learn functional atomic operators while specific operand-values are changeable variables. We show that models learn atomic logic even from randomized operands. Thus, synthetic data should favor operators over operands for generalizability.

\section*{Limitation}
Due to computational resource constraints, we do not conduct experiments on models with 32B parameters or larger. Furthermore, while our ATOM framework demonstrates robustness across general instruction following and mathematical reasoning tasks, we have not yet extended the atomic decomposition methodology to code generation or other domains, which we leave for future work.

\section*{Acknowledgement}
This work was supported by the National Natural Science Foundation of China (No.62406321) and CIPS-SMP-Zhipu Large Model Fund.

\bibliography{custom}

\begin{thebibliography}{57}
\providecommand{\natexlab}[1]{#1}

\bibitem[{Alajrami et~al.(2025)Alajrami, Tan, and Aletras}]{alajrami2025finetuningnoisyinstructionseffects}
Ahmed Alajrami, Xingwei Tan, and Nikolaos Aletras. 2025.
\newblock \href {https://arxiv.org/abs/2510.03528} {Fine-tuning on noisy instructions: Effects on generalization and performance}.
\newblock \emph{Preprint}, arXiv:2510.03528.

\bibitem[{Amini et~al.(2019)Amini, Gabriel, Lin, Koncel-Kedziorski, Choi, and Hajishirzi}]{amini2019mathqainterpretablemathword}
Aida Amini, Saadia Gabriel, Peter Lin, Rik Koncel-Kedziorski, Yejin Choi, and Hannaneh Hajishirzi. 2019.
\newblock \href {https://arxiv.org/abs/1905.13319} {Mathqa: Towards interpretable math word problem solving with operation-based formalisms}.
\newblock \emph{Preprint}, arXiv:1905.13319.

\bibitem[{Bai et~al.(2024)Bai, Lv, Zhang, Lyu, Tang, Huang, Du, Liu, Zeng, Hou, Dong, Tang, and Li}]{bai2024longbenchbilingualmultitaskbenchmark}
Yushi Bai, Xin Lv, Jiajie Zhang, Hongchang Lyu, Jiankai Tang, Zhidian Huang, Zhengxiao Du, Xiao Liu, Aohan Zeng, Lei Hou, Yuxiao Dong, Jie Tang, and Juanzi Li. 2024.
\newblock \href {https://arxiv.org/abs/2308.14508} {Longbench: A bilingual, multitask benchmark for long context understanding}.
\newblock \emph{Preprint}, arXiv:2308.14508.

\bibitem[{Bisk et~al.(2019)Bisk, Zellers, Bras, Gao, and Choi}]{bisk2019piqareasoningphysicalcommonsense}
Yonatan Bisk, Rowan Zellers, Ronan~Le Bras, Jianfeng Gao, and Yejin Choi. 2019.
\newblock \href {https://arxiv.org/abs/1911.11641} {Piqa: Reasoning about physical commonsense in natural language}.
\newblock \emph{Preprint}, arXiv:1911.11641.

\bibitem[{Cao et~al.(2025)Cao, Zhang, Li, Zhang, Liu, Duan, Zhang, and Chen}]{maosongcao-etal-2025-condor}
Maosong Cao, Taolin Zhang, Mo~Li, Chuyu Zhang, Yunxin Liu, Haodong Duan, Songyang Zhang, and Kai Chen. 2025.
\newblock \href {https://doi.org/10.18653/v1/2025.acl-long.1091} {Condor: Enhance {LLM} alignment with knowledge-driven data synthesis and refinement}.
\newblock In \emph{Proceedings of the 63rd Annual Meeting of the Association for Computational Linguistics (Volume 1: Long Papers)}, pages 22392--22412, Vienna, Austria. Association for Computational Linguistics.

\bibitem[{Cao et~al.(2024)Cao, Chen, Jin, Chen, Liu, and Zhao}]{cao2024mindtonguesdeepdive}
Pengfei Cao, Yuheng Chen, Zhuoran Jin, Yubo Chen, Kang Liu, and Jun Zhao. 2024.
\newblock \href {https://arxiv.org/abs/2411.17401} {One mind, many tongues: A deep dive into language-agnostic knowledge neurons in large language models}.
\newblock \emph{Preprint}, arXiv:2411.17401.

\bibitem[{Chen et~al.(2025{\natexlab{a}})Chen, Xiao, Zhang, Luo, Lian, and Liu}]{bge-m3}
Jianlv Chen, Shitao Xiao, Peitian Zhang, Kun Luo, Defu Lian, and Zheng Liu. 2025{\natexlab{a}}.
\newblock \href {https://arxiv.org/abs/2402.03216} {M3-embedding: Multi-linguality, multi-functionality, multi-granularity text embeddings through self-knowledge distillation}.
\newblock \emph{Preprint}, arXiv:2402.03216.

\bibitem[{Chen et~al.(2025{\natexlab{b}})Chen, Perin, Chen, Chen, Han, Hirata, Hong, and Kailkhura}]{chen2025extractingunderstandingsuperficialknowledge}
Runjin Chen, Gabriel~Jacob Perin, Xuxi Chen, Xilun Chen, Yan Han, Nina S.~T. Hirata, Junyuan Hong, and Bhavya Kailkhura. 2025{\natexlab{b}}.
\newblock \href {https://arxiv.org/abs/2502.04602} {Extracting and understanding the superficial knowledge in alignment}.
\newblock \emph{Preprint}, arXiv:2502.04602.

\bibitem[{Chen et~al.(2025{\natexlab{c}})Chen, Li, Hu, Ma, Ye, and Chen}]{chen2025migautomaticdataselection}
Yicheng Chen, Yining Li, Kai Hu, Zerun Ma, Haochen Ye, and Kai Chen. 2025{\natexlab{c}}.
\newblock \href {https://arxiv.org/abs/2504.13835} {Mig: Automatic data selection for instruction tuning by maximizing information gain in semantic space}.
\newblock \emph{Preprint}, arXiv:2504.13835.

\bibitem[{Cho(2024{\natexlab{a}})}]{cho-2024-unveiling}
Hyunsoo Cho. 2024{\natexlab{a}}.
\newblock \href {https://doi.org/10.18653/v1/2024.findings-acl.6} {Unveiling imitation learning: Exploring the impact of data falsity to large language model}.
\newblock In \emph{Findings of the Association for Computational Linguistics: ACL 2024}, pages 62--73, Bangkok, Thailand. Association for Computational Linguistics.

\bibitem[{Cho(2024{\natexlab{b}})}]{cho2024unveilingimitationlearningexploring}
Hyunsoo Cho. 2024{\natexlab{b}}.
\newblock \href {https://arxiv.org/abs/2404.09717} {Unveiling imitation learning: Exploring the impact of data falsity to large language model}.
\newblock \emph{Preprint}, arXiv:2404.09717.

\bibitem[{Clark et~al.(2019)Clark, Lee, Chang, Kwiatkowski, Collins, and Toutanova}]{clark2019boolqexploringsurprisingdifficulty}
Christopher Clark, Kenton Lee, Ming-Wei Chang, Tom Kwiatkowski, Michael Collins, and Kristina Toutanova. 2019.
\newblock \href {https://arxiv.org/abs/1905.10044} {Boolq: Exploring the surprising difficulty of natural yes/no questions}.
\newblock \emph{Preprint}, arXiv:1905.10044.

\bibitem[{Clark et~al.(2018)Clark, Cowhey, Etzioni, Khot, Sabharwal, Schoenick, and Tafjord}]{clark2018thinksolvedquestionanswering}
Peter Clark, Isaac Cowhey, Oren Etzioni, Tushar Khot, Ashish Sabharwal, Carissa Schoenick, and Oyvind Tafjord. 2018.
\newblock \href {https://arxiv.org/abs/1803.05457} {Think you have solved question answering? try arc, the ai2 reasoning challenge}.
\newblock \emph{Preprint}, arXiv:1803.05457.

\bibitem[{Cobbe et~al.(2021)Cobbe, Kosaraju, Bavarian, Chen, Jun, Kaiser, Plappert, Tworek, Hilton, Nakano, Hesse, and Schulman}]{cobbe2021trainingverifierssolvemath}
Karl Cobbe, Vineet Kosaraju, Mohammad Bavarian, Mark Chen, Heewoo Jun, Lukasz Kaiser, Matthias Plappert, Jerry Tworek, Jacob Hilton, Reiichiro Nakano, Christopher Hesse, and John Schulman. 2021.
\newblock \href {https://arxiv.org/abs/2110.14168} {Training verifiers to solve math word problems}.
\newblock \emph{Preprint}, arXiv:2110.14168.

\bibitem[{Covington and McFall(2010)}]{Covington2010CuttingTG}
Michael~A. Covington and Joe~D. McFall. 2010.
\newblock \href {https://api.semanticscholar.org/CorpusID:18924254} {Cutting the gordian knot: The moving-average type–token ratio (mattr)}.
\newblock \emph{Journal of Quantitative Linguistics}, 17:100 -- 94.

\bibitem[{Dai et~al.(2025)Dai, Zhang, Ma, and Peng}]{dai2025improvinginfluencebasedinstructiontuning}
Qirun Dai, Dylan Zhang, Jiaqi~W. Ma, and Hao Peng. 2025.
\newblock \href {https://arxiv.org/abs/2501.12147} {Improving influence-based instruction tuning data selection for balanced learning of diverse capabilities}.
\newblock \emph{Preprint}, arXiv:2501.12147.

\bibitem[{DeepSeek-AI et~al.(2025)DeepSeek-AI, Guo, Yang, Zhang, Song, Zhang, Xu, Zhu, Ma, Wang, Bi, Zhang, Yu, Wu, Wu, Gou, Shao, Li, Gao, Liu, Xue, Wang, Wu, Feng, Lu, Zhao, Deng, Zhang, Ruan, Dai, Chen, Ji, Li, Lin, Dai, Luo, Hao, Chen, Li, Zhang, Bao, Xu, Wang, Ding, Xin, Gao, Qu, Li, Guo, Li, Wang, Chen, Yuan, Qiu, Li, Cai, Ni, Liang, Chen, Dong, Hu, Gao, Guan, Huang, Yu, Wang, Zhang, Zhao, Wang, Zhang, Xu, Xia, Zhang, Zhang, Tang, Li, Wang, Li, Tian, Huang, Zhang, Wang, Chen, Du, Ge, Zhang, Pan, Wang, Chen, Jin, Chen, Lu, Zhou, Chen, Ye, Wang, Yu, Zhou, Pan, Li, Zhou, Wu, Ye, Yun, Pei, Sun, Wang, Zeng, Zhao, Liu, Liang, Gao, Yu, Zhang, Xiao, An, Liu, Wang, Chen, Nie, Cheng, Liu, Xie, Liu, Yang, Li, Su, Lin, Li, Jin, Shen, Chen, Sun, Wang, Song, Zhou, Wang, Shan, Li, Wang, Wei, Zhang, Xu, Li, Zhao, Sun, Wang, Yu, Zhang, Shi, Xiong, He, Piao, Wang, Tan, Ma, Liu, Guo, Ou, Wang, Gong, Zou, He, Xiong, Luo, You, Liu, Zhou, Zhu, Xu, Huang, Li, Zheng, Zhu, Ma, Tang, Zha, Yan, Ren, Ren, Sha, Fu, Xu, Xie, Zhang,
  Hao, Ma, Yan, Wu, Gu, Zhu, Liu, Li, Xie, Song, Pan, Huang, Xu, Zhang, and Zhang}]{deepseekai2025deepseekr1incentivizingreasoningcapability}
DeepSeek-AI, Daya Guo, Dejian Yang, Haowei Zhang, Junxiao Song, Ruoyu Zhang, Runxin Xu, Qihao Zhu, Shirong Ma, Peiyi Wang, Xiao Bi, Xiaokang Zhang, Xingkai Yu, Yu~Wu, Z.~F. Wu, Zhibin Gou, Zhihong Shao, Zhuoshu Li, Ziyi Gao, and 181 others. 2025.
\newblock \href {https://arxiv.org/abs/2501.12948} {Deepseek-r1: Incentivizing reasoning capability in llms via reinforcement learning}.
\newblock \emph{Preprint}, arXiv:2501.12948.

\bibitem[{Frenay and Verleysen(2014)}]{6685834}
Benoit Frenay and Michel Verleysen. 2014.
\newblock \href {https://doi.org/10.1109/TNNLS.2013.2292894} {Classification in the presence of label noise: A survey}.
\newblock \emph{IEEE Transactions on Neural Networks and Learning Systems}, 25(5):845--869.

\bibitem[{Friedman and Dieng(2023)}]{friedman2023the}
Dan Friedman and Adji~Bousso Dieng. 2023.
\newblock \href {https://openreview.net/forum?id=g97OHbQyk1} {The vendi score: A diversity evaluation metric for machine learning}.
\newblock \emph{Transactions on Machine Learning Research}.

\bibitem[{Gao et~al.(2024)Gao, Tow, Abbasi, Biderman, Black, DiPofi, Foster, Golding, Hsu, Le~Noac'h, Li, McDonell, Muennighoff, Ociepa, Phang, Reynolds, Schoelkopf, Skowron, Sutawika, Tang, Thite, Wang, Wang, and Zou}]{eval-harness}
Leo Gao, Jonathan Tow, Baber Abbasi, Stella Biderman, Sid Black, Anthony DiPofi, Charles Foster, Laurence Golding, Jeffrey Hsu, Alain Le~Noac'h, Haonan Li, Kyle McDonell, Niklas Muennighoff, Chris Ociepa, Jason Phang, Laria Reynolds, Hailey Schoelkopf, Aviya Skowron, Lintang Sutawika, and 5 others. 2024.
\newblock \href {https://doi.org/10.5281/zenodo.12608602} {The language model evaluation harness}.

\bibitem[{Gemini-Team et~al.(2025)Gemini-Team, Anil, Borgeaud, Alayrac, Yu, Soricut, Schalkwyk, Dai, Hauth, Millican, Silver, Johnson, Antonoglou, Schrittwieser, Glaese, Chen, Pitler, Lillicrap, Lazaridou, Firat, Molloy, Isard, Barham, Hennigan, Lee, Viola, Reynolds, Xu, Doherty, Collins, Meyer, Rutherford, Moreira, Ayoub, Goel, Krawczyk, Du, Chi, Cheng, Ni, Shah, Kane, Chan, Faruqui, Severyn, Lin, Li, Cheng, Ittycheriah, Mahdieh, Chen, Sun, Tran, Bagri, Lakshminarayanan, Liu, Orban, Güra, Zhou, Song, Boffy, Ganapathy, Zheng, Choe, Ágoston Weisz, Zhu, Lu, Gopal, Kahn, Kula, Pitman, Shah, Taropa, Merey, Baeuml, Chen, Shafey, Zhang, Sercinoglu, Tucker, Piqueras, Krikun, Barr, Savinov, Danihelka, Roelofs, White, Andreassen, von Glehn, Yagati, Kazemi, Gonzalez, Khalman, Sygnowski, Frechette, Smith, Culp, Proleev, Luan, Chen, Lottes, Schucher, Lebron, Rrustemi, Clay, Crone, Kocisky, Zhao, Perz, Yu, Howard, Bloniarz, Rae, Lu, Sifre, Maggioni, Alcober, Garrette, Barnes, Thakoor, Austin, Barth-Maron, Wong, Joshi,
  Chaabouni, Fatiha, Ahuja, Tomar, Senter, Chadwick, Kornakov, Attaluri, Iturrate, Liu, Li, Cogan, Chen, Jia, Gu, Zhang, Grimstad, Hartman, Garcia, Pillai, Devlin, Laskin, de~Las~Casas, Valter, Tao, Blanco, Badia, Reitter, Chen, Brennan, Rivera, Brin, Iqbal, Surita, Labanowski, Rao, Winkler, Parisotto, Gu, Olszewska, Addanki, Miech, Louis, Teplyashin, Brown, Catt, Balaguer, Xiang, Wang, Ashwood, Briukhov, Webson, Ganapathy, Sanghavi, Kannan, Chang, Stjerngren, Djolonga, Sun, Bapna, Aitchison, Pejman, Michalewski, Yu, Wang, Love, Ahn, Bloxwich, Han, Humphreys, Sellam, Bradbury, Godbole, Samangooei, Damoc, Kaskasoli, Arnold, Vasudevan, Agrawal, Riesa, Lepikhin, Tanburn, Srinivasan, Lim, Hodkinson, Shyam, Ferret, Hand, Garg, Paine, Li, Li, Giang, Neitz, Abbas, York, Reid, Cole, Chowdhery, Das, Rogozińska, Nikolaev, Sprechmann, Nado, Zilka, Prost, He, Monteiro, Mishra, Welty, Newlan, Jia, Allamanis, Hu, de~Liedekerke, Gilmer, Saroufim, Rijhwani, Hou, Shrivastava, Baddepudi, Goldin, Ozturel, Cassirer, Xu, Sohn,
  Sachan, Amplayo, Swanson, Petrova, Narayan, Guez, Brahma, Landon, Patel, Zhao, Villela, Wang, Jia, Rahtz, Giménez, Yeung, Keeling, Georgiev, Mincu, Wu, Haykal, Saputro, Vodrahalli, Qin, Cankara, Sharma, Fernando, Hawkins, Neyshabur, Kim, Hutter, Agrawal, Castro-Ros, van~den Driessche, Wang, Yang, yiin Chang, Komarek, McIlroy, Lučić, Zhang, Farhan, Sharman, Natsev, Michel, Bansal, Qiao, Cao, Shakeri, Butterfield, Chung, Rubenstein, Agrawal, Mensch, Soparkar, Lenc, Chung, Pope, Maggiore, Kay, Jhakra, Wang, Maynez, Phuong, Tobin, Tacchetti, Trebacz, Robinson, Katariya, Riedel, Bailey, Xiao, Ghelani, Aroyo, Slone, Houlsby, Xiong, Yang, Gribovskaya, Adler, Wirth, Lee, Li, Kagohara, Pavagadhi, Bridgers, Bortsova, Ghemawat, Ahmed, Liu, Powell, Bolina, Iinuma, Zablotskaia, Besley, Chung, Dozat, Comanescu, Si, Greer, Su, Polacek, Kaufman, Tokumine, Hu, Buchatskaya, Miao, Elhawaty, Siddhant, Tomasev, Xing, Greer, Miller, Ashraf, Roy, Zhang, Ma, Filos, Besta, Blevins, Klimenko, Yeh, Changpinyo, Mu, Chang,
  Pajarskas, Muir, Cohen, Lan, Haridasan, Marathe, Hansen, Douglas, Samuel, Wang, Austin, Lan, Jiang, Chiu, Lorenzo, Sjösund, Cevey, Gleicher, Avrahami, Boral, Srinivasan, Selo, May, Aisopos, Hussenot, Soares, Baumli, Chang, Recasens, Caine, Pritzel, Pavetic, Pardo, Gergely, Frye, Ramasesh, Horgan, Badola, Kassner, Roy, Dyer, Campos, Tomala, Tang, Badawy, White, Mustafa, Lang, Jindal, Vikram, Gong, Caelles, Hemsley, Thornton, Feng, Stokowiec, Zheng, Thacker, Çağlar Ünlü, Zhang, Saleh, Svensson, Bileschi, Patil, Anand, Ring, Tsihlas, Vezer, Selvi, Shevlane, Rodriguez, Kwiatkowski, Daruki, Rong, Dafoe, FitzGerald, Gu-Lemberg, Khan, Hendricks, Pellat, Feinberg, Cobon-Kerr, Sainath, Rauh, Hashemi, Ives, Hasson, Noland, Cao, Byrd, Hou, Wang, Sottiaux, Paganini, Lespiau, Moufarek, Hassan, Shivakumar, van Amersfoort, Mandhane, Joshi, Goyal, Tung, Brock, Sheahan, Misra, Li, Rakićević, Dehghani, Liu, Mittal, Oh, Noury, Sezener, Huot, Lamm, Cao, Chen, Mudgal, Stella, Brooks, Vasudevan, Liu, Chain, Melinkeri,
  Cohen, Wang, Seymore, Zubkov, Goel, Yue, Krishnakumaran, Albert, Hurley, Sano, Mohananey, Joughin, Filonov, Kępa, Eldawy, Lim, Rishi, Badiezadegan, Bos, Chang, Jain, Padmanabhan, Puttagunta, Krishna, Baker, Kalb, Bedapudi, Kurzrok, Lei, Yu, Litvin, Zhou, Wu, Sobell, Siciliano, Papir, Neale, Bragagnolo, Toor, Chen, Anklin, Wang, Feng, Gholami, Ling, Liu, Walter, Moghaddam, Kishore, Adamek, Mercado, Mallinson, Wandekar, Cagle, Ofek, Garrido, Lombriser, Mukha, Sun, Mohammad, Matak, Qian, Peswani, Janus, Yuan, Schelin, David, Garg, He, Duzhyi, Älgmyr, Lottaz, Li, Yadav, Xu, Chinien, Shivanna, Chuklin, Li, Spadine, Wolfe, Mohamed, Das, Dai, He, von Dincklage, Upadhyay, Maurya, Chi, Krause, Salama, Rabinovitch, M, Selvan, Dektiarev, Ghiasi, Guven, Gupta, Liu, Sharma, Shtacher, Paul, Akerlund, Aubet, Huang, Zhu, Zhu, Teixeira, Fritze, Bertolini, Marinescu, Bölle, Paulus, Gupta, Latkar, Chang, Sanders, Wilson, Wu, Tan, Thiet, Doshi, Lall, Mishra, Chen, Luong, Benjamin, Lee, Andrejczuk, Rabiej, Ranjan, Styrc,
  Yin, Simon, Harriott, Bansal, Robsky, Bacon, Greene, Mirylenka, Zhou, Sarvana, Goyal, Andermatt, Siegler, Horn, Israel, Pongetti, Chen, Selvatici, Silva, Wang, Tolins, Guu, Yogev, Cai, Agostini, Shah, Nguyen, Donnaile, Pereira, Friso, Stambler, Kurzrok, Kuang, Romanikhin, Geller, Yan, Jang, Lee, Fica, Malmi, Tan, Banica, Balle, Pham, Huang, Avram, Shi, Singh, Hidey, Ahuja, Saxena, Dooley, Potharaju, O'Neill, Gokulchandran, Foley, Zhao, Dusenberry, Liu, Mehta, Kotikalapudi, Safranek-Shrader, Goodman, Kessinger, Globen, Kolhar, Gorgolewski, Ibrahim, Song, Eichenbaum, Brovelli, Potluri, Lahoti, Baetu, Ghorbani, Chen, Crawford, Pal, Sridhar, Gurita, Mujika, Petrovski, Cedoz, Li, Chen, Santo, Goyal, Punjabi, Kappaganthu, Kwak, LV, Velury, Choudhury, Hall, Shah, Figueira, Thomas, Lu, Zhou, Kumar, Jurdi, Chikkerur, Ma, Yu, Kwak, Ähdel, Rajayogam, Choma, Liu, Barua, Ji, Park, Hellendoorn, Bailey, Bilal, Zhou, Khatir, Sutton, Rzadkowski, Macintosh, Vij, Shagin, Medina, Liang, Zhou, Shah, Bi, Dankovics, Banga,
  Lehmann, Bredesen, Lin, Hoffmann, Lai, Chung, Yang, Balani, Bražinskas, Sozanschi, Hayes, Alcalde, Makarov, Chen, Stella, Snijders, Mandl, Kärrman, Nowak, Wu, Dyck, Vaidyanathan, R, Mallet, Rudominer, Johnston, Mittal, Udathu, Christensen, Verma, Irving, Santucci, Elsayed, Davoodi, Georgiev, Tenney, Hua, Cideron, Leurent, Alnahlawi, Georgescu, Wei, Zheng, Scandinaro, Jiang, Snoek, Sundararajan, Wang, Ontiveros, Karo, Cole, Rajashekhar, Tumeh, Ben-David, Jain, Uesato, Datta, Bunyan, Wu, Zhang, Stanczyk, Zhang, Steiner, Naskar, Azzam, Johnson, Paszke, Chiu, Elias, Mohiuddin, Muhammad, Miao, Lee, Vieillard, Park, Zhang, Stanway, Garmon, Karmarkar, Dong, Lee, Kumar, Zhou, Evens, Isaac, Irving, Loper, Fink, Arkatkar, Chen, Shafran, Petrychenko, Chen, Jia, Levskaya, Zhu, Grabowski, Mao, Magni, Yao, Snaider, Casagrande, Palmer, Suganthan, Castaño, Giannoumis, Kim, Rybiński, Sreevatsa, Prendki, Soergel, Goedeckemeyer, Gierke, Jafari, Gaba, Wiesner, Wright, Wei, Vashisht, Kulizhskaya, Hoover, Le, Li, Iwuanyanwu,
  Liu, Ramirez, Khorlin, Cui, LIN, Wu, Aguilar, Pallo, Chakladar, Perng, Abellan, Zhang, Dasgupta, Kushman, Penchev, Repina, Wu, van~der Weide, Ponnapalli, Kaplan, Simsa, Li, Dousse, Yang, Piper, Ie, Pasumarthi, Lintz, Vijayakumar, Andor, Valenzuela, Lui, Paduraru, Peng, Lee, Zhang, Greene, Nguyen, Kurylowicz, Hardin, Dixon, Janzer, Choo, Feng, Zhang, Singhal, Du, McKinnon, Antropova, Bolukbasi, Keller, Reid, Finchelstein, Raad, Crocker, Hawkins, Dadashi, Gaffney, Franko, Bulanova, Leblond, Chung, Askham, Cobo, Xu, Fischer, Xu, Sorokin, Alberti, Lin, Evans, Dimitriev, Forbes, Banarse, Tung, Omernick, Bishop, Sterneck, Jain, Xia, Amid, Piccinno, Wang, Banzal, Mankowitz, Polozov, Krakovna, Brown, Bateni, Duan, Firoiu, Thotakuri, Natan, Geist, tan Girgin, Li, Ye, Roval, Tojo, Kwong, Lee-Thorp, Yew, Sinopalnikov, Ramos, Mellor, Sharma, Wu, Miller, Sonnerat, Vnukov, Greig, Beattie, Caveness, Bai, Eisenschlos, Korchemniy, Tsai, Jasarevic, Kong, Dao, Zheng, Liu, Yang, Zhu, Teh, Sanmiya, Gladchenko, Trdin, Toyama,
  Rosen, Tavakkol, Xue, Elkind, Woodman, Carpenter, Papamakarios, Kemp, Kafle, Grunina, Sinha, Talbert, Wu, Owusu-Afriyie, Du, Thornton, Pont-Tuset, Narayana, Li, Fatehi, Wieting, Ajmeri, Uria, Ko, Knight, Héliou, Niu, Gu, Pang, Li, Levine, Stolovich, Santamaria-Fernandez, Goenka, Yustalim, Strudel, Elqursh, Deck, Lee, Li, Levin, Hoffmann, Holtmann-Rice, Bachem, Arora, Koh, Yeganeh, Põder, Tariq, Sun, Ionita, Seyedhosseini, Tafti, Liu, Gulati, Liu, Ye, Chrzaszcz, Wang, Sethi, Li, Brown, Singh, Fan, Parisi, Stanton, Koverkathu, Choquette-Choo, Li, Lu, Ittycheriah, Shroff, Varadarajan, Bahargam, Willoughby, Gaddy, Desjardins, Cornero, Robenek, Mittal, Albrecht, Shenoy, Moiseev, Jacobsson, Ghaffarkhah, Rivière, Walton, Crepy, Parrish, Zhou, Farabet, Radebaugh, Srinivasan, van~der Salm, Fidjeland, Scellato, Latorre-Chimoto, Klimczak-Plucińska, Bridson, de~Cesare, Hudson, Mendolicchio, Walker, Morris, Mauger, Guseynov, Reid, Odoom, Loher, Cotruta, Yenugula, Grewe, Petrushkina, Duerig, Sanchez, Yadlowsky, Shen,
  Globerson, Webb, Dua, Li, Bhupatiraju, Hurt, Qureshi, Agarwal, Shani, Eyal, Khare, Belle, Wang, Tekur, Kale, Wei, Sang, Saeta, Liechty, Sun, Zhao, Lee, Nayak, Fritz, Vuyyuru, Aslanides, Vyas, Wicke, Ma, Eltyshev, Martin, Cate, Manyika, Amiri, Kim, Xiong, Kang, Luisier, Tripuraneni, Madras, Guo, Waters, Wang, Ainslie, Baldridge, Zhang, Pruthi, Bauer, Yang, Mansour, Gelman, Xu, Polovets, Liu, Cai, Chen, Sheng, Xue, Ozair, Angermueller, Li, Sinha, Wang, Wiesinger, Koukoumidis, Tian, Iyer, Gurumurthy, Goldenson, Shah, Blake, Yu, Urbanowicz, Palomaki, Fernando, Durden, Mehta, Momchev, Rahimtoroghi, Georgaki, Raul, Ruder, Redshaw, Lee, Zhou, Jalan, Li, Hechtman, Schuh, Nasr, Milan, Mikulik, Franco, Green, Nguyen, Kelley, Mahendru, Hu, Howland, Vargas, Hui, Bansal, Rao, Ghiya, Wang, Ye, Sarr, Preston, Elish, Li, Kaku, Gupta, Pasupat, Juan, Someswar, M., Chen, Amini, Fabrikant, Chu, Dong, Muthal, Buthpitiya, Jauhari, Hua, Khandelwal, Hitron, Ren, Rinaldi, Drath, Dabush, Jiang, Godhia, Sachs, Chen, Fan, Taitelbaum,
  Noga, Dai, Wang, Liang, Hamer, Ferng, Elkind, Atias, Lee, Listík, Carlen, van~de Kerkhof, Pikus, Zaher, Müller, Zykova, Stefanec, Gatsko, Hirnschall, Sethi, Xu, Ahuja, Tsai, Stefanoiu, Feng, Dhandhania, Katyal, Gupta, Parulekar, Pitta, Zhao, Bhatia, Bhavnani, Alhadlaq, Li, Danenberg, Tu, Pine, Filippova, Ghosh, Limonchik, Urala, Lanka, Clive, Sun, Li, Wu, Hongtongsak, Li, Thakkar, Omarov, Majmundar, Alverson, Kucharski, Patel, Jain, Zabelin, Pelagatti, Kohli, Kumar, Kim, Sankar, Shah, Ramachandruni, Zeng, Bariach, Weidinger, Vu, Andreev, He, Hui, Kashem, Subramanya, Hsiao, Hassabis, Kavukcuoglu, Sadovsky, Le, Strohman, Wu, Petrov, Dean, and Vinyals}]{geminiteam2025geminifamilyhighlycapable}
Gemini-Team, Rohan Anil, Sebastian Borgeaud, Jean-Baptiste Alayrac, Jiahui Yu, Radu Soricut, Johan Schalkwyk, Andrew~M. Dai, Anja Hauth, Katie Millican, David Silver, Melvin Johnson, Ioannis Antonoglou, Julian Schrittwieser, Amelia Glaese, Jilin Chen, Emily Pitler, Timothy Lillicrap, Angeliki Lazaridou, and 1332 others. 2025.
\newblock \href {https://arxiv.org/abs/2312.11805} {Gemini: A family of highly capable multimodal models}.
\newblock \emph{Preprint}, arXiv:2312.11805.

\bibitem[{Gunasekar et~al.(2023)Gunasekar, Zhang, Aneja, Mendes, Giorno, Gopi, Javaheripi, Kauffmann, de~Rosa, Saarikivi, Salim, Shah, Behl, Wang, Bubeck, Eldan, Kalai, Lee, and Li}]{gunasekar2023textbooksneed}
Suriya Gunasekar, Yi~Zhang, Jyoti Aneja, Caio César~Teodoro Mendes, Allie~Del Giorno, Sivakanth Gopi, Mojan Javaheripi, Piero Kauffmann, Gustavo de~Rosa, Olli Saarikivi, Adil Salim, Shital Shah, Harkirat~Singh Behl, Xin Wang, Sébastien Bubeck, Ronen Eldan, Adam~Tauman Kalai, Yin~Tat Lee, and Yuanzhi Li. 2023.
\newblock \href {https://arxiv.org/abs/2306.11644} {Textbooks are all you need}.
\newblock \emph{Preprint}, arXiv:2306.11644.

\bibitem[{Havrilla and Iyer(2024)}]{havrilla2024understandingeffectnoisellm}
Alex Havrilla and Maia Iyer. 2024.
\newblock \href {https://arxiv.org/abs/2402.04004} {Understanding the effect of noise in llm training data with algorithmic chains of thought}.
\newblock \emph{Preprint}, arXiv:2402.04004.

\bibitem[{Hendrycks et~al.(2021)Hendrycks, Burns, Kadavath, Arora, Basart, Tang, Song, and Steinhardt}]{hendrycks2021measuringmathematicalproblemsolving}
Dan Hendrycks, Collin Burns, Saurav Kadavath, Akul Arora, Steven Basart, Eric Tang, Dawn Song, and Jacob Steinhardt. 2021.
\newblock \href {https://arxiv.org/abs/2103.03874} {Measuring mathematical problem solving with the math dataset}.
\newblock \emph{Preprint}, arXiv:2103.03874.

\bibitem[{Jiang et~al.(2025)Jiang, Li, Song, Zhang, Zhu, Zhao, Xu, Taura, and Wang}]{jiang2025importanceawaredataselectionefficient}
Tingyu Jiang, Shen Li, Yiyao Song, Lan Zhang, Hualei Zhu, Yuan Zhao, Xiaohang Xu, Kenjiro Taura, and Hao~Henry Wang. 2025.
\newblock \href {https://arxiv.org/abs/2511.07074} {Importance-aware data selection for efficient llm instruction tuning}.
\newblock \emph{Preprint}, arXiv:2511.07074.

\bibitem[{Kornblith et~al.(2019)Kornblith, Norouzi, Lee, and Hinton}]{pmlr-v97-kornblith19a}
Simon Kornblith, Mohammad Norouzi, Honglak Lee, and Geoffrey Hinton. 2019.
\newblock \href {https://proceedings.mlr.press/v97/kornblith19a.html} {Similarity of neural network representations revisited}.
\newblock In \emph{Proceedings of the 36th International Conference on Machine Learning}, volume~97 of \emph{Proceedings of Machine Learning Research}, pages 3519--3529. PMLR.

\bibitem[{Lai et~al.(2017)Lai, Xie, Liu, Yang, and Hovy}]{lai2017racelargescalereadingcomprehension}
Guokun Lai, Qizhe Xie, Hanxiao Liu, Yiming Yang, and Eduard Hovy. 2017.
\newblock \href {https://arxiv.org/abs/1704.04683} {Race: Large-scale reading comprehension dataset from examinations}.
\newblock \emph{Preprint}, arXiv:1704.04683.

\bibitem[{Li et~al.(2024)Li, Zhang, He, Li, Zhao, Wang, Cheng, and Zhou}]{li-etal-2024-superfiltering}
Ming Li, Yong Zhang, Shwai He, Zhitao Li, Hongyu Zhao, Jianzong Wang, Ning Cheng, and Tianyi Zhou. 2024.
\newblock \href {https://doi.org/10.18653/v1/2024.acl-long.769} {Superfiltering: Weak-to-strong data filtering for fast instruction-tuning}.
\newblock In \emph{Proceedings of the 62nd Annual Meeting of the Association for Computational Linguistics (Volume 1: Long Papers)}, pages 14255--14273, Bangkok, Thailand. Association for Computational Linguistics.

\bibitem[{Li et~al.(2025)Li, Hua, Vu, Zhan, Qu, and Haffari}]{li-etal-2025-scar}
Zhuang Li, Yuncheng Hua, Thuy-Trang Vu, Haolan Zhan, Lizhen Qu, and Gholamreza Haffari. 2025.
\newblock \href {https://aclanthology.org/2025.acl-long.625/} {{SCAR}: Data selection via style consistency-aware response ranking for efficient instruction-tuning of large language models}.
\newblock In \emph{Proceedings of the 63rd Annual Meeting of the Association for Computational Linguistics (Volume 1: Long Papers)}, pages 12756--12790, Vienna, Austria. Association for Computational Linguistics.

\bibitem[{Lin et~al.(2023)Lin, Ravichander, Lu, Dziri, Sclar, Chandu, Bhagavatula, and Choi}]{lin2023unlockingspellbasellms}
Bill~Yuchen Lin, Abhilasha Ravichander, Ximing Lu, Nouha Dziri, Melanie Sclar, Khyathi Chandu, Chandra Bhagavatula, and Yejin Choi. 2023.
\newblock \href {https://arxiv.org/abs/2312.01552} {The unlocking spell on base llms: Rethinking alignment via in-context learning}.
\newblock \emph{Preprint}, arXiv:2312.01552.

\bibitem[{Llama~Team(2024)}]{grattafiori2024llama3herdmodels}
Meta Llama~Team. 2024.
\newblock \href {https://arxiv.org/abs/2407.21783} {The llama 3 herd of models}.
\newblock \emph{Preprint}, arXiv:2407.21783.

\bibitem[{McCarthy and Jarvis(2010)}]{mccarthy2010mtld}
Philip~M McCarthy and Scott Jarvis. 2010.
\newblock \href {https://doi.org/10.3758/BRM.42.2.381} {Mtld, vocd-d, and hd-d: A validation study of several new measures of lexical diversity}.
\newblock \emph{Behavior Research Methods}, 42(2):381--392.

\bibitem[{Mihaylov et~al.(2018)Mihaylov, Clark, Khot, and Sabharwal}]{mihaylov2018suitarmorconductelectricity}
Todor Mihaylov, Peter Clark, Tushar Khot, and Ashish Sabharwal. 2018.
\newblock \href {https://arxiv.org/abs/1809.02789} {Can a suit of armor conduct electricity? a new dataset for open book question answering}.
\newblock \emph{Preprint}, arXiv:1809.02789.

\bibitem[{OpenAI(2024)}]{openai2024gpt4technicalreport}
OpenAI. 2024.
\newblock \href {https://arxiv.org/abs/2303.08774} {Gpt-4 technical report}.
\newblock \emph{Preprint}, arXiv:2303.08774.

\bibitem[{Qwen-Team(2024)}]{qwen2.5}
Qwen-Team. 2024.
\newblock \href {https://qwenlm.github.io/blog/qwen2.5/} {Qwen2.5: A party of foundation models}.

\bibitem[{Raghavendra et~al.(2025)Raghavendra, Nath, and Hendryx}]{raghavendra2025revisiting}
Mohit Raghavendra, Vaskar Nath, and Sean~M. Hendryx. 2025.
\newblock \href {https://openreview.net/forum?id=SIzjhS9kEF} {Revisiting the superficial alignment hypothesis}.

\bibitem[{Rajpurkar et~al.(2018)Rajpurkar, Jia, and Liang}]{rajpurkar2018knowdontknowunanswerable}
Pranav Rajpurkar, Robin Jia, and Percy Liang. 2018.
\newblock \href {https://arxiv.org/abs/1806.03822} {Know what you don't know: Unanswerable questions for squad}.
\newblock \emph{Preprint}, arXiv:1806.03822.

\bibitem[{Sakaguchi et~al.(2019)Sakaguchi, Bras, Bhagavatula, and Choi}]{sakaguchi2019winograndeadversarialwinogradschema}
Keisuke Sakaguchi, Ronan~Le Bras, Chandra Bhagavatula, and Yejin Choi. 2019.
\newblock \href {https://arxiv.org/abs/1907.10641} {Winogrande: An adversarial winograd schema challenge at scale}.
\newblock \emph{Preprint}, arXiv:1907.10641.

\bibitem[{Sap et~al.(2019)Sap, Rashkin, Chen, LeBras, and Choi}]{sap2019socialiqacommonsensereasoningsocial}
Maarten Sap, Hannah Rashkin, Derek Chen, Ronan LeBras, and Yejin Choi. 2019.
\newblock \href {https://arxiv.org/abs/1904.09728} {Socialiqa: Commonsense reasoning about social interactions}.
\newblock \emph{Preprint}, arXiv:1904.09728.

\bibitem[{Schlichtkrull et~al.(2018)Schlichtkrull, Kipf, Bloem, Berg, Titov, and Welling}]{fb15k}
Michael Schlichtkrull, Thomas~N Kipf, Peter Bloem, Rianne van~den Berg, Ivan Titov, and Max Welling. 2018.
\newblock Modeling relational data with graph convolutional networks.
\newblock In \emph{European semantic web conference}, pages 593--607. Springer.

\bibitem[{Sener and Savarese(2018)}]{sener2018active}
Ozan Sener and Silvio Savarese. 2018.
\newblock \href {https://openreview.net/forum?id=H1aIuk-RW} {Active learning for convolutional neural networks: A core-set approach}.
\newblock In \emph{International Conference on Learning Representations}.

\bibitem[{Taori et~al.(2023)Taori, Gulrajani, Zhang, Dubois, Li, Guestrin, Liang, and Hashimoto}]{alpaca}
Rohan Taori, Ishaan Gulrajani, Tianyi Zhang, Yann Dubois, Xuechen Li, Carlos Guestrin, Percy Liang, and Tatsunori~B. Hashimoto. 2023.
\newblock Stanford alpaca: An instruction-following llama model.
\newblock \url{https://github.com/tatsu-lab/stanford_alpaca}.

\bibitem[{Tran et~al.(2022)Tran, Ouchi, Watanabe, and Matsumoto}]{zsre}
Van-Hien Tran, Hiroki Ouchi, Taro Watanabe, and Yuji Matsumoto. 2022.
\newblock \href {https://doi.org/10.18653/v1/2022.spanlp-1.1} {Improving discriminative learning for zero-shot relation extraction}.
\newblock In \emph{Proceedings of the 1st Workshop on Semiparametric Methods in NLP: Decoupling Logic from Knowledge}, pages 1--6, Dublin, Ireland and Online. Association for Computational Linguistics.

\bibitem[{Wang et~al.(2024)Wang, Zhang, Tian, Xi, Yao, Xu, Wang, Mao, Wang, Cheng, Liu, Ni, Zheng, and Chen}]{wang2023easyedit}
Peng Wang, Ningyu Zhang, Bozhong Tian, Zekun Xi, Yunzhi Yao, Ziwen Xu, Mengru Wang, Shengyu Mao, Xiaohan Wang, Siyuan Cheng, Kangwei Liu, Yuansheng Ni, Guozhou Zheng, and Huajun Chen. 2024.
\newblock \href {https://arxiv.org/abs/2308.07269} {Easyedit: An easy-to-use knowledge editing framework for large language models}.
\newblock \emph{Preprint}, arXiv:2308.07269.

\bibitem[{Wang et~al.(2025)Wang, Li, Zhang, Wu, Liu, Hu, Guo, Huang, Xin, Yang, Su, Chen, and Li}]{wang2025epicoder}
Yaoxiang Wang, Haoling Li, Xin Zhang, Jie Wu, Xiao Liu, Wenxiang Hu, Zhongxin Guo, Yangyu Huang, Ying Xin, Yujiu Yang, Jinsong Su, Qi~Chen, and Scarlett Li. 2025.
\newblock \href {https://openreview.net/forum?id=RAxe7nF4Oz} {Epicoder: Encompassing diversity and complexity in code generation}.
\newblock In \emph{Forty-second International Conference on Machine Learning}.

\bibitem[{Wang et~al.(2023)Wang, Kordi, Mishra, Liu, Smith, Khashabi, and Hajishirzi}]{wang2023selfinstructaligninglanguagemodels}
Yizhong Wang, Yeganeh Kordi, Swaroop Mishra, Alisa Liu, Noah~A. Smith, Daniel Khashabi, and Hannaneh Hajishirzi. 2023.
\newblock \href {https://arxiv.org/abs/2212.10560} {Self-instruct: Aligning language models with self-generated instructions}.
\newblock \emph{Preprint}, arXiv:2212.10560.

\bibitem[{Xie et~al.(2024)Xie, Aggarwal, Ahmad, and Lau}]{xie2024controlledautomatictaskspecificsynthetic}
Yong Xie, Karan Aggarwal, Aitzaz Ahmad, and Stephen Lau. 2024.
\newblock \href {https://arxiv.org/abs/2410.12278} {Controlled automatic task-specific synthetic data generation for hallucination detection}.
\newblock \emph{Preprint}, arXiv:2410.12278.

\bibitem[{Xu et~al.(2025)Xu, Sun, Zheng, Geng, Zhao, Feng, Tao, Lin, and Jiang}]{xu2025wizardlmempoweringlargepretrained}
Can Xu, Qingfeng Sun, Kai Zheng, Xiubo Geng, Pu~Zhao, Jiazhan Feng, Chongyang Tao, Qingwei Lin, and Daxin Jiang. 2025.
\newblock \href {https://arxiv.org/abs/2304.12244} {Wizardlm: Empowering large pre-trained language models to follow complex instructions}.
\newblock \emph{Preprint}, arXiv:2304.12244.

\bibitem[{Xu et~al.(2024)Xu, Jiang, Niu, Deng, Poovendran, Choi, and Lin}]{xu2024magpiealignmentdatasynthesis}
Zhangchen Xu, Fengqing Jiang, Luyao Niu, Yuntian Deng, Radha Poovendran, Yejin Choi, and Bill~Yuchen Lin. 2024.
\newblock \href {https://arxiv.org/abs/2406.08464} {Magpie: Alignment data synthesis from scratch by prompting aligned llms with nothing}.
\newblock \emph{Preprint}, arXiv:2406.08464.

\bibitem[{Yang et~al.(2025{\natexlab{a}})Yang, Li, Yang, Zhang, Hui, Zheng, Yu, Gao, Huang, Lv, Zheng, Liu, Zhou, Huang, Hu, Ge, Wei, Lin, Tang, Yang, Tu, Zhang, Yang, Yang, Zhou, Zhou, Lin, Dang, Bao, Yang, Yu, Deng, Li, Xue, Li, Zhang, Wang, Zhu, Men, Gao, Liu, Luo, Li, Tang, Yin, Ren, Wang, Zhang, Ren, Fan, Su, Zhang, Zhang, Wan, Liu, Wang, Cui, Zhang, Zhou, and Qiu}]{qwen3}
An~Yang, Anfeng Li, Baosong Yang, Beichen Zhang, Binyuan Hui, Bo~Zheng, Bowen Yu, Chang Gao, Chengen Huang, Chenxu Lv, Chujie Zheng, Dayiheng Liu, Fan Zhou, Fei Huang, Feng Hu, Hao Ge, Haoran Wei, Huan Lin, Jialong Tang, and 41 others. 2025{\natexlab{a}}.
\newblock Qwen3 technical report.
\newblock \emph{arXiv preprint arXiv:2505.09388}.

\bibitem[{Yang et~al.(2025{\natexlab{b}})Yang, Nan, Ye, Dou, Wang, Li, Lv, Gui, Zhang, and Huang}]{yang-etal-2025-measuring}
Yuming Yang, Yang Nan, Junjie Ye, Shihan Dou, Xiao Wang, Shuo Li, Huijie Lv, Tao Gui, Qi~Zhang, and Xuanjing Huang. 2025{\natexlab{b}}.
\newblock \href {https://doi.org/10.18653/v1/2025.acl-long.908} {Measuring data diversity for instruction tuning: A systematic analysis and a reliable metric}.
\newblock In \emph{Proceedings of the 63rd Annual Meeting of the Association for Computational Linguistics (Volume 1: Long Papers)}, pages 18530--18549, Vienna, Austria. Association for Computational Linguistics.

\bibitem[{Yu et~al.(2023)Yu, Jiang, Shi, Yu, Liu, Zhang, Kwok, Li, Weller, and Liu}]{yu2023metamath}
Longhui Yu, Weisen Jiang, Han Shi, Jincheng Yu, Zhengying Liu, Yu~Zhang, James~T Kwok, Zhenguo Li, Adrian Weller, and Weiyang Liu. 2023.
\newblock Metamath: Bootstrap your own mathematical questions for large language models.
\newblock \emph{arXiv preprint arXiv:2309.12284}.

\bibitem[{Zhang et~al.(2026)Zhang, Yao, Qin, Xu, Zhu, Yu, Wang, Tang, Gu, Deng, and Chen}]{Zhang_2026}
Ningyu Zhang, Yunzhi Yao, Jiaxin Qin, Haoming Xu, Yuqi Zhu, Zeping Yu, Mengru Wang, Yuqi Tang, Jia-Chen Gu, Shumin Deng, and Huajun Chen. 2026.
\newblock \href {https://doi.org/10.1038/s42256-026-01276-y} {Towards principled knowledge editing methods for large language model reasoning}.
\newblock \emph{Nature Machine Intelligence}, 8(8):1189–1200.

\bibitem[{Zhao et~al.(2024)Zhao, Andriushchenko, Croce, and Flammarion}]{zhao2024long}
Hao Zhao, Maksym Andriushchenko, Francesco Croce, and Nicolas Flammarion. 2024.
\newblock \href {https://openreview.net/forum?id=0AZAjkXhit} {Long is more for alignment: A simple but tough-to-beat baseline for instruction fine-tuning}.
\newblock In \emph{Forty-first International Conference on Machine Learning}.

\bibitem[{Zheng et~al.(2024{\natexlab{a}})Zheng, Zhang, Zhang, Ye, and Luo}]{zheng-etal-2024-llamafactory}
Yaowei Zheng, Richong Zhang, Junhao Zhang, Yanhan Ye, and Zheyan Luo. 2024{\natexlab{a}}.
\newblock \href {https://doi.org/10.18653/v1/2024.acl-demos.38} {{L}lama{F}actory: Unified efficient fine-tuning of 100+ language models}.
\newblock In \emph{Proceedings of the 62nd Annual Meeting of the Association for Computational Linguistics (Volume 3: System Demonstrations)}, pages 400--410, Bangkok, Thailand. Association for Computational Linguistics.

\bibitem[{Zheng et~al.(2024{\natexlab{b}})Zheng, Zhang, Zhang, Ye, Luo, Feng, and Ma}]{zheng2024llamafactory}
Yaowei Zheng, Richong Zhang, Junhao Zhang, Yanhan Ye, Zheyan Luo, Zhangchi Feng, and Yongqiang Ma. 2024{\natexlab{b}}.
\newblock \href {http://arxiv.org/abs/2403.13372} {Llamafactory: Unified efficient fine-tuning of 100+ language models}.
\newblock In \emph{Proceedings of the 62nd Annual Meeting of the Association for Computational Linguistics (Volume 3: System Demonstrations)}, Bangkok, Thailand. Association for Computational Linguistics.

\bibitem[{Zhou et~al.(2023)Zhou, Liu, Xu, Iyer, Sun, Mao, Ma, Efrat, Yu, Yu, Zhang, Ghosh, Lewis, Zettlemoyer, and Levy}]{zhou2023limaalignment}
Chunting Zhou, Pengfei Liu, Puxin Xu, Srini Iyer, Jiao Sun, Yuning Mao, Xuezhe Ma, Avia Efrat, Ping Yu, Lili Yu, Susan Zhang, Gargi Ghosh, Mike Lewis, Luke Zettlemoyer, and Omer Levy. 2023.
\newblock \href {https://arxiv.org/abs/2305.11206} {Lima: Less is more for alignment}.
\newblock \emph{Preprint}, arXiv:2305.11206.

\end{thebibliography}

\appendix
\section{Expanded Related Work}
\label{sec:exrelated}
\paragraph{Construction and Filtering of High-Quality Data for SFT}
Early work highlights the importance of training data quality over sheer data scale in SFT.
LIMA \citep{zhou2023limaalignment} and Textbooks Are All You Need \citep{gunasekar2023textbooksneed}
demonstrate that effective alignment does not strictly require large-scale datasets, but can emerge from relatively small yet high-quality training corpora.
In a similar spirit, generate-then-filter pipelines such as Self-Instruct \citep{wang2023selfinstructaligninglanguagemodels}, WizardLM \citep{xu2025wizardlmempoweringlargepretrained}, and Magpie \citep{xu2024magpiealignmentdatasynthesis} explore scalable, high-quality data construction through one or more rounds of data generation and heuristic filtering. More recently, Condor \citep{maosongcao-etal-2025-condor} refines synthetic training data during generation by constructing instructions of varying difficulty and improving response quality.

Other work in this line focuses on filtering and selecting representative high-quality training data. Many approaches rely on surface-level quality metrics or heuristic-based strategies, including length-based filtering \citep{zhao2024long}, perplexity-based selection \citep{li-etal-2024-superfiltering}, and style-consistency constraints \citep{li-etal-2025-scar}. More recent work has proposed methods that estimate sample contributions to downstream performance, such as BIDS \citep{dai2025improvinginfluencebasedinstructiontuning}, MIG \citep{chen2025migautomaticdataselection}, and MIWV \citep{jiang2025importanceawaredataselectionefficient}.

However, most data synthesis, refinement, and selection methods lack a guiding framework for quantitatively defining data quality. Consequently, they oscillate between two extremes: overly aggressive filtering that removes valuable supervisory signals, and overly permissive filtering that retains training instances with potential errors.

\paragraph{Training Data Noise Effects and Learning Mechanisms in SFT}
Prior work has investigated noisy supervision in SFT under a variety of controlled settings. 
FTNI \citep{alajrami2025finetuningnoisyinstructionseffects} observes that training on perturbed instructions (such as removing stop words or shuffling words) can improve downstream performance in some cases. For reasoning supervision, TInt \citep{havrilla2024understandingeffectnoisellm} studies robustness to noise in algorithmic chains of thought by injecting errors during reasoning, and FACO \citep{cho2024unveilingimitationlearningexploring} studies how different proportions of erroneous reasoning-chain samples in the training data affect model performance.

From a learning-mechanism perspective, the superficial alignment hypothesis characterizes SFT as primarily focusing on adopting the language style of responsible AI assistants and relying largely on the knowledge already acquired by base LLMs \citep{zhou2023limaalignment,lin2023unlockingspellbasellms,cao2024mindtonguesdeepdive}. Beyond this view, subsequent evidence shows that SFT enhances models’ reasoning and contextual understanding compared to their base counterparts \citep{chen2025extractingunderstandingsuperficialknowledge}. \citet{raghavendra2025revisiting} further observe that, similar to the pre-training scaling laws, post-training task performance scales as a power law against the number of finetuning examples, reflecting that language models are not necessarily confined to using only the knowledge learned during pretraining.

Current analyses of noise in training data and SFT learning mechanisms remain confined to specific types of datasets or lack fine-grained, sample-level structural analysis. As a result, they are unable to determine which errors are tolerable to the model and which genuinely disrupt learning, leaving the critical boundary between benign and detrimental noise unclear and the quantification of error tolerance in synthetic data unexplored.

\section{Pipeline for ATOM}
This section presents the detailed process of ATOM. We elaborate from the perspectives of the Pipeline for Atomic Tree Construction and Atomic Data Synthesis, respectively.

\subsection{Pipeline for Atomic Tree Construction}
\label{sec:appendix-ConstructionPrompt}
To facilitate clarity, we first detail the prompts employed in our pipeline, followed by a formal description using pseudo-code.

\begin{figure*}[!ht]
    \centering
    \includegraphics[
        width=1\textwidth,
    ]{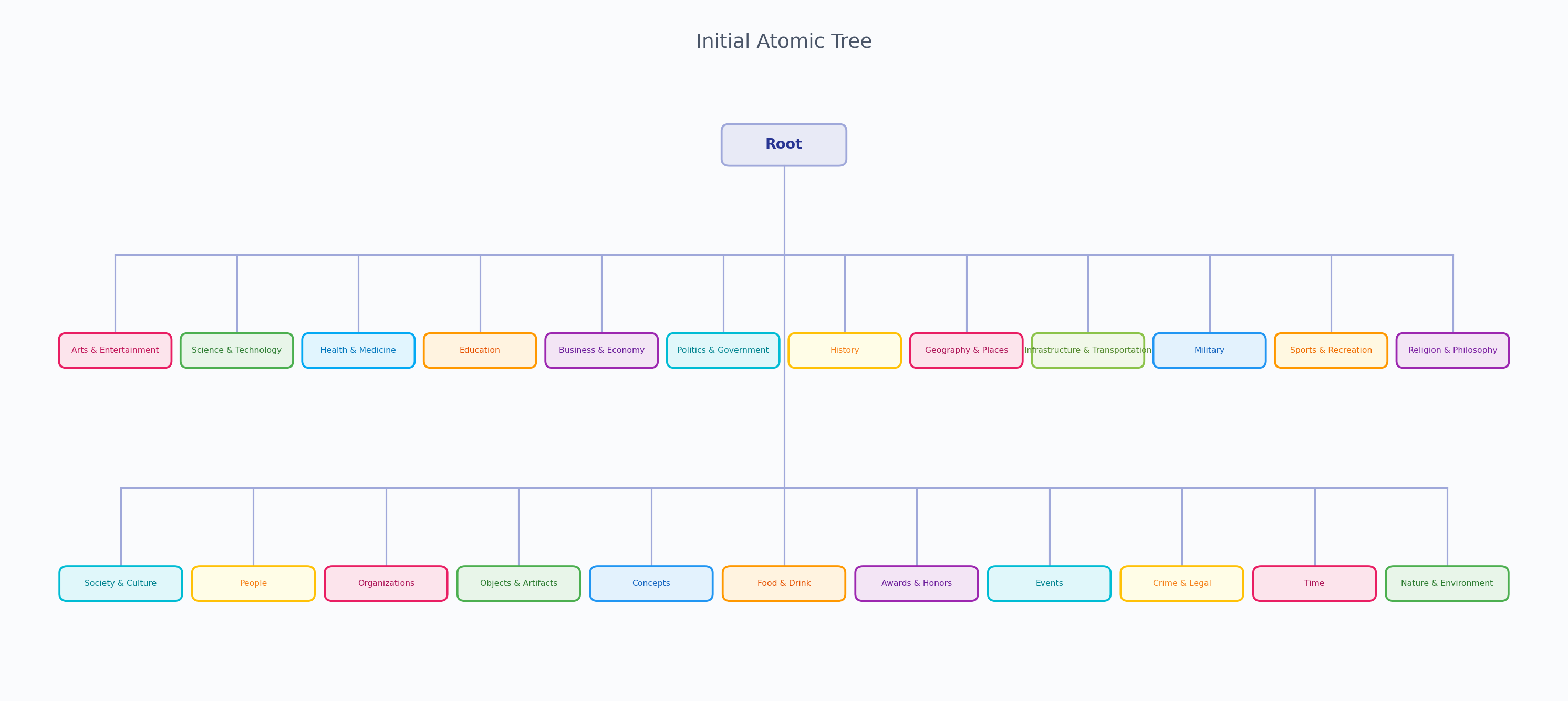}
    \caption{The initial operator tree of Atomic-QA. Internal nodes represent abstract operator categories, and leaf nodes define concrete atomic operators.}
    \label{fig:Atomic-QA_Tree}
\end{figure*}

\begin{figure*}[!ht]
    \centering
    \includegraphics[
        width=1\textwidth,
    ]{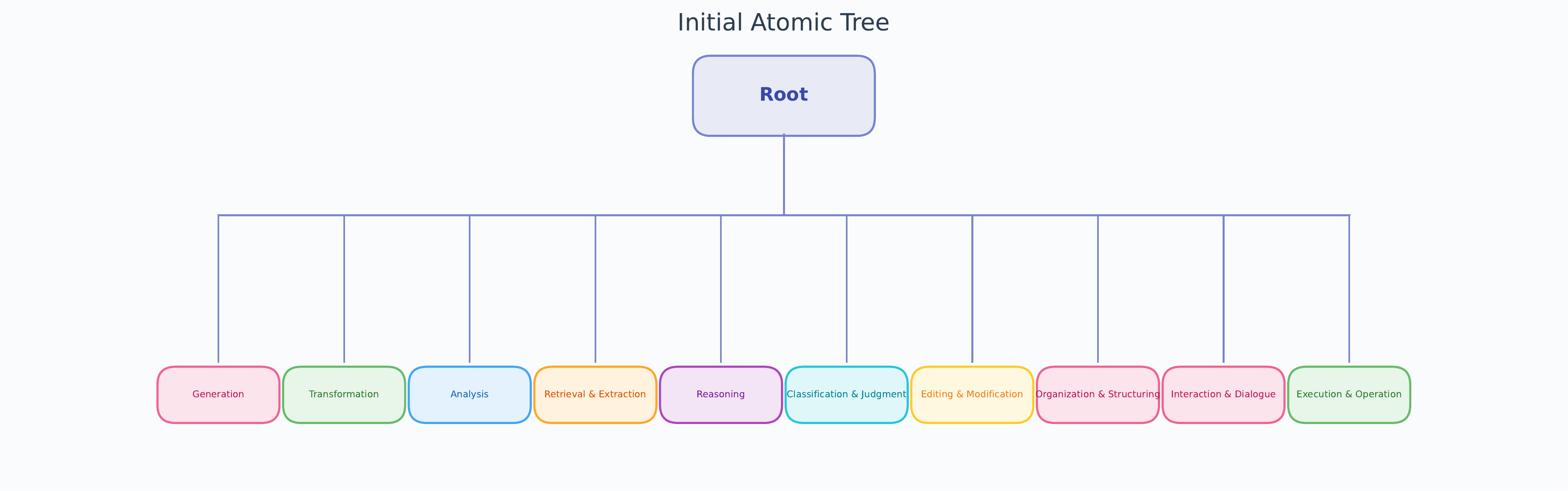}
    \caption{The initial operator tree of Alpaca.}
    \label{fig:Alpaca_Tree}
\end{figure*}

\begin{figure*}[!ht]
    \centering
    \includegraphics[
        width=1\textwidth,
    ]{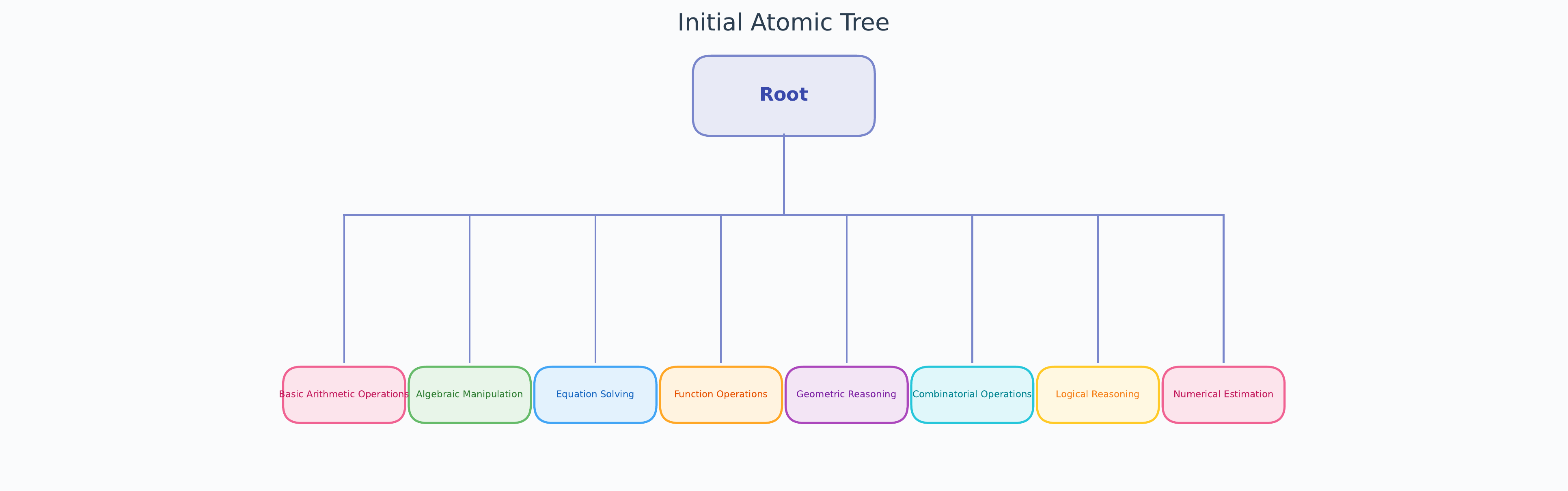}
    \caption{The initial operator tree of MetaMath.}
    \label{fig:MetaMath_Tree}
\end{figure*}

ATOM initiates the process by extracting atomic operators from representative data, guided by predefined definitions via Prompt \ref{Atomic-Function-Extraction}. Subsequently, we organize these operators into an Initial Operator Tree using Prompt \ref{Initial-Tree-Generation}, where internal nodes represent abstract operator categories and leaf nodes represent concrete atomic operators. To guarantee structural completeness, we generate trees for various subsets of atomic functions under diverse sampling schemes, while simultaneously pruning duplicate nodes at the same hierarchy level via Prompt \ref{Tree-Duplication}. After the above steps, we obtain the operator trees shown in Figure \ref{fig:Atomic-QA_Tree}, \ref{fig:Alpaca_Tree} and \ref{fig:MetaMath_Tree}. Finally, building upon this initial structure and its enumerated first-layer nodes, we expand the tree across both depth and breadth dimensions using Prompt \ref{Tree-Expansion}. The comprehensive procedure for Atomic Tree Construction is outlined in Algorithm \ref{alg:atomic_tree}.

\paragraph{Human Expert Verification.}
To ensure the quality and validity of the constructed atomic trees, we involve four human experts throughout the pipeline. During \textbf{Initial Tree Generation} (Phase 2), the experts evaluate whether the category hierarchy is \textit{reasonable} (i.e., semantically coherent groupings) and \textit{comprehensive} (i.e., sufficient coverage of the operator space). An initial tree is accepted only when all four experts reach unanimous agreement; otherwise, the tree is revised and re-evaluated. During \textbf{Tree Expansion} (Phase 3), the experts verify whether the newly generated leaf nodes conform to the atomic operator definition in our formulation---namely, that each leaf represents a semantically irreducible operator. We compare the expert judgments with the LLM-based assessment and observe an agreement rate of 97.3\%, indicating strong alignment between model and human evaluation.
\begin{algorithm}[t]
    \caption{Atomic Tree Construction Pipeline}
    \label{alg:atomic_tree}
    \SetKwInOut{Input}{Input}
    \SetKwInOut{Output}{Output}
    \SetKwFunction{Encode}{Encode}
    \SetKwFunction{KCenter}{KCenterGreedy}
    \SetKwFunction{GenTree}{GenSubTree}
    \SetKwFunction{Merge}{Merge}
    \SetKwFunction{Dedup}{SemanticAssessment}
    \SetKwFunction{ExpandDepth}{DepthExpansion}
    \SetKwFunction{ExpandBreadth}{BreadthExpansion}
    \SetKwFunction{Ext}{Extract}
    
    \Input{Candidate Datasets $\mathcal{D}$, Embedding Model $\mathcal{M}_{emb}$, Generation Model $\mathcal{M}_{gen}$, Max Interaction $\mathcal{I}$}
    \Output{Atomic Tree $\mathcal{T}$}

    \tcp{Phase 1: Representative Data Selection}
    $V \leftarrow \Encode(\mathcal{D}, \mathcal{M}_{bge})$,
    $\mathcal{S}_{rep} \leftarrow \KCenter(V)$,
    $\mathcal{T} \leftarrow \{root\}$

    $\mathbb{A}_{rep} \leftarrow \Ext({S}_{rep})$

    \tcp{Phase 2: Level 1 - Initial Tree Generation}
    \For{batch $\mathbb{A}_{sub} \in sample(\mathbb{A}_{rep})$}{
        $\mathcal{T}_{sub} \leftarrow \GenTree(\mathbb{A}_{sub}, \mathcal{M}_{gen})$
        $\mathcal{T} \leftarrow \Merge(\mathcal{T}, \mathcal{T}_{sub})$ 
    }

    \tcp{Phase 3: Level 2 - Tree Expansion}
    \For{iter $i \in \mathcal{I}$}{
    $\mathcal{T}_i \leftarrow \emptyset$
    
    \For{node $n \in \text{LeafNodes}(\mathcal{T})$}{
        $\mathcal{N}_{d} \leftarrow \ExpandDepth(n, \mathcal{M}_{gen})$
        $\mathcal{N}_{b} \leftarrow \ExpandBreadth(n, \mathcal{M}_{gen})$
        
        \If{$\mathcal{N}_{d} \neq \emptyset \lor \mathcal{N}_{b} \neq \emptyset$}{
            $\mathcal{T}_i \leftarrow \Merge(\mathcal{T}_i, \mathcal{N}_{d} \cup \mathcal{N}_{b})$\
        }}
    \If{$\mathcal{T}_i = \emptyset$}{
        break
    }
    $\mathcal{T} \leftarrow \Merge(\mathcal{T}, \mathcal{T}_i)$
    }
    \Return{$\mathcal{T}$}
\end{algorithm}

\subsection{Pipeline for Atomic Data Synthesis}
\label{sec:appendix-SynthesisPrompt}
This section details the Atomic-QA Generation pipeline. The Atomic-X Generation pipeline is described in $\S$\ref{sec:Atomic-X} of the main text.

\subsubsection{Atomic-QA Generation}
Atomic-QA generation follows the pipeline described in $\S$\ref{sec:Atomic-QA}. Starting from each leaf node of the operator tree, we first employ an LLM to generate diverse operands $x$ within the operator's domain, guided by Prompt \ref{Operand-Generation}. We then compute the corresponding outputs $y = f(x)$ via a subsequent LLM call to form complete atomic triplets $\mathcal{A} = \langle x, f, y \rangle$. Finally, we utilize Prompt \ref{Verification} to validate the logical correctness of the generated atomic operations.
\subsubsection{Atomic-X Generation}
The Atomic-X pipeline is introduced in $\S$\ref{sec:Atomic-X}; here we detail the prompt usage. The framework operates in the reverse direction: given existing data, it decomposes responses into atomic operators. The framework comprises three core modules: the Encoder, the Perturber, and the Decoder. Guided by Prompt \ref{Encoder}, the Encoder maps the response content onto the operator tree, recursively decomposing the structure until reaching leaf operators, which are resolved into atomic operations. Next, the Perturber applies operator perturbations to each atomic operation utilizing the Prompt \ref{Perturber}. Finally, the Decoder modifies the original data by employing the SEARCH-REPLACE and TERMINATE tools defined in Prompt \ref{Decoder}.

\section{Details about Experimental Setup}
In this section we will introduce our experimental details. First, we introduce the details of data generation used in the experiments, followed by a description of the model training and evaluation procedures. \textbf{All experiments are run three times.}
\subsection{Details about Dataset Generation}
\label{sec:DatasetGeneration}
For the Atomic-QA generation process, we commence by employing the KCenterGreedy algorithm to identify 5,000 cluster centers. The initial operator tree is constructed through an iterative procedure spanning 500 rounds, with 50 subjects randomly sampled in each round. During the subsequent evolution stage, the tree evolves for 1,715 iterations. Notably, in each iteration, a single leaf operator node undergoes five depth expansions and five breadth expansions. This results in an operator tree with 6,931 leaf nodes (max depth 4), where each leaf defines a concrete atomic operator. Following the pipeline in $\S$\ref{sec:Atomic-QA}, we generate diverse operands for each leaf operator, yielding the Atomic-QA dataset with approximately 1.37 million instances.

Regarding the Atomic-X generation process, we construct a separate operator tree for each target dataset. We initiate construction by randomly sampling 10,000 instances, generating 100 cluster centers at each step. The initial tree construction involves 200 evolution rounds, wherein 50 data points are selected per round. In the final stage, the evolution iterations differ by dataset: Alpaca evolves for 1,203 iterations, while MetaMath evolves for 1,347 iterations. This results in operator trees with 276 and 398 leaf operators respectively for Alpaca (max depth 5) and MetaMath (max depth 6). From the remaining samples, we randomly select 10,000 instances and apply the Atomic-X pipeline ($\S$\ref{sec:Atomic-X}) to construct the corresponding Atomic-Alpaca and Atomic-MetaMath.

\subsection{Details about Model Training and Evaluation}
\label{sec:ModelTraining}
We conduct experiments on four representative open-source pre-trained LLMs that span different architectures and scales, including LLaMA3.1-8B, Qwen3-8B, Qwen2.5-7B, and Qwen2.5-14B.
All models are fine-tuned with LLaMA-Factory~\citep{zheng-etal-2024-llamafactory} and evaluated with the LM Evaluation Harness~\citep{eval-harness}.

\paragraph{Datasets.}
We compare models fine-tuned separately under different training dataset configurations, as summarized in Tables~\ref{tab:overall_performance} and~\ref{tab:qwen3_performance}. The comparison set includes Atomic-QA and its variants (Wiki, Wiki-Rewrite, and Atomic-LIFD), the official \emph{Instruct} versions, and six representative open-source datasets: LIMA~\citep{zhou2023limaalignment}, Condor~\citep{maosongcao-etal-2025-condor}, Magpie~\citep{xu2024magpiealignmentdatasynthesis}, WizardLM~\citep{xu2025wizardlmempoweringlargepretrained}, Alpaca~\citep{alpaca}, and Self-Instruct~\citep{wang2023selfinstructaligninglanguagemodels}. Specifically, \emph{Instruct} refers to the official tuned versions released for each pretrained model. \emph{Wiki} consists of synthetic instructions paired with Wikipedia entity summaries as responses, where the entities are drawn from the Atomic-QA entity set with available Wikipedia pages. \emph{Wiki-Rewrite} reformulates these instructions to increase diversity, serving as a variant derived from Wiki. In addition, \emph{Atomic-LIFD} is a compact distilled subset of Atomic-QA obtained via perplexity-based selection, with construction details provided in Appendix \ref{sec:trainingtoken}.

\begin{figure}[!ht]
    \centering
    \includegraphics[width=1\linewidth]{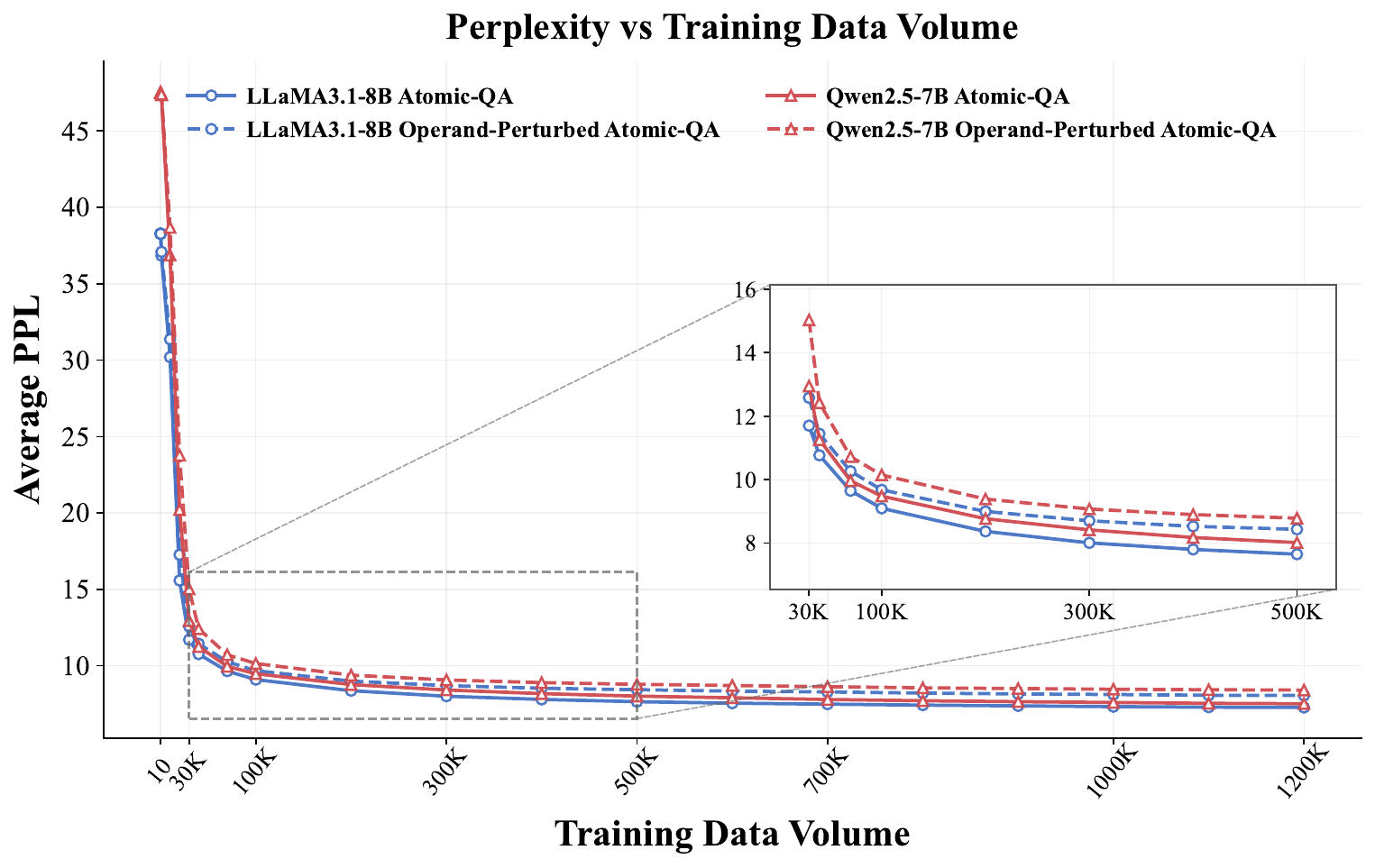}
    \caption{
        Perplexity trends of LLaMA and Qwen models as a function of training data volume. 
        The curve displays average PPL for both the original and entity-replaced variants, 
        with an inset showing a zoomed region for the 30K--500K data range. 
    }
    \label{fig:ppl_trend}
    \vspace{-10pt}
\end{figure}

\begin{figure}[!ht]
    \centering
    \includegraphics[width=1\linewidth]{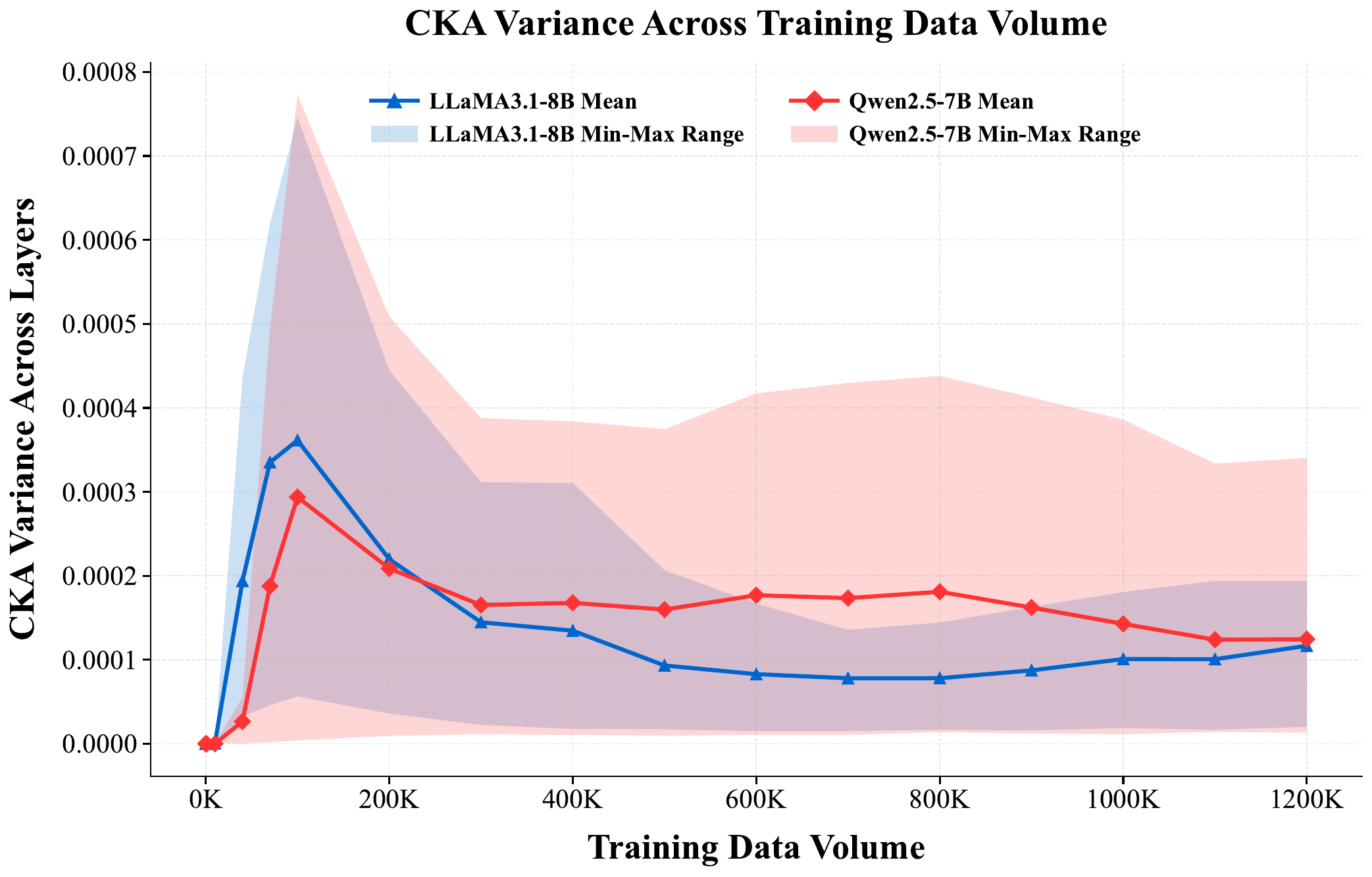}
    \caption{
    CKA variance across layers versus training data volume for LLaMA3.1-8B and Qwen2.5-7B, showing mean values (lines) and min–max ranges (shaded areas).
    }
    \label{fig:variance_aggregate}
    \vspace{-10pt}
\end{figure}

\paragraph{Training Configuration.} All models are fine-tuned using Low-Rank Adaptation (LoRA) with rank $r=16$ and a weight decay of 0.01 for 2 epochs.
We adopt a learning rate of $1 \times 10^{-4}$ by default. For experiments involving scaled training data, specifically Atomic-QA MCQ, perplexity analysis, and CKA variance analysis, we use a learning rate of $1 \times 10^{-5}$, as shown respectively in Figure~\ref{fig:mcq_main_acc}, Figure~\ref{fig:ppl_trend}, and Figure~\ref{fig:variance_aggregate}.
\paragraph{Evaluation Configuration.}
We evaluate Atomic-QA alongside other dataset baselines across a benchmark suite spanning multiple capability dimensions. In the domain of text comprehension and information integration, assessment is conducted using BoolQ \citep{clark2019boolqexploringsurprisingdifficulty}, LongBench-HotpotQA \citep{bai2024longbenchbilingualmultitaskbenchmark}, RACE \citep{lai2017racelargescalereadingcomprehension}, and SQuADv2 \citep{rajpurkar2018knowdontknowunanswerable}. With regard to knowledge-driven reasoning, capabilities are measured through ARC-Easy/Challenge \citep{clark2018thinksolvedquestionanswering} and OpenBookQA \citep{mihaylov2018suitarmorconductelectricity}. Additionally, for commonsense reasoning, evaluation is performed via PIQA \citep{bisk2019piqareasoningphysicalcommonsense}, SIQA \citep{sap2019socialiqacommonsensereasoningsocial}, and WinoGrande \citep{sakaguchi2019winograndeadversarialwinogradschema}.

To analyze the impact of increasing atomic error rates (from 0\% to 100\%) on model performance, we evaluate datasets generated via the Atomic-X pipeline. Specifically, Atomic-Alpaca is tested on the identical ten benchmarks introduced above to observe how different error levels influence general capability dimensions. In parallel, Atomic-MetaMath is evaluated on specialized mathematical reasoning tasks, including GSM8K~\citep{cobbe2021trainingverifierssolvemath}, Math500 \citep{hendrycks2021measuringmathematicalproblemsolving}, and MathQA~\citep{amini2019mathqainterpretablemathword}. 
Regarding the evaluation protocol, the majority of benchmarks are conducted under a 3-shot setting, while LongBench-HotpotQA and Atomic-QA MCQ follow a zero-shot paradigm. Results on Atomic-QA MCQ are reported in Figure~\ref{fig:mcq_main_acc}, and detailed evaluation configurations for all benchmarks are summarized in Table~\ref{tab:benchmarks}.
\begin{table}[t]
\centering
\caption{Evaluation benchmarks and configurations.}
\label{tab:benchmarks}
\resizebox{\linewidth}{!}{
\begin{tabular}{>{\centering\arraybackslash}p{3.6cm}ccc}
\toprule
\multicolumn{1}{c}{\textbf{Category}} & \textbf{Benchmark} & \textbf{Shots} & \textbf{Metric} \\
\midrule
\multirow{4}{=}{Text Comprehension and Information Integration} 
& BoolQ & 3 & Accuracy \\
& LongBench-HotpotQA & 0 & F1 \\
& RACE & 3 & Accuracy \\
& SQuADv2 & 3 & F1 \\
\midrule
\multirow{3}{=}{Knowledge-Driven Reasoning} 
& ARC-Challenge & 3 & Accuracy \\
& ARC-Easy & 3 & Accuracy \\
& OpenBookQA & 3 & Accuracy \\
\midrule
\multirow{3}{=}{Commonsense Reasoning} 
& PIQA & 3 & Accuracy \\
& SIQA & 3 & Accuracy \\
& WinoGrande & 3 & Accuracy \\
\midrule
\multirow{3}{=}{Mathematical Reasoning} 
& GSM8K & 3 & Exact Match \\
& Hendrycks Math500 & 3 & Exact Match \\
& MathQA & 3 & Accuracy \\
\bottomrule
\end{tabular}
}
\end{table}

\section{Additional Results}
\label{ref:addresult}
In this section, we present additional experimental results, structured to mirror the organization of the main text. We begin with supplementary findings on Quantifying the Cost of Strict Filtering, followed by further details on Measuring Operator–Operand and Dominance.

\begin{table*}[!ht]
  \centering
  \scriptsize
  \caption{Overall performance across all training datasets on Qwen2.5-7B and Qwen2.5-14B. Color-coded cells (\textcolor{excellent}{green}/\textcolor{verypoor}{red}) are used to indicate performance above or below the mean across methods (All Average), with darker colors representing better or worse performance.}
  \resizebox{0.9\linewidth}{!}{
  \begin{tabular}{c|c*{11}{c}}
    \toprule
    \multicolumn{2}{c|}{\textbf{Model Family}} & \makecell{\textbf{ARC-}\\\textbf{Challenge}} & \makecell{\textbf{ARC-}\\\textbf{Easy}} & \textbf{BoolQ} & \makecell{\textbf{Hotpot-}\\\textbf{QA}} & \makecell{\textbf{Open-}\\\textbf{BookQA}} & \textbf{PIQA} & \textbf{RACE} & \textbf{SIQA} & \textbf{SQuADv2} & \makecell{\textbf{Wino-}\\\textbf{Grande}} & \textbf{Avg} \\
    \hline

    % Qwen2.5-7B 部分
    \multirow{13}{*}{\rotatebox{90}{\textbf{Qwen2.5-7B}}} 
    
    & Base & \cellcolor{average}0.590 & \cellcolor{average}0.860 & \cellcolor{average}0.872 & \cellcolor{average}0.216 & \cellcolor{average}0.390 & \cellcolor{average}0.803 & \cellcolor{average}0.459 & \cellcolor{average}0.576 & \cellcolor{average}0.421 & \cellcolor{average}0.754 & \cellcolor{average}0.594 \\
    
  & All Average & \cellcolor{average}0.586 & \cellcolor{average}0.859 & \cellcolor{average}0.875 & \cellcolor{average}0.190 & \cellcolor{average}0.399 & \cellcolor{average}0.800 & \cellcolor{average}0.474 & \cellcolor{average}0.572 & \cellcolor{average}0.364 & \cellcolor{average}0.747 & \cellcolor{average}0.587 \\
\cline{2-13}

& Instruct & \cellcolor{excellent}0.616 & \cellcolor{excellent}0.873 & \cellcolor{poor}0.865 & \cellcolor{belowavg}0.143 & \cellcolor{slightabove}0.404 & \cellcolor{verypoor}0.786 & \cellcolor{excellent}0.512 & \cellcolor{slightbelow}0.566 & \cellcolor{belowavg}0.324 & \cellcolor{poor}0.735 & \cellcolor{slightbelow}0.582 \\

& LIMA & \cellcolor{average}0.586 & \cellcolor{slightabove}0.861 & \cellcolor{slightbelow}0.871 & \cellcolor{slightabove}0.207 & \cellcolor{slightbelow}0.396 & \cellcolor{slightabove}0.801 & \cellcolor{belowavg}0.459 & \cellcolor{slightabove}0.574 & \cellcolor{excellent}0.424 & \cellcolor{good}0.757 & \cellcolor{slightabove}0.594 \\
& Condor & \cellcolor{aboveavg}0.597 & \cellcolor{aboveavg}0.864 & \cellcolor{slightbelow}0.874 & \cellcolor{slightbelow}0.185 & \cellcolor{slightbelow}0.392 & \cellcolor{slightabove}0.801 & \cellcolor{slightbelow}0.471 & \cellcolor{good}0.580 & \cellcolor{good}0.396 & \cellcolor{aboveavg}0.755 & \cellcolor{slightabove}0.591 \\
& Magpie & \cellcolor{slightbelow}0.582 & \cellcolor{slightbelow}0.857 & \cellcolor{slightabove}0.876 & \cellcolor{aboveavg}0.241 & \cellcolor{verypoor}0.364 & \cellcolor{belowavg}0.795 & \cellcolor{belowavg}0.455 & \cellcolor{average}0.572 & \cellcolor{excellent}0.412 & \cellcolor{belowavg}0.740 & \cellcolor{slightabove}0.589 \\

& WizardLM & \cellcolor{good}0.606 & \cellcolor{aboveavg}0.863 & \cellcolor{aboveavg}0.879 & \cellcolor{verypoor}0.050 & \cellcolor{aboveavg}0.412 & \cellcolor{excellent}0.811 & \cellcolor{slightbelow}0.464 & \cellcolor{excellent}0.584 & \cellcolor{verypoor}0.259 & \cellcolor{good}0.756 & \cellcolor{belowavg}0.568 \\
& Alpaca & \cellcolor{average}0.586 & \cellcolor{slightabove}0.862 & \cellcolor{excellent}0.883 & \cellcolor{verypoor}0.037 & \cellcolor{slightbelow}0.392 & \cellcolor{excellent}0.808 & \cellcolor{slightabove}0.482 & \cellcolor{excellent}0.587 & \cellcolor{verypoor}0.265 & \cellcolor{average}0.747 & \cellcolor{belowavg}0.565 \\
& Self-Instruct & \cellcolor{verypoor}0.490 & \cellcolor{verypoor}0.816 & \cellcolor{verypoor}0.859 & \cellcolor{verypoor}0.062 & \cellcolor{poor}0.376 & \cellcolor{poor}0.792 & \cellcolor{verypoor}0.430 & \cellcolor{verypoor}0.524 & \cellcolor{poor}0.304 & \cellcolor{verypoor}0.723 & \cellcolor{verypoor}0.537 \\

& Wiki & \cellcolor{slightbelow}0.583 & \cellcolor{good}0.868 & \cellcolor{slightabove}0.876 & \cellcolor{aboveavg}0.248 & \cellcolor{slightbelow}0.398 & \cellcolor{slightabove}0.803 & \cellcolor{slightbelow}0.470 & \cellcolor{aboveavg}0.576 & \cellcolor{excellent}0.426 & \cellcolor{excellent}0.763 & \cellcolor{aboveavg}0.601 \\
& Wiki-Rewrite & \cellcolor{slightbelow}0.584 & \cellcolor{excellent}0.870 & \cellcolor{slightabove}0.876 & \cellcolor{slightabove}0.204 & \cellcolor{slightabove}0.404 & \cellcolor{slightabove}0.802 & \cellcolor{slightbelow}0.470 & \cellcolor{slightabove}0.576 & \cellcolor{excellent}0.422 & \cellcolor{aboveavg}0.751 & \cellcolor{aboveavg}0.596 \\
& Atomic-LIFD & \cellcolor{good}0.606 & \cellcolor{average}0.859 & \cellcolor{excellent}0.882 & \cellcolor{excellent}0.383 & \cellcolor{excellent}0.432 & \cellcolor{slightbelow}0.797 & \cellcolor{good}0.500 & \cellcolor{slightabove}0.576 & \cellcolor{aboveavg}0.392 & \cellcolor{slightbelow}0.743 & \cellcolor{excellent}0.617 \\
& Atomic-QA & \cellcolor{excellent}0.609 & \cellcolor{aboveavg}0.860 & \cellcolor{good}0.880 & \cellcolor{good}0.327
& \cellcolor{good}0.419 & \cellcolor{slightabove}0.801 & \cellcolor{excellent}0.505 & \cellcolor{aboveavg}0.577 & \cellcolor{aboveavg}0.386 & \cellcolor{slightbelow}0.746 & \cellcolor{excellent}0.611 \\
\hline

\multirow{13}{*}{\rotatebox{90}{\textbf{Qwen2.5-14B}}}

& Base
& \cellcolor{average}0.637
& \cellcolor{average}0.878
& \cellcolor{average}0.888
& \cellcolor{average}0.147
& \cellcolor{average}0.404
& \cellcolor{average}0.822
& \cellcolor{average}0.487
& \cellcolor{average}0.579
& \cellcolor{average}0.392
& \cellcolor{average}0.793
& \cellcolor{average}0.603 \\

& All Average & \cellcolor{average}0.630 & \cellcolor{average}0.879 & \cellcolor{average}0.884 & \cellcolor{average}0.169 & \cellcolor{average}0.409 & \cellcolor{average}0.816 & \cellcolor{average}0.508 & \cellcolor{average}0.586 & \cellcolor{average}0.393 & \cellcolor{average}0.791 & \cellcolor{average}0.607 \\
\cline{2-13}

& Instruct & \cellcolor{excellent}0.699 & \cellcolor{excellent}0.902 & \cellcolor{slightbelow}0.883 & \cellcolor{slightbelow}0.160 & \cellcolor{aboveavg}0.430 & \cellcolor{good}0.821 & \cellcolor{excellent}0.577 & \cellcolor{excellent}0.611 & \cellcolor{good}0.411 & \cellcolor{slightbelow}0.785 & \cellcolor{good}0.628 \\

& LIMA & \cellcolor{slightabove}0.634 & \cellcolor{slightabove}0.880 & \cellcolor{slightabove}0.886 & \cellcolor{slightbelow}0.139 & \cellcolor{slightbelow}0.404 & \cellcolor{good}0.820 & \cellcolor{poor}0.484 & \cellcolor{slightbelow}0.582 & \cellcolor{slightabove}0.395 & \cellcolor{aboveavg}0.800 & \cellcolor{slightbelow}0.602 \\
& Condor & \cellcolor{slightabove}0.631 & \cellcolor{average}0.879 & \cellcolor{slightabove}0.885 & \cellcolor{belowavg}0.133 & \cellcolor{belowavg}0.396 & \cellcolor{slightabove}0.818 & \cellcolor{belowavg}0.490 & \cellcolor{slightabove}0.588 & \cellcolor{slightbelow}0.379 & \cellcolor{aboveavg}0.800 & \cellcolor{slightbelow}0.600 \\
& Magpie & \cellcolor{average}0.630 & \cellcolor{slightabove}0.882 & \cellcolor{good}0.893 & \cellcolor{verypoor}0.037 & \cellcolor{verypoor}0.368 & \cellcolor{belowavg}0.811 & \cellcolor{belowavg}0.496 & \cellcolor{slightbelow}0.580 & \cellcolor{good}0.415 & \cellcolor{poor}0.776 & \cellcolor{belowavg}0.589 \\

& WizardLM & \cellcolor{slightabove}0.640 & \cellcolor{aboveavg}0.888 & \cellcolor{good}0.893 & \cellcolor{poor}0.076 & \cellcolor{slightbelow}0.400 & \cellcolor{excellent}0.824 & \cellcolor{slightabove}0.520 & \cellcolor{slightabove}0.590 & \cellcolor{poor}0.351 & \cellcolor{slightbelow}0.789 & \cellcolor{belowavg}0.597 \\
& Alpaca & \cellcolor{slightbelow}0.609 & \cellcolor{slightabove}0.880 & \cellcolor{excellent}0.898 & \cellcolor{slightbelow}0.143 & \cellcolor{belowavg}0.394 & \cellcolor{good}0.822 & \cellcolor{slightbelow}0.500 & \cellcolor{aboveavg}0.597 & \cellcolor{excellent}0.427 & \cellcolor{aboveavg}0.796 & \cellcolor{slightabove}0.607 \\
& Self-Instruct & \cellcolor{verypoor}0.518 & \cellcolor{verypoor}0.826 & \cellcolor{verypoor}0.850 & \cellcolor{aboveavg}0.263 & \cellcolor{verypoor}0.368 & \cellcolor{verypoor}0.800 & \cellcolor{verypoor}0.463 & \cellcolor{verypoor}0.534 & \cellcolor{verypoor}0.319 & \cellcolor{verypoor}0.763 & \cellcolor{verypoor}0.570 \\

& Wiki & \cellcolor{slightbelow}0.629 & \cellcolor{slightabove}0.880 & \cellcolor{belowavg}0.874 & \cellcolor{verypoor}0.068 & \cellcolor{slightabove}0.416 & \cellcolor{slightabove}0.817 & \cellcolor{average}0.508 & \cellcolor{aboveavg}0.597 & \cellcolor{good}0.417 & \cellcolor{excellent}0.807 & \cellcolor{slightbelow}0.601 \\
& Wiki-Rewrite & \cellcolor{slightbelow}0.626 & \cellcolor{slightabove}0.880 & \cellcolor{slightbelow}0.882 & \cellcolor{verypoor}0.063 & \cellcolor{slightbelow}0.408 & \cellcolor{good}0.820 & \cellcolor{slightabove}0.515 & \cellcolor{slightabove}0.592 & \cellcolor{good}0.416 & \cellcolor{excellent}0.811 & \cellcolor{slightbelow}0.601 \\
& Atomic-LIFD & \cellcolor{aboveavg}0.653 & \cellcolor{slightbelow}0.878 & \cellcolor{good}0.894 & \cellcolor{excellent}0.414 & \cellcolor{excellent}0.466 & \cellcolor{belowavg}0.808 & \cellcolor{slightbelow}0.507 & \cellcolor{slightbelow}0.582 & \cellcolor{slightbelow}0.392 & \cellcolor{poor}0.770 & \cellcolor{excellent}0.636 \\
& Atomic-QA & \cellcolor{aboveavg}0.666 & \cellcolor{aboveavg}0.889 & \cellcolor{slightbelow}0.880 & \cellcolor{excellent}0.367 & \cellcolor{good}0.449 & \cellcolor{slightabove}0.817
& \cellcolor{aboveavg}0.539 & \cellcolor{slightabove}0.592 & \cellcolor{aboveavg}0.405 & \cellcolor{good}0.803 & \cellcolor{excellent}0.641 \\

\bottomrule
  \end{tabular}
  }
  \label{tab:qwen3_performance}
\end{table*}

\subsection{Training Tokens Comparison}
\label{sec:trainingtoken}
Table~\ref{tab:training_tokens} reports the total number of training tokens for each dataset. The datasets vary substantially in scale, ranging from fewer than one million tokens (e.g., LIMA) to over two hundred million tokens (e.g., Magpie), with several datasets clustered in the range of a few million to several tens of millions of tokens. Given this substantial variation, it is important to determine whether dataset scale or structural diversity serves as the primary determinant of performance. To investigate this question, we construct Atomic-LIFD, a compact dataset of approximately 2.7M tokens that is larger than LIMA but smaller than other considered datasets. Specifically, this subset is derived by applying perplexity-based selection (IFD) to Atomic-QA, a logically verified and structurally diverse source. 

Remarkably, Atomic-LIFD achieves performance comparable to that of the full Atomic-QA dataset, as shown in Table~\ref{tab:overall_performance} and Table~\ref{tab:qwen3_performance}. Given that data scale is the sole variable distinguishing these two settings, this minimal performance gap strongly suggests that data quantity is not the decisive factor behind Atomic-QA's superiority.
\begin{table}[t]
    \centering
    \small
    \caption{The number of training tokens in different datasets.
    }
    \begin{tabular}{c r}
        \toprule
        \textbf{Dataset} & \textbf{Training Tokens} \\
        \midrule
        LIMA                  &  627{,}993\\
        Condor                &  21{,}131{,}550\\
        Magpie                &  203{,}256{,}865\\
        WizardLM              &  27{,}673{,}025\\
        Alpaca                &  3{,}766{,}455\\
        Self-Instruct         &  4{,}536{,}225\\
        Wiki                  &  11{,}082{,}377\\
        Wiki-Rewrite          &  11{,}356{,}362\\
        Atomic-QA             &  34{,}901{,}525\\
        \textbf{Atomic-LIFD} & \textbf{2{,}701{,}510} \\
        \bottomrule
    \end{tabular}
    \label{tab:training_tokens}
\end{table}

\subsection{Quantifying the Cost of Strict Filtering}
Table \ref{tab:qwen3_performance} extends our evaluation to Qwen2.5-7B and Qwen2.5-14B to verify the robustness of our approach. Consistent with the findings in the main text, Atomic-QA demonstrates strong generalizability, achieving the highest average performance across all baselines on both backbones (61.2\% on Qwen2.5-7B and 62.5\% on Qwen3-8B).

The persistent performance gap over LIMA (+1.8\% to +2.3\%) and other synthetic methods (Alpaca, WizardLM) confirms that our conclusion is model-agnostic: prioritizing structural scale over strict filtering consistently yields better instruction-following and reasoning capabilities, regardless of the underlying model architecture or size.

\begin{figure*}[!ht]
    \centering
    \includegraphics[width=0.9\textwidth]{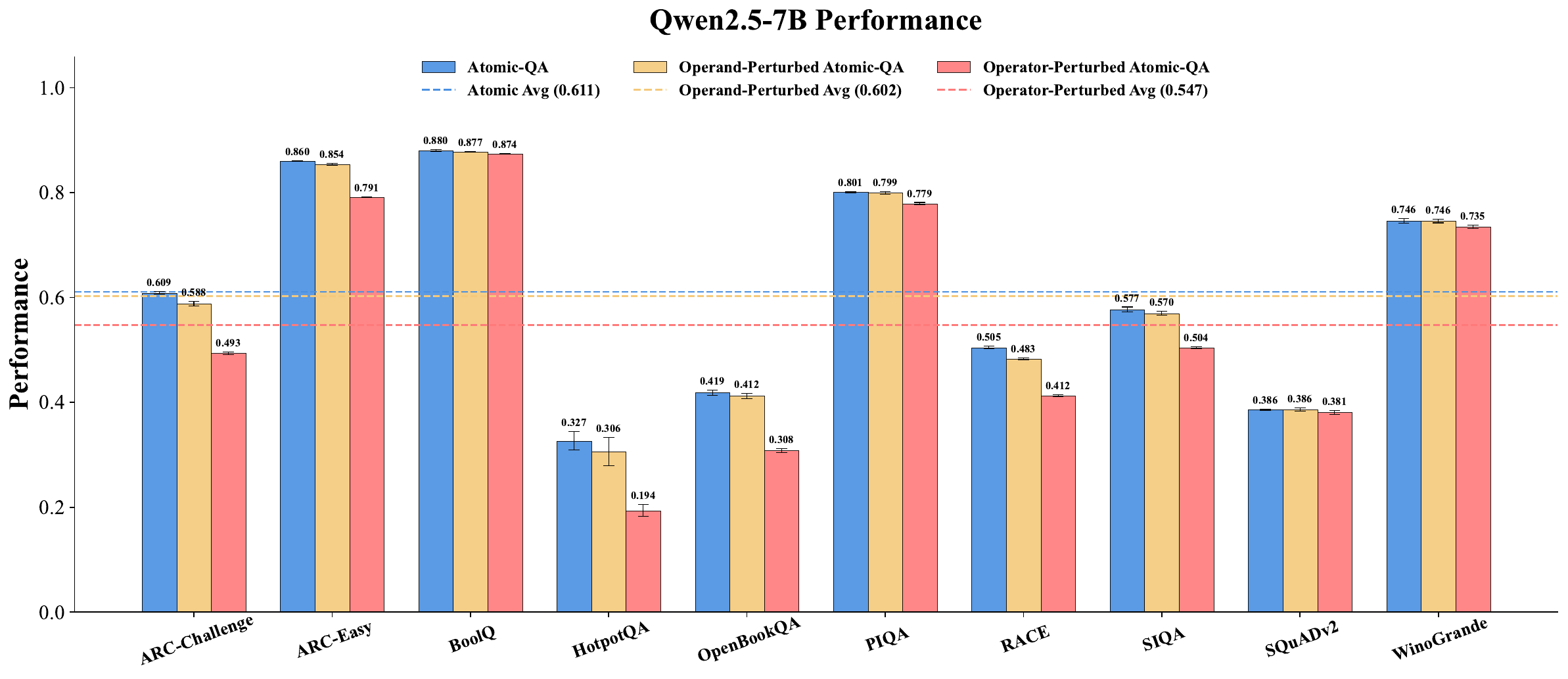}
    \caption{
    Performance comparison of Qwen2.5-7B across the Atomic-QA, Operand-Perturbed Atomic-QA, and Operator-Perturbed Atomic-QA settings. Each bar shows the model’s accuracy on individual tasks, while the dashed lines denote the average performance for each setting.
    }
    \label{fig:qwen2_5_7b}
\end{figure*}

\begin{figure*}[!ht]
    \centering
    \includegraphics[width=0.9\textwidth]{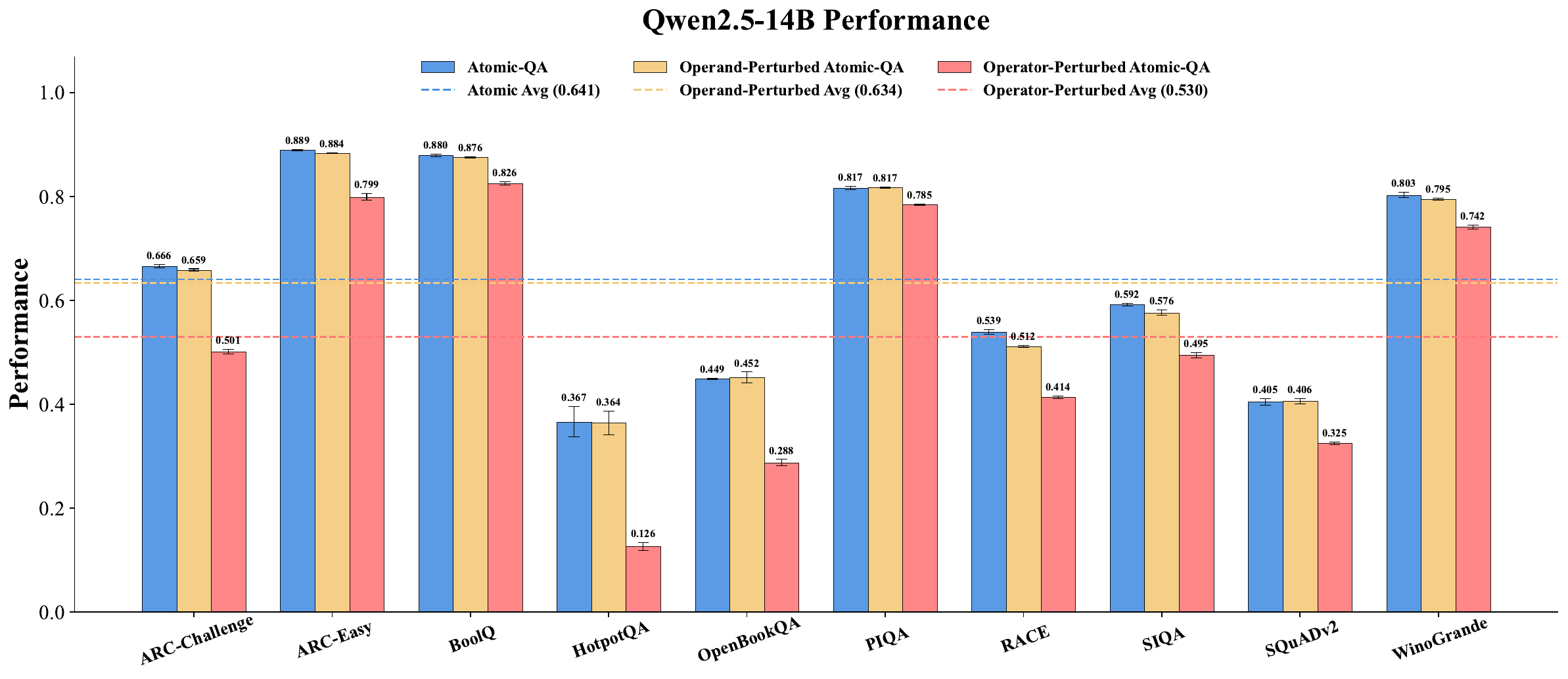}
    \caption{
    Performance comparison of Qwen2.5-14B across the Atomic-QA, Operand-Perturbed Atomic-QA, and Operator-Perturbed Atomic-QA settings. Each bar shows the model’s accuracy on individual tasks, while the dashed lines denote the average performance for each setting.
    }
    \label{fig:qwen2_5_14b}
\end{figure*}

\begin{figure*}[!ht]
    \centering
    \includegraphics[width=0.9\textwidth]{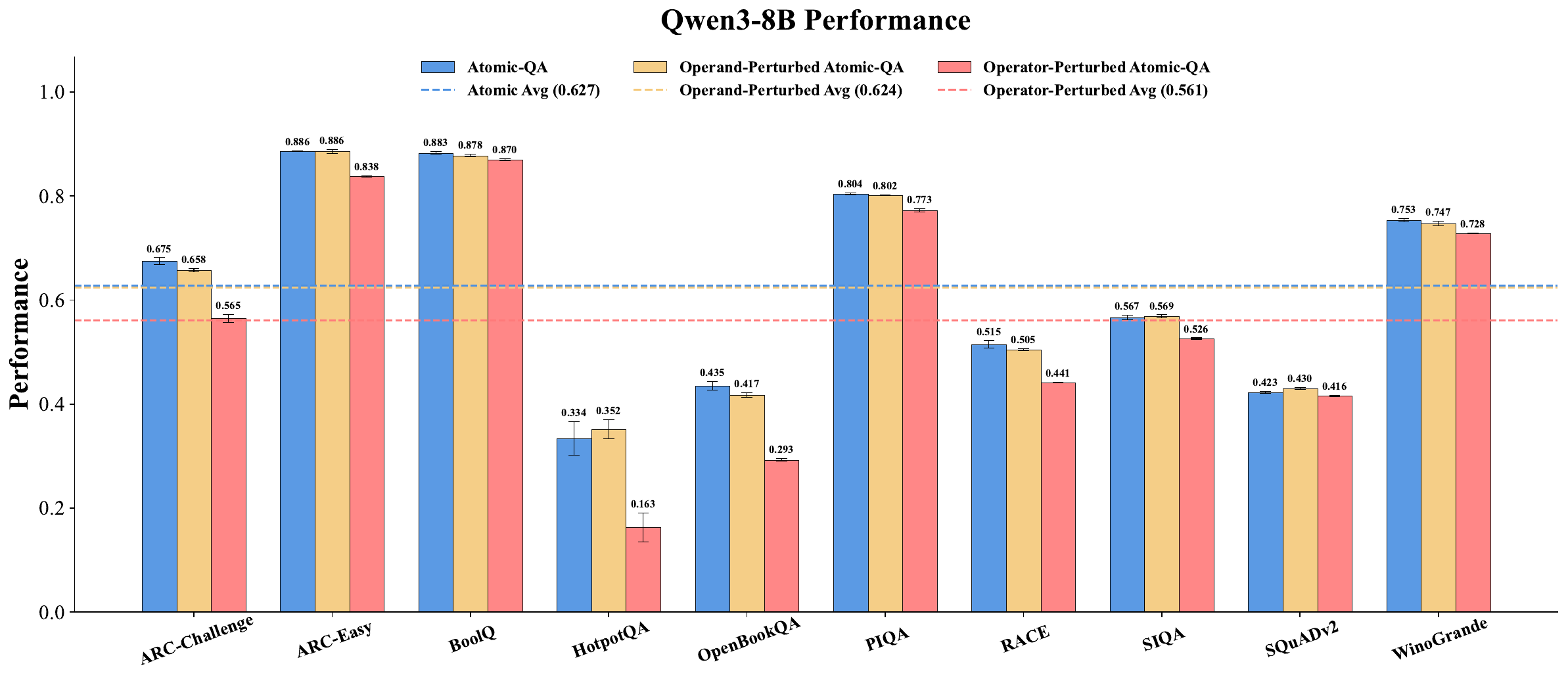}
    \caption{
    Performance comparison of Qwen3-8B across the Atomic-QA, Operand-Perturbed Atomic-QA, and Operator-Perturbed Atomic-QA settings. Each bar shows the model’s accuracy on individual tasks, while the dashed lines denote the average performance for each setting.
    }
    \label{fig:qwen3_8b}
    \vspace{-10pt}
\end{figure*}

\subsection{Measuring Operator–Operand and Dominance}
We further validated our findings on the Qwen model family, as illustrated in Figures \ref{fig:qwen2_5_7b}, \ref{fig:qwen2_5_14b}, and \ref{fig:qwen3_8b}. Across varying model sizes (7B, 8B, and 14B), Operand-Perturbed Atomic-QA consistently aligns with standard Atomic-QA performance with negligible degradation (e.g., an average drop of only 0.002 for Qwen3-8B). This stability demonstrates that robustness to operand errors is a generalizable phenomenon, independent of specific model architectures.

To substantiate our main findings, we visualize the training dynamics in Figures \ref{fig:ppl_trend} and \ref{fig:variance_aggregate}. Figure \ref{fig:ppl_trend} shows that the perplexity curves for the Operand-Perturbed Atomic-QA and the original dataset are nearly indistinguishable, demonstrating that operand errors impose almost no additional optimization burden. This indicates the model learns the structural logic with equal efficiency. Furthermore, Figure \ref{fig:variance_aggregate} illustrates that the variance in CKA similarity initially spikes but subsequently stabilizes and decreases as data volume increases. This convergence confirms that the high representational alignment discussed in the main text is a robust and stable outcome of training, rather than a transient state.

\begin{figure}[!ht]
    \centering
    \includegraphics[width=0.9\linewidth]{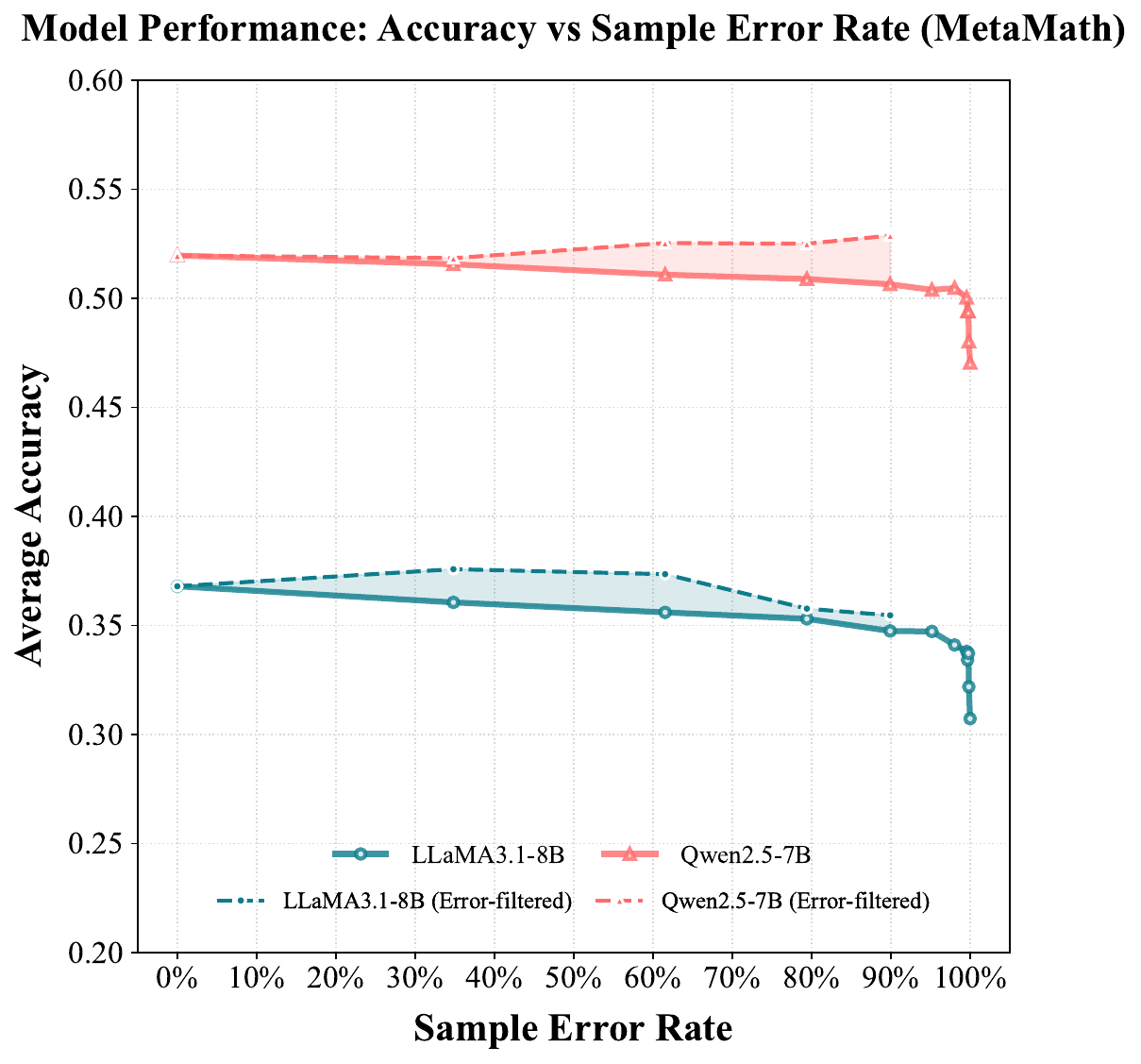}
    \caption{
     Accuracy on GSM8K, MATH-500, and MathQA for LLaMA 3.1-8B and Qwen 2.5-7B fine-tuned on \textbf{Atomic-MetaMath} with operator-level perturbations at controlled rates. Corresponds to Figure \ref{fig:real_error_sensitivity-alpaca}, which reports Atomic-Alpaca on general benchmarks.
    }
    \label{fig:real_error_sensitivity-math}
    \vspace{-10pt}
\end{figure}

Finally, to verify the generalizability of the degradation patterns in operation errors, we extended our analysis to mathematical reasoning using the MetaMath dataset. As illustrated in Figure \ref{fig:real_error_sensitivity-math}, consistent with the collapse observed in Alpaca, both LLaMA 3.1-8B and Qwen 2.5-7B eventually exhibit a sharp performance decline. However, a key distinction lies in the critical threshold. While models trained on Alpaca begin to collapse around an error rate of $0.6$, those trained on MetaMath demonstrate higher robustness, maintaining stability until the error rate approaches saturation ($>0.9$). This confirms that while the non-linear degradation dynamics are universal across tasks, the tipping point for structural collapse depends on domain complexity. Overall, the models exhibit exceptional resilience to operand perturbations. Remarkably, even when the operand error rate reaches almost 100\%, performance remains largely unaffected. In stark contrast, a 100\% error rate in operators precipitates a significant performance decline across both the Alpaca and MetaMath datasets.

%\begin{figure}[!ht]
%    \centering

    % --- Top: LLaMA ---
%    \begin{minipage}[!ht]{0.95\linewidth}
%        \centering
%        \includegraphics[width=\linewidth]{images/alpaca_error_merged_cka_heatmap.pdf}
%        \vspace{2pt}
%        {\small (a) LLaMA Alpaca-Subset Last-Layer CKA Similarity}
%    \end{minipage}

%    \vspace{8pt}  % spacing between top & bottom

    % --- Bottom: Qwen ---
%    \begin{minipage}[!ht]{0.95\linewidth}
%        \centering
%        \includegraphics[width=\linewidth]{images/math_error_merged_cka_heatmap.pdf}
%        \vspace{2pt}
%        {\small (b) Qwen Alpaca-Subset Last-Layer CKA Similarity}
%    \end{minipage}

%    \caption{
%        Comparison of last-layer CKA similarity under different Alpaca subset sizes for 
%        LLaMA (a) and Qwen (b). Increasing Alpaca proportions (0--100\%) show consistent
%        representational shifts across both model families.
%    }
%    \label{fig:alpaca_last_layer_heatmap_vertical}
%\end{figure}

\subsection{Layer-Wise CKA Analysis}
\label{sec:appendix-layerwise-cka}

We computed CKA at every layer for LLaMA3.1-8B and Qwen2.5-7B. Each row of Table~\ref{tab:layerwise_cka} reports the CKA with the model trained on Atomic-QA, averaged over the early, middle, and late thirds of the layers. Across depth, operand-perturbed models stay close to the clean model while operator-perturbed models sit lower. The gap is largest at the last layer on both backbones (LLaMA 0.935 vs.\ 0.832; Qwen 0.950 vs.\ 0.899).

\begin{table}[t]
    \centering
    \small
    \caption{Layer-wise CKA similarity against the model trained on clean Atomic-QA, averaged over the early, middle, and late thirds of the layers.}
    \label{tab:layerwise_cka}
    \resizebox{\linewidth}{!}{
    \begin{tabular}{llcccc}
        \toprule
        \textbf{Model} & \textbf{Perturbation} & \textbf{Early} & \textbf{Middle} & \textbf{Late} & \textbf{Last layer} \\
        \midrule
        \multirow{3}{*}{LLaMA3.1-8B} & Operand & 0.988 & 0.964 & 0.961 & 0.935 \\
         & Operator & 0.932 & 0.865 & 0.888 & 0.832 \\
         & Base Model & 0.942 & 0.743 & 0.668 & 0.745 \\
        \midrule
        \multirow{3}{*}{Qwen2.5-7B} & Operand & 0.998 & 0.999 & 0.994 & 0.950 \\
         & Operator & 0.998 & 0.999 & 0.988 & 0.899 \\
         & Base Model & 0.997 & 0.999 & 0.973 & 0.773 \\
        \bottomrule
    \end{tabular}}
    \vspace{-10pt}
\end{table}

\subsection{Surface-Level Coherence of Operator vs.\ Operand Perturbations}
\label{sec:appendix-ppl-analysis}

A natural question is how operator/operand perturbation relates to the distinction between coherent but non-factual and incoherent text. We examine this by measuring textual coherence via perplexity, computed by base models, as shown in Table~\ref{tab:ppl_comparison}.

\begin{table}[t]
\centering
\small
\caption{Average perplexity of training corpora evaluated by pretrained checkpoints. Operand-Pert.\ is reported for both the random-string construction (w/ r.s., used in $\S$\ref{sec:Measuring}) and the LLM-rewritten synthetic-entity construction (w/ s.e., Appendix~\ref{sec:appendix-operand-realism}).}
\label{tab:ppl_comparison}
\resizebox{\linewidth}{!}{
\begin{tabular}{lcccc}
\toprule
\multirow{2}{*}{\textbf{Model}} & \multirow{2}{*}{\textbf{Atomic-QA}} & \multirow{2}{*}{\textbf{Operator-Pert.}} & \multicolumn{2}{c}{\textbf{Operand-Pert.}} \\
\cmidrule(l){4-5}
 & & & \textbf{w/ r.s.} & \textbf{w/ s.e.} \\
\midrule
LLaMA3.1-8B   & 35.80 & 36.53 & 99.76  & 35.82 \\
Qwen3-8B      & 36.00 & 33.04 & 90.78  & 32.98 \\
Qwen2.5-7B    & 43.33 & 42.17 & 136.24 & 42.32 \\
Qwen2.5-14B   & 52.25 & 39.48 & 131.47 & 39.91 \\
\bottomrule
\end{tabular}}
\end{table}

Operator-Perturbed data shows perplexity nearly identical to  Atomic-QA like 35.80 vs. 36.53 on LLaMA3.1-8B, indicating high surface fluency despite logical corruption. Operand-Perturbed data with random strings, by contrast, yields roughly 3× higher perplexity (99.76), reflecting strong surface incoherence caused by random-string operands. This exposes a counterintuitive asymmetry: the surface-incoherent operand variant is benign for training, whereas the surface-coherent operator variant is catastrophic.

\subsection{About the Realism of Operand Perturbation}
\label{sec:appendix-operand-realism}

To address the concern regarding the realism of operand perturbation, we adopt LLM rewriting to construct a set of plausible but synthetic entities as an external form of operand perturbation, following the same protocol as the operator perturbation, and train four models on this data. Below, \textbf{w/ r.s.} denotes the random-string construction of $\S$\ref{sec:Measuring} and \textbf{w/ s.e.} the synthetic-entity construction. Table~\ref{tab:operand_realism_perf} shows that the models trained on the two operand-perturbation variants perform equivalently on every backbone (per-benchmark performance deltas within $\pm 0.021$).

\begin{table*}[!ht]
  \centering
  \small
  \caption{Downstream performance of the two operand-perturbation constructions. \textbf{w/ r.s.}: random strings ($\S$\ref{sec:Measuring}); \textbf{w/ s.e.}: plausible but synthetic entities generated with the same LLM-rewriting protocol as the operator perturbation. Parenthesised values are deltas relative to w/ r.s.}
  \label{tab:operand_realism_perf}
  \resizebox{\linewidth}{!}{
  \begin{tabular}{llcccccccccc}
    \toprule
    \textbf{Model} & \textbf{Perturbation} & \makecell{\textbf{ARC-}\\\textbf{Challenge}} & \makecell{\textbf{ARC-}\\\textbf{Easy}} & \textbf{BoolQ} & \makecell{\textbf{Hotpot-}\\\textbf{QA}} & \makecell{\textbf{Open-}\\\textbf{BookQA}} & \textbf{PIQA} & \textbf{RACE} & \textbf{SIQA} & \textbf{SQuADv2} & \makecell{\textbf{Wino-}\\\textbf{Grande}} \\
    \midrule
    \multirow{2}{*}{LLaMA3.1-8B} & w/ r.s. & 0.587 & 0.856 & 0.858 & 0.159 & 0.413 & 0.814 & 0.478 & 0.549 & 0.394 & 0.770 \\
     & w/ s.e. & 0.572 \tiny{($-$0.015)} & 0.862 \tiny{($+$0.006)} & 0.860 \tiny{($+$0.002)} & 0.158 \tiny{($-$0.001)} & 0.408 \tiny{($-$0.005)} & 0.825 \tiny{($+$0.011)} & 0.476 \tiny{($-$0.002)} & 0.552 \tiny{($+$0.003)} & 0.401 \tiny{($+$0.007)} & 0.770 \tiny{($+$0.000)} \\
    \midrule
    \multirow{2}{*}{Qwen2.5-7B} & w/ r.s. & 0.588 & 0.854 & 0.877 & 0.306 & 0.412 & 0.799 & 0.483 & 0.570 & 0.386 & 0.746 \\
     & w/ s.e. & 0.609 \tiny{($+$0.021)} & 0.842 \tiny{($-$0.012)} & 0.884 \tiny{($+$0.007)} & 0.313 \tiny{($+$0.007)} & 0.408 \tiny{($-$0.004)} & 0.801 \tiny{($+$0.002)} & 0.476 \tiny{($-$0.007)} & 0.570 \tiny{($+$0.000)} & 0.381 \tiny{($-$0.005)} & 0.746 \tiny{($+$0.000)} \\
    \midrule
    \multirow{2}{*}{Qwen2.5-14B} & w/ r.s. & 0.659 & 0.884 & 0.876 & 0.364 & 0.452 & 0.817 & 0.512 & 0.576 & 0.406 & 0.795 \\
     & w/ s.e. & 0.659 \tiny{($+$0.000)} & 0.877 \tiny{($-$0.007)} & 0.880 \tiny{($+$0.004)} & 0.362 \tiny{($-$0.002)} & 0.455 \tiny{($+$0.003)} & 0.809 \tiny{($-$0.008)} & 0.527 \tiny{($+$0.015)} & 0.576 \tiny{($+$0.000)} & 0.408 \tiny{($+$0.002)} & 0.793 \tiny{($-$0.002)} \\
    \midrule
    \multirow{2}{*}{Qwen3-8B} & w/ r.s. & 0.658 & 0.886 & 0.878 & 0.352 & 0.417 & 0.802 & 0.505 & 0.569 & 0.430 & 0.747 \\
     & w/ s.e. & 0.662 \tiny{($+$0.004)} & 0.884 \tiny{($-$0.002)} & 0.878 \tiny{($+$0.000)} & 0.348 \tiny{($-$0.004)} & 0.435 \tiny{($+$0.018)} & 0.804 \tiny{($+$0.002)} & 0.511 \tiny{($+$0.006)} & 0.567 \tiny{($-$0.002)} & 0.430 \tiny{($+$0.000)} & 0.750 \tiny{($+$0.003)} \\
    \bottomrule
  \end{tabular}}
\end{table*}

\paragraph{Representational Alignment.} To further validate this point, we conduct a CKA similarity evaluation in Table~\ref{tab:operand_realism_cka}. The representations learned by the models trained on the two variants are also relatively consistent.

\begin{table}[t]
    \centering
    \small
    \caption{Last-layer CKA similarity among models trained on clean Atomic-QA and the two operand-perturbation constructions.}
    \label{tab:operand_realism_cka}
    \resizebox{\linewidth}{!}{
    \begin{tabular}{llccc}
        \toprule
        \textbf{Model} & \textbf{Training Data} & \textbf{Atomic-QA} & \textbf{w/ r.s.} & \textbf{w/ s.e.} \\
        \midrule
        \multirow{3}{*}{LLaMA3.1-8B} & Atomic-QA & 1.0 & -- & -- \\
         & w/ r.s. & 0.935 & 1.0 & -- \\
         & w/ s.e. & 0.942 & 0.956 & 1.0 \\
        \midrule
        \multirow{3}{*}{Qwen2.5-7B} & Atomic-QA & 1.0 & -- & -- \\
         & w/ r.s. & 0.950 & 1.0 & -- \\
         & w/ s.e. & 0.945 & 0.967 & 1.0 \\
        \bottomrule
    \end{tabular}}
\end{table}

\subsection{Closed-World Training Data}
\label{sec:appendix-closed-world}

One might concern that the degradation under operator perturbation results from conflict with pretrained factual knowledge. To control for this, we build a 20k closed-world training data over synthetic entities and a 1k data for evaluation. Results are in Table~\ref{tab:closed_world}. In this setting, there is no pretrained factual association for the perturbed outputs to contradict. The same operator--operand asymmetry still holds.

\begin{table}[t]
    \centering
    \small
    \caption{Performance on the closed-world evaluation set for models trained on 20k closed-world data over synthetic entities with artificial relations.}
    \label{tab:closed_world}
    \begin{tabular}{lcc}
        \toprule
        \textbf{Training Data} & \textbf{LLaMA3.1-8B} & \textbf{Qwen3-8B} \\
        \midrule
        Naive & 0.457 & 0.534 \\
        Operand-Perturbed & 0.439 & 0.558 \\
        Operator-Perturbed & 0.298 & 0.392 \\
        \bottomrule
    \end{tabular}
\end{table}

\subsection{Statistical Significance}
\label{sec:appendix-significance}

We perform paired Wilcoxon signed-rank tests across the 10 evaluation benchmarks, comparing the proposed Atomic-QA against the mean performance of all baseline methods on each model family. The improvements are statistically significant in every case ($p < 0.01$). On LLaMA 3.1-8B, Atomic-QA raises the average score from 0.5716 to 0.5980 ($\Delta$=+0.0264, $W$=0.0, $p$=0.00195). On Qwen2.5-7B, the gain is +0.0313 (0.5805$\to$0.6118, $W$=0.0, $p$=0.00195). The larger Qwen2.5-14B also benefits substantially ($\Delta$=+0.0377, $W$=3.0, $p$=0.00977), as does Qwen3-8B ($\Delta$=+0.0375, $W$=1.0, $p$=0.00391). These results confirm that the advantages of Atomic-QA over existing data synthesis and selection strategies are robust and not attributable to chance variation.

\subsection{Comparison at Matched Data Scale}
\label{sec:matched-scale}
To investigate whether the advantages of our method persist under a controlled data-scale comparison, we construct \textbf{Atomic-LIFD-20K}, a more aggressively filtered subset of Atomic-QA designed to match the token budget of LIMA ($\sim$600K tokens). In Table~\ref{tab:matched_scale}, Atomic-LIFD-20K consistently outperforms LIMA across all model architectures, despite using approximately 15\% fewer tokens.

\begin{table}[t]
    \centering
    \small
    \caption{Performance comparison between Atomic-LIFD-20K and LIMA at matched data scale.}
    \label{tab:matched_scale}
    \resizebox{\linewidth}{!}{
    \begin{tabular}{l r l c}
        \toprule
        \textbf{Training Data} & \textbf{Tokens} & \textbf{Model} & \textbf{Avg. Perf.} \\
        \midrule
        \multirow{4}{*}{Atomic-LIFD-20K} & \multirow{4}{*}{535{,}706} & LLaMA3.1-8B & 0.583 (\textcolor{teal}{$\uparrow$0.018}) \\
        & & Qwen3-8B & 0.627 (\textcolor{teal}{$\uparrow$0.025}) \\
        & & Qwen2.5-7B & 0.623 (\textcolor{teal}{$\uparrow$0.029}) \\
        & & Qwen2.5-14B & 0.638 (\textcolor{teal}{$\uparrow$0.036}) \\
        \midrule
        \multirow{4}{*}{LIMA} & \multirow{4}{*}{627{,}993} & LLaMA3.1-8B & 0.565 \\
        & & Qwen3-8B & 0.602 \\
        & & Qwen2.5-7B & 0.594 \\
        & & Qwen2.5-14B & 0.602 \\
        \bottomrule
    \end{tabular}}
\end{table}

\subsection{Diversity Retention under LIFD Filtering}
\label{sec:diversity-retention}
To assess how much corpus diversity is retained after LIFD-based filtering, we compute diversity metrics for Atomic-QA, Atomic-LIFD-20K, and Atomic-LIFD-100K. As shown in Table~\ref{tab:diversity_retention}, the filtered subsets preserve comparable or even higher diversity than the full dataset, demonstrating that LIFD effectively retains diverse samples while reducing data volume.

\begin{table}[t]
    \centering
    \small
    \caption{Diversity metrics of Atomic-QA and its LIFD-filtered subsets.}
    \label{tab:diversity_retention}
    \resizebox{\linewidth}{!}{
    \begin{tabular}{l c c c c c c}
        \toprule
        \textbf{Dataset} & \textbf{TTR} & \textbf{MTLD} & \textbf{MSTTR} & \textbf{MATTR} & \textbf{Vendi} & \textbf{LogDet} \\
        \midrule
        Atomic-QA & 0.2688 & 226.97 & 0.8689 & 0.8695 & 4.2840 & $-$1187.36 \\
        Atomic-LIFD-20K & 0.2671 & 263.11 & 0.8817 & 0.8817 & 4.2963 & $-$1169.71 \\
        Atomic-LIFD-100K & 0.2616 & 253.22 & 0.8820 & 0.8806 & 4.2854 & $-$1182.44 \\
        \bottomrule
    \end{tabular}
    }
\end{table}

\subsection{Representation Stability of LIMA}
\label{sec:lima-stability}
To investigate whether the high CKA similarity between LIMA-finetuned models and their base counterparts (Figure~\ref{fig:variance_aggregate}) is an artifact of under-training (e.g., insufficient epochs or an unadjusted learning rate schedule), we extend the training of LIMA from 2 to 5 epochs and measure CKA similarity relative to the base model at each stage. As shown in Table~\ref{tab:cka_epochs}, the marginal decrease in CKA across extended epochs indicates that the models have effectively converged and are not significantly drifting from the base model's representations.

\begin{table}[t]
    \centering
    \small
    \caption{CKA similarity between LIMA-finetuned and base models across training epochs.}
    \resizebox{\linewidth}{!}{
    \label{tab:cka_epochs}
    \begin{tabular}{l c c c c}
        \toprule
        \textbf{Model} & \textbf{Epoch 2} & \textbf{Epoch 3} & \textbf{Epoch 4} & \textbf{Epoch 5} \\
        \midrule
        LLaMA-3.1-8B & 0.9995 & 0.9984 & 0.9961 & 0.9922 \\
        Qwen2.5-7B & 0.9996 & 0.9986 & 0.9970 & 0.9944 \\
        \bottomrule
    \end{tabular}}
\end{table}

We further evaluate the downstream performance to check for potential gains from extended training. As shown in Table~\ref{tab:lima_perf_epochs}, the average performance saturates after epoch 2, with negligible improvements in subsequent epochs, confirming that the training has converged and additional epochs do not yield meaningful representation changes.

\begin{table}[t]
    \centering
    \small
    \caption{Average performance of LIMA-finetuned models across training epochs.}
    \label{tab:lima_perf_epochs}
    \resizebox{\linewidth}{!}{
    \begin{tabular}{l c c c c}
        \toprule
        \textbf{Model} & \textbf{Epoch 2} & \textbf{Epoch 3} & \textbf{Epoch 4} & \textbf{Epoch 5} \\
        \midrule
        LLaMA-3.1-8B & 0.565 & 0.563 & 0.566 & 0.564 \\
        Qwen2.5-7B & 0.594 & 0.592 & 0.601 & 0.594 \\
        \bottomrule
    \end{tabular}}
\end{table}

\section{Beta-Distribution Based Error Assignment}
\label{sec:Beta-Distribution}

In this section, we present a Beta-Distribution based approach for assigning errors to samples while maintaining a target global error rate.We first formulate the problem, then introduce our stochastic assignment method that leverages the Beta distribution to create realistic error diversity across samples.

\subsection{Problem Formulation}
We consider a dataset of $M$ samples, where sample $i$ comprises $n_i$ atomic functions. Our objective is to introduce errors while matching a prescribed global error rate $r \in [0,1]$ at the atomic-function level. The total number of atomic functions is $N=\sum_{i=1}^{M} n_i$.

For each sample $i$, we pre-generate $n_i+1$ candidate versions indexed by $k \in \{0,1,\ldots,n_i\}$, where the $k$-th version contains exactly $k$ corrupted functions. Each version is constructed by introducing one additional corruption relative to the $(k-1)$-th version. Exactly one version must be selected for each sample to satisfy the following condition: \[
\sum_{i=1}^{M} e_i = \text{round}(r \cdot N),
\]where $e_i$ is the number of errors in the selected version of sample $i$, and $\text{round}(\cdot)$ rounds to the nearest integer.

\subsection{Beta Distribution Based Assignment}

\paragraph{Motivation.} A naive uniform approach assigns errors deterministically by setting $e_i = \lfloor r \cdot n_i \rfloor$ for each sample. While this guarantees the global error rate, it produces homogeneous error distributions where every sample exhibits approximately the same error rate. Such uniformity fails to capture the natural heterogeneity observed in real-world data, where some samples are nearly perfect while others contain multiple errors. To address this limitation, we introduce controlled stochasticity through the Beta distribution.

\paragraph{Beta Distribution Properties.}
The Beta distribution is a continuous probability distribution defined on the interval $[0,1]$, making it particularly well suited for modeling proportions and probabilities. Its flexible shape is controlled by two parameters $\alpha$ and $\beta$. Under the parameterization $\alpha = r\kappa$ and $\beta = (1-r)\kappa$, the distribution has mean
\[
\mathbb{E} = \frac{\alpha}{\alpha + \beta} = r
\]
and the concentration parameter $\kappa$ controls the shape of the distribution.

\paragraph{Stochastic Sampling Phase.}
We model each sample’s error probability using a Beta distribution parameterized by the target error rate $r$ and concentration parameter $\kappa$. For each sample $i$, we independently draw an error probability
\[
q_i \sim \mathrm{Beta}(r\kappa, (1-r)\kappa).
\]
The parameterization ensures that the expected per-sample error probability satisfies $\mathbb{E}[q_i]=r$, so that the expected global error rate converges to the target value $r$. This stochastic sampling introduces natural heterogeneity across samples, reflecting realistic scenarios in which data quality varies at the instance level. In our experiments, we set $\kappa = 10$, which provides moderate variance and produces the relationship between atomic-level error rates and the proportion of corrupted samples shown in Figure~\ref{fig:atomic_error_rate_math}.

\paragraph{Normalization, Adjustment, and Quantization.} 
To ensure that the expected total number of errors matches the target, we first compute the aggregate statistic\[
S = \sum_{j=1}^{M} n_j q_j,
\]and derive the normalization constant as\[
c = \frac{r \cdot N}{S}.
\]The adjusted error probability for sample $i$ is then given by
:
\[
p_i = \text{clip}(c \cdot q_i, 0, 1).
\]
This correction enforces $\sum_{i=1}^M n_i p_i \approx r \cdot N$ while maintaining the sample-level heterogeneity introduced by the Beta sampling. The clipping operation ensures all probabilities remain within $[0,1]$.

Since error counts must be integers, we set the error count for each sample to
\[
e_i = \operatorname{round}(n_i \cdot p_i),
\]
which rounds $n_i \cdot p_i$ to the nearest integer, and compute the residual $\Delta = \operatorname{round}(r \cdot N) - \sum_{i=1}^{M} e_i$. To correct this discretization error, we iteratively adjust error assignments to ensure that the final assignment satisfies
\[
\sum_{i=1}^{M} e_i = \operatorname{round}(r \cdot N).
\]
The complete procedure is detailed in Algorithm~\ref{alg:beta_assignment}, where $\operatorname{sign}(x)$ denotes the sign function that returns $+1$ if $x > 0$, $-1$ if $x < 0$, and $0$ if $x = 0$.

\begin{algorithm}[t]
    \caption{Beta Distribution Error Assignment}
    \label{alg:beta_assignment}
    \SetKwInOut{Input}{Input}
    \SetKwInOut{Output}{Output}
    
    \Input{atomic function counts $\mathbf{n} = \{n_1, \ldots, n_M\} \in \mathbb{N}^M$}
    \Input{target error rate $r \in [0,1]$, concentration $\kappa$ (default $10$)}
    \Output{error counts $\mathbf{e} \in \mathbb{N}^M$ with $\sum_i e_i = T$}
    
    $N \leftarrow \sum_{i} n_i$;\quad $T \leftarrow \text{round}(r N)$
    
    \tcp{Sample per-sample error probabilities}
    \ForAll{$i \in \{1,\ldots,M\}$}{
        $q_i \sim \text{Beta}(r\kappa,\,(1-r)\kappa)$
    }
    
    \tcp{Normalize so expected total matches $T$}
    $c \leftarrow r N \,\big/\, \sum_{j} n_j q_j$
    
    \ForAll{$i \in \{1,\ldots,M\}$}{
        $p_i \leftarrow \text{clip}(c\, q_i,\, 0,\, 1)$
        
        $e_i \leftarrow \text{round}(n_i\, p_i)$
    }
    
    \tcp{Correct rounding residual}
    $\Delta \leftarrow T - \sum_{i} e_i$;\quad $s \leftarrow \operatorname{sign}(\Delta)$
    
    $\sigma \leftarrow \text{argsort}(-s \cdot \mathbf{n})$ \tcp*{large-$n_i$ first if $\Delta>0$}
    
    \For{$i \in \sigma$}{
        \lIf{$\Delta = 0$}{\textbf{break}}
        \If{$0 \le e_i + s \le n_i$}{
            $e_i \leftarrow e_i + s$
            
            $\Delta \leftarrow \Delta - s$
        }
    }
    
    \Return $\mathbf{e}$
\end{algorithm}

\begin{figure}[!ht]
    \centering
    \begin{subfigure}[b]{\linewidth}
        \centering
        \includegraphics[width=0.9\linewidth, keepaspectratio]{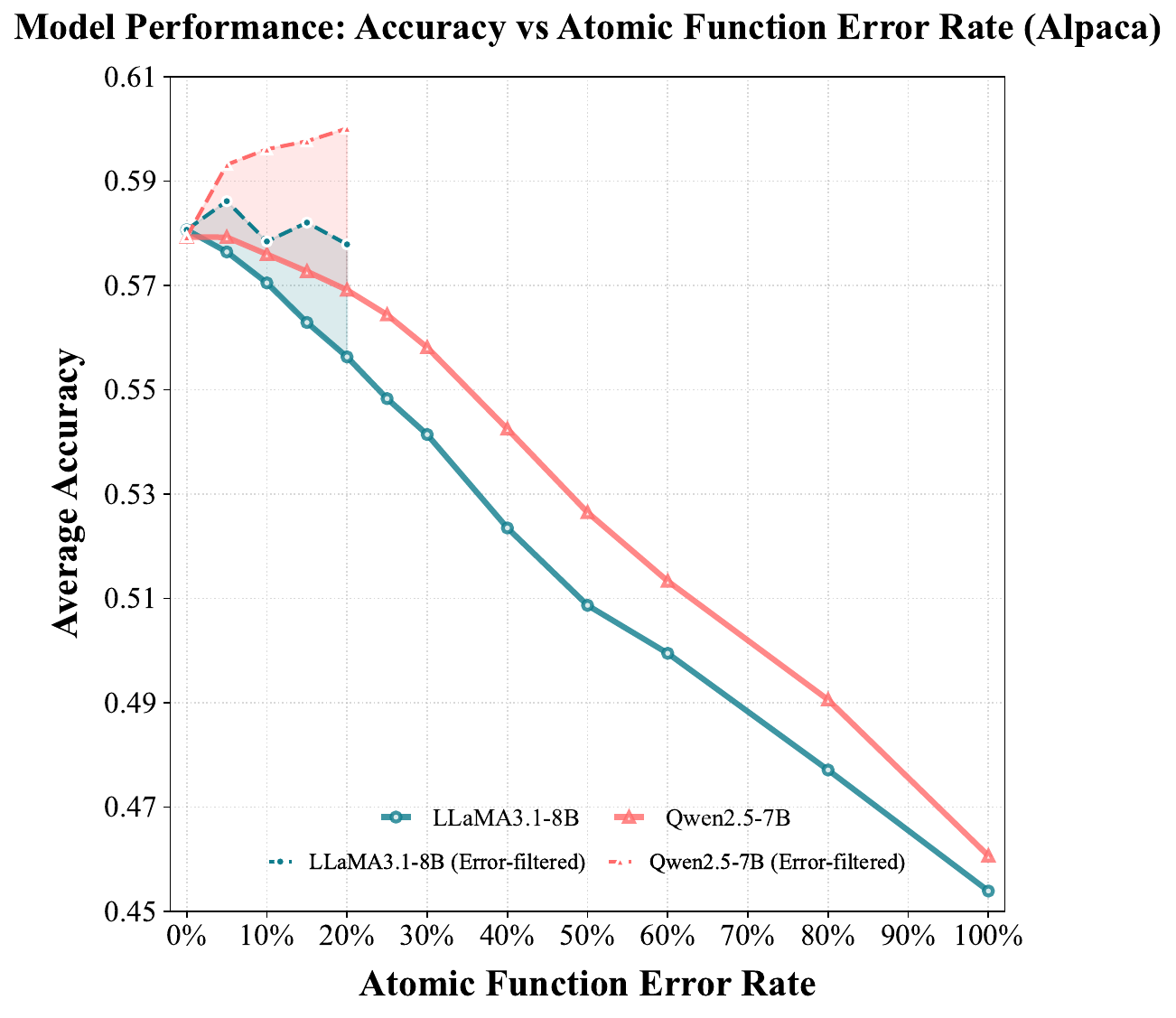}
        \caption{Alpaca accuracy degradation under the perspective of erroneous atomic functions ratio.}
        \label{fig:alpaca_atomic}
    \end{subfigure}
    
    \vspace{10pt}
    
    \begin{subfigure}[b]{\linewidth}
        \centering
        \includegraphics[width=0.9\linewidth, keepaspectratio]{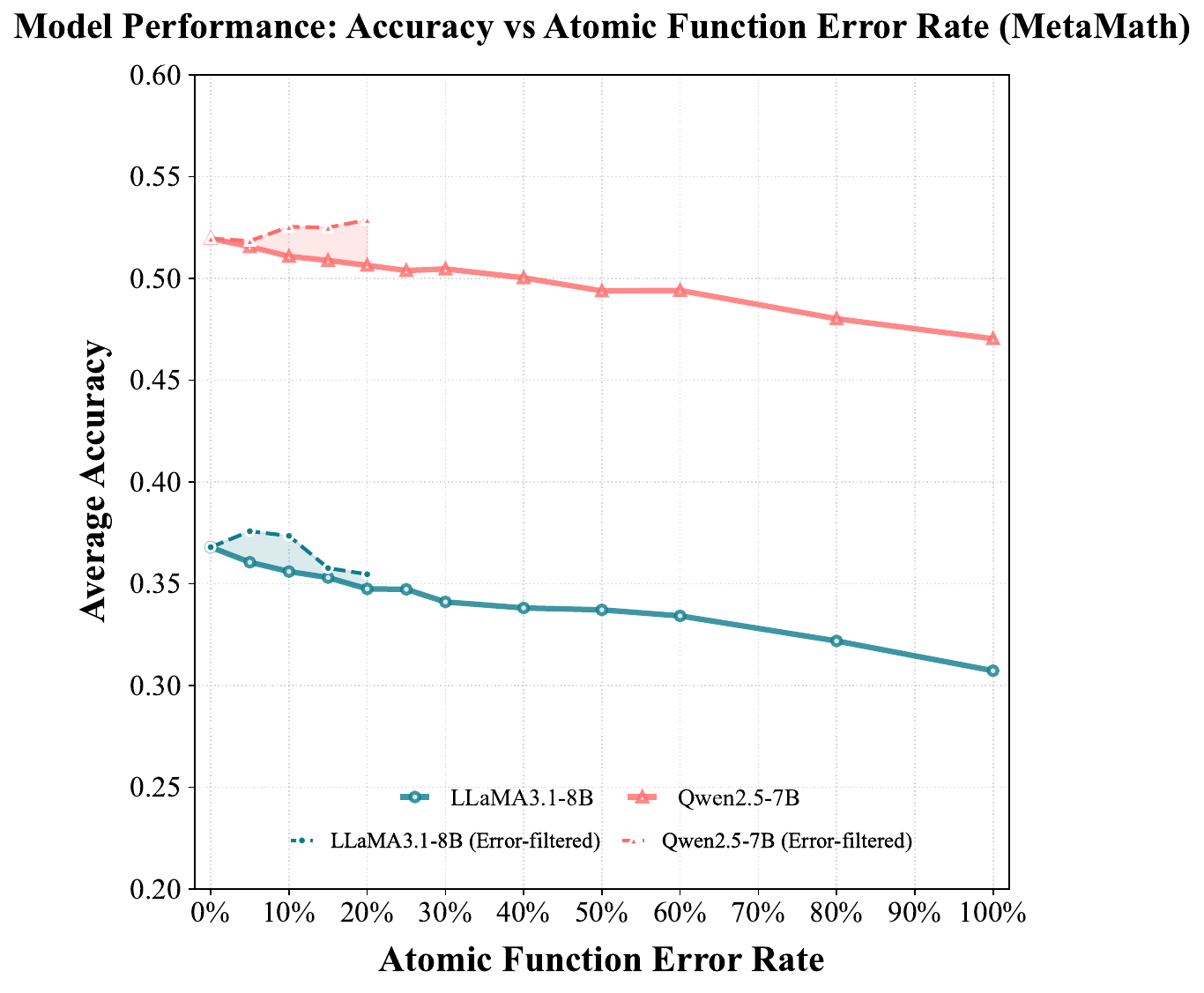}
        \caption{MetaMath accuracy degradation under the perspective of erroneous atomic functions ratio.}
        \label{fig:math_atomic}
    \end{subfigure}
    
    \caption{Accuracy degradation under increasing atomic function error rates for (a) LLaMA and (b) Qwen. The x-axis denotes the proportion of erroneous atomic functions in the training set, while the y-axis represents the performance of the trained model.}
    \label{fig:atomic_error_rate_alpaca_math}
\end{figure}

\begin{figure}[!ht]
    \centering
    \includegraphics[width=0.9\linewidth]{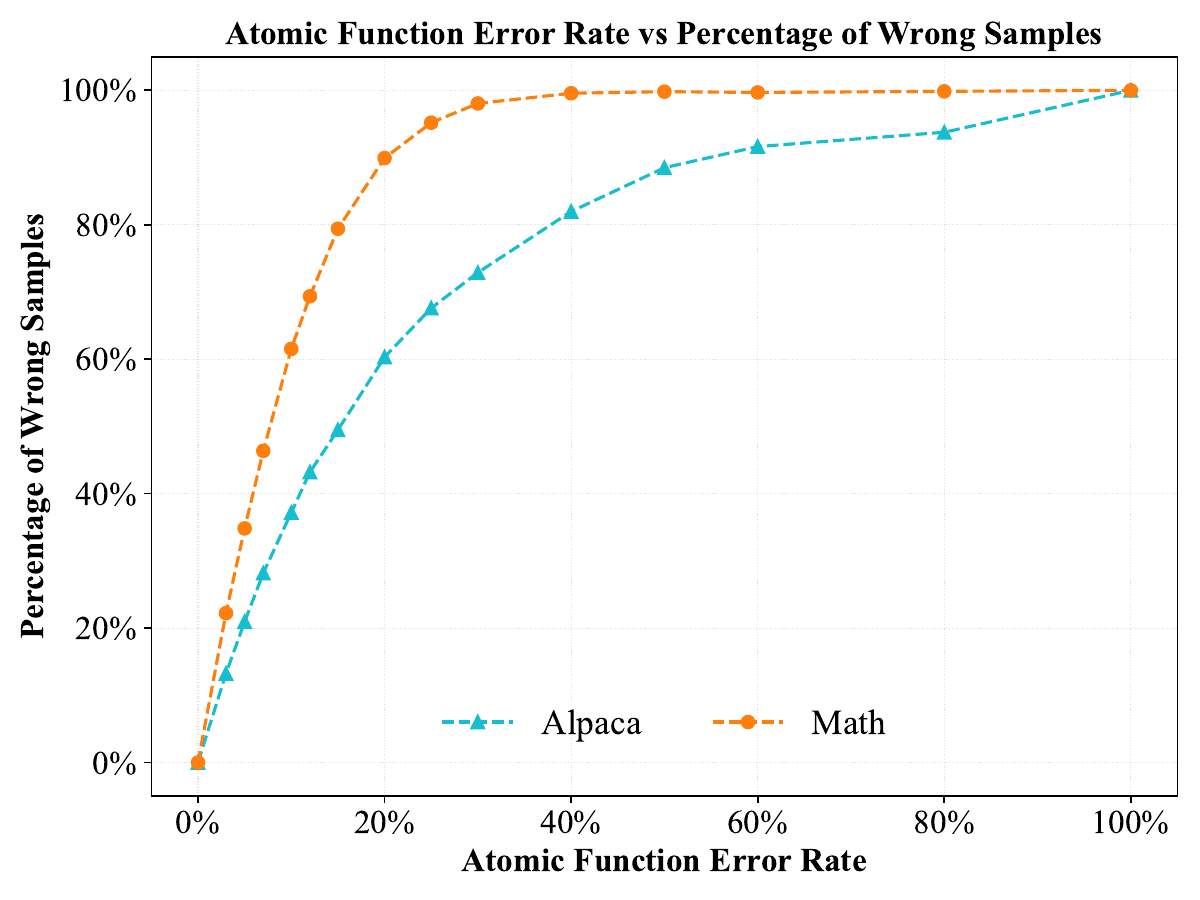}
    \caption{
        Relationship between the atomic function error rate and the resulting percentage of wrong samples. 
        }
    \label{fig:atomic_error_rate_math}
\end{figure}

\section{Prompts}

\begin{figure*}[!ht]
\begin{promptbox}[Atomic-Function-Extraction]{Atomic Function Extraction}
You are the ENCODER agent. Your job is to extract atomic functions from the Sample Output and represent each input as (operand, operator, output) triples.

{atomic function definition}

Return **exactly one JSON array** of atomic functions. If no atomic functions can be extracted, return [].

## Atomic Function Definition
{atomic function definition}

## Rules
1) Minimality: one predicate per triple. Do not merge multiple assertions. If the current sample is already the smallest unit satisfying the definition, no splitting is required.
2) Order: emit triples in the order their evidence appears in the Sample Output.
3) Numbers/dates: keep literal values and units as written (e.g., "18", "2019", "3.14", "USD").
4) Determinism: no randomness; if uncertain, omit the item.
5) Strict output: JSON array only, no comments, no trailing commas, no extra keys.

## Sample Input
This section provides the original input of the data sample. Such as a question, instruction, or code comment.
{sample input}

## Sample Output
This section provides the original answer of the data sample. It may contain multiple sentences, equations, or code snippets.
{sample output}

## Output Format
Your answer should wrapped in a proper Json code block using triple backticks like this:
```json
{example}
```
\end{promptbox}
\end{figure*}

\begin{figure*}[!ht]
\begin{promptbox}[Initial-Tree-Generation]{Initial Tree Generation}
The following are atomic functions, followed by their definitions. 

Input atomic functions:
{atomic functions}
The definitions of the atomic functions:
{definitions of atomic functions}

Requirements:
1. Format the output as a JSON tree structure.
2. Consider both coarse-grained and fine-grained classification of atomic functions' categories, building parent-child relationships.
3. Each node is a dict with category names as keys.
4. The root node should be {root name}.
5. Leaf nodes should contain empty dicts.
6. Wrap the JSON with <begin> and <end>.

Think step by step:
1. Identify the categories of the atomic functions
2. Organize the categories (not atomic functions) into a hierarchical tree structure.

Output format:
<think>
[your reasoning process]
</think>
<begin>
{example}
<end>
\end{promptbox}
\end{figure*}

\begin{figure*}[!ht]
\begin{promptbox}[Tree-Duplication]{Tree Duplication}
Please perform strict deduplication on the following subcategories under the parent category "{parent category}". 

Requirements:
1. Remove any semantic duplicates (even if names are different)
2. Ensure all items strictly belong to "{parent category}", remove any unrelated items
3. Eliminate any hierarchical relationships between subcategories, remove subcategories that are included in other subcategories
4. Keep the most appropriate names
5. Output the cleaned list

Current subcategories:
{subcategories}

Output format:
```json
{{"unique_categories": ["example1", "example2"]}}
```
\end{promptbox}
\end{figure*}

\begin{figure*}[!ht]
\begin{promptbox}[Tree-Expansion]{Tree Expansion}
As a taxonomy expert, suggest 3-5 possible {width/depth} expansion candidates for category: [{category}]

Related Nodes:
- Parent: {parent}
- Siblings: {siblings}

Definitions:
- Current Node: {definition}
- Parent: {parent_definition}
- Siblings: {siblings_definitions}

Guidelines:
1. Ensure logical consistency with parent/sibling definitions
2. Avoid overlaps with existing nodes
3. Use clear and specific naming

Output JSON format:
```json
{{"new_nodes": ["name1", "name2"]}}
```
\end{promptbox}
\end{figure*}

\begin{figure*}[!ht]
\begin{promptbox}[Operand-Generation]{Operand Generation}
You are given the following information:
    - Category
    - Operator
    - Existing Operand

Your task is to generate new operands that are valid inputs for the given operator under the specified category.
Your generation should consider the following aspects:
1. The generated operands should be specific and fall under the given category.
2. The generated operands should be valid inputs for the given operator.
3. The generated operands should not be similar to the existing operands.
4. Each generated operand should have a brief description that help understand the operands.

## Category ##
This section contains the category that you need to generate operands for.
- Category Name: {category}
- Category Definition: {definition}

## Operator ##
- Operator: {operator}
- Operator Definition: {operator_definition}

## Existing operands ##
This section contains the existing operands that are already classified under the category. 
You should avoid generating operands that are similar to the existing ones.
- Existing operands:
{existing operands}

## Output format ##
You should output the generated operands in the JSON format, wrapped in a proper Json code block using triple backticks like this:
```json
{{
    "operand_name1": "Description of operand 1",
    "operand_name2": "Description of operands 2",
    ...
}}
```
\end{promptbox}
\end{figure*}

\begin{figure*}[!ht]
\begin{promptbox}[Verification]{Verification}
You are given the following information:
    - Function: {function}
    - Operand: {operand}
    - Output: {output}

Your task is to verify whether the function is logically consistent, that is, whether **Function(Operand)** equals the **Output**.

## Output format ##
You should output the judgement in the JSON format, wrapped in a proper Json code block using triple backticks like this:
```json
{{
    "judgement": "pass"/"fail",
    "reason": "reason for your judgement"
}}
```
\end{promptbox}
\end{figure*}

\begin{figure*}[!ht]
\begin{promptbox}[Encoder]{Encoder}
You are the ENCODER agent. Your job is to extract atomic functions from the Sample Output and represent each atomic function as a (subject, relation, object) triple.

## Categories
This chapter enumerates the possible candidate categories for decomposing atomic functions. You need to determine whether the **## Payload** can be further decomposed within some category.
{categories}

## Sample Input
This section provides the original input of the data sample. Such as a question, instruction, or code comment.
{sample input}

## Sample Output
This section provides the original answer of the data sample. It may contain multiple sentences, equations, or code snippets.
{sample output}

## Payload
This section provides the original answer of the data sample. It may contain multiple sentences, equations, or code snippets.
{sample output}

## Output Format
Your answer should wrapped in a proper Json code block using triple backticks like this:
```json
{example}
```
\end{promptbox}
\end{figure*}

\begin{figure*}[!ht]
\begin{promptbox}[Perturber]{Perturber}
You are the PERTURBER agent. Your job is to apply controlled edits to specific atomic functions in the Sample Output, guided by the provided (subject, relation, object) triple and payload location.

A *controlled edit* is a minimal, deterministic change that makes the function incorrect while preserving overall coherence and fluency.

Return **exactly one JSON object** describing the edit.

Use the Sample Input only to preserve coherence (e.g., pronouns); all edits **must** be grounded in the Sample Output and the given function.


## Sample Input
This section provides the original input of the data sample (e.g., question/instruction/code comment).
{input_sample}

## Sample Output
This section provides the original answer of the data sample (sentences, equations, or code).
{output_sample}

## Atomic function to be Perturbed (from ENCODER)
This section provides the function to be perturbed, along with its unique fid and the corresponding payload location in the Sample Output.
function: {triple}
Payload: {payload}

## Output Format
Your answer should wrapped in a proper Json code block using triple backticks like this:
## Output Format
Your answer should wrapped in a proper Json code block using triple backticks like this:
```json
{example}
```
\end{promptbox}
\end{figure*}

\begin{figure*}[!ht]
\begin{promptbox}[Decoder]{Decoder}
You are the DECODER agent. Your job is to reconstruct an output sample from the perturbed atomic function that is provided.

You shoudld use the SEARCH-REPLACE strategy to ensure the perturbed function is correctly integrated into the Sample Output while maintaining overall coherence and fluency.

You are allowed to use the following tools, and you must use one tool at a time:
- SEARCH-REPLACE: Locate the content in the Sample Output and replace it with the given content.
For each change, follow this format exactly:
```[TOOL_CALL]
>>> SEARCH
<original snippet copied from the Sample Output>
======
>>> REPLACE
<modified snippet you want to replace the original snippet with>
======
```

- TERMINATE: Indicate the end of the editing process and output the final revised Sample Output.
To finish the editing process, follow this format exactly:
```[TOOL_CALL]
Terminate(reason="The reason for termination, e.g., no more changes needed")
```

## Sample Input
This section provides the original input of the data sample (e.g., question/instruction/code comment).
{input_sample}

## Sample Output
This section provides the original answer of the data sample (sentences, equations, or code).
{output_sample}

## Perturbed Atomic function (from PERTURBER)
This section provides the perturbed function along with the type of edit applied. You should use change the corresponding content that matches the Original function in the Sample Output to the Perturbed function.
Original function: {pre_triple}
Perturbed function: {post_triple}

## Output Format
IMPORTANT: Only use one tool call per response. Do not use multiple tool calls in a single response or code block.
IMPORTANT: You ONLY need to edit the sepcific function that is perturbed. Do NOT make any other changes to the Sample Output. Do NOT consider the effect of edited function on other parts of the Sample Output.
Your tool calls should be wrapped in a proper [TOOL_CALL] code block using triple backticks like this:
```[TOOL_CALL]
<your tool call here>
```
\end{promptbox}
\end{figure*}
This section enumerates all prompts utilized in our framework. Prompts \ref{Atomic-Function-Extraction}, \ref{Initial-Tree-Generation}, \ref{Tree-Duplication}, and \ref{Tree-Expansion} detail the construction process of the Atomic Tree. Subsequently, Prompts \ref{Operand-Generation} and \ref{Verification} outline the creation of Atomic-QA, while Prompts \ref{Encoder}, \ref{Perturber}, and \ref{Decoder} correspond to Atomic-X pipeline in $\S$\ref{sec:Atomic-X}.

\section{Examples of Atomic Function}
\label{sec:appendix-atomicexample}
\begin{figure*}[!ht]
    \centering
    \includegraphics[
        width=1\textwidth,
        trim=12 132 12 12,
        clip
    ]{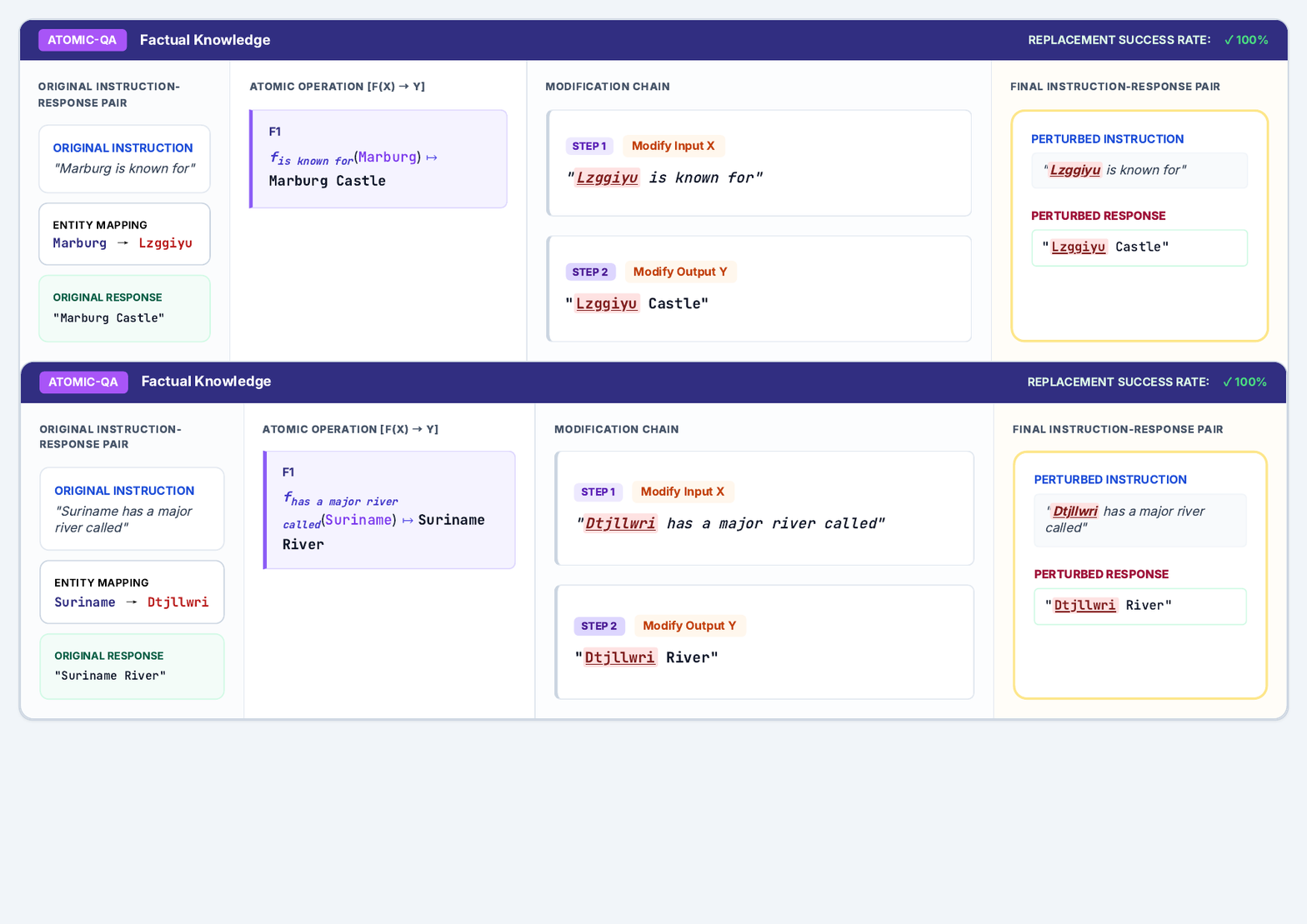}
    \caption{The example of Atomic-QA.}
    \label{fig:Atomic-QA_example}
\end{figure*}

\begin{figure*}[!ht]
    \centering
    \includegraphics[
        width=1\textwidth,
        trim=12 35 12 12,
        clip
    ]
    {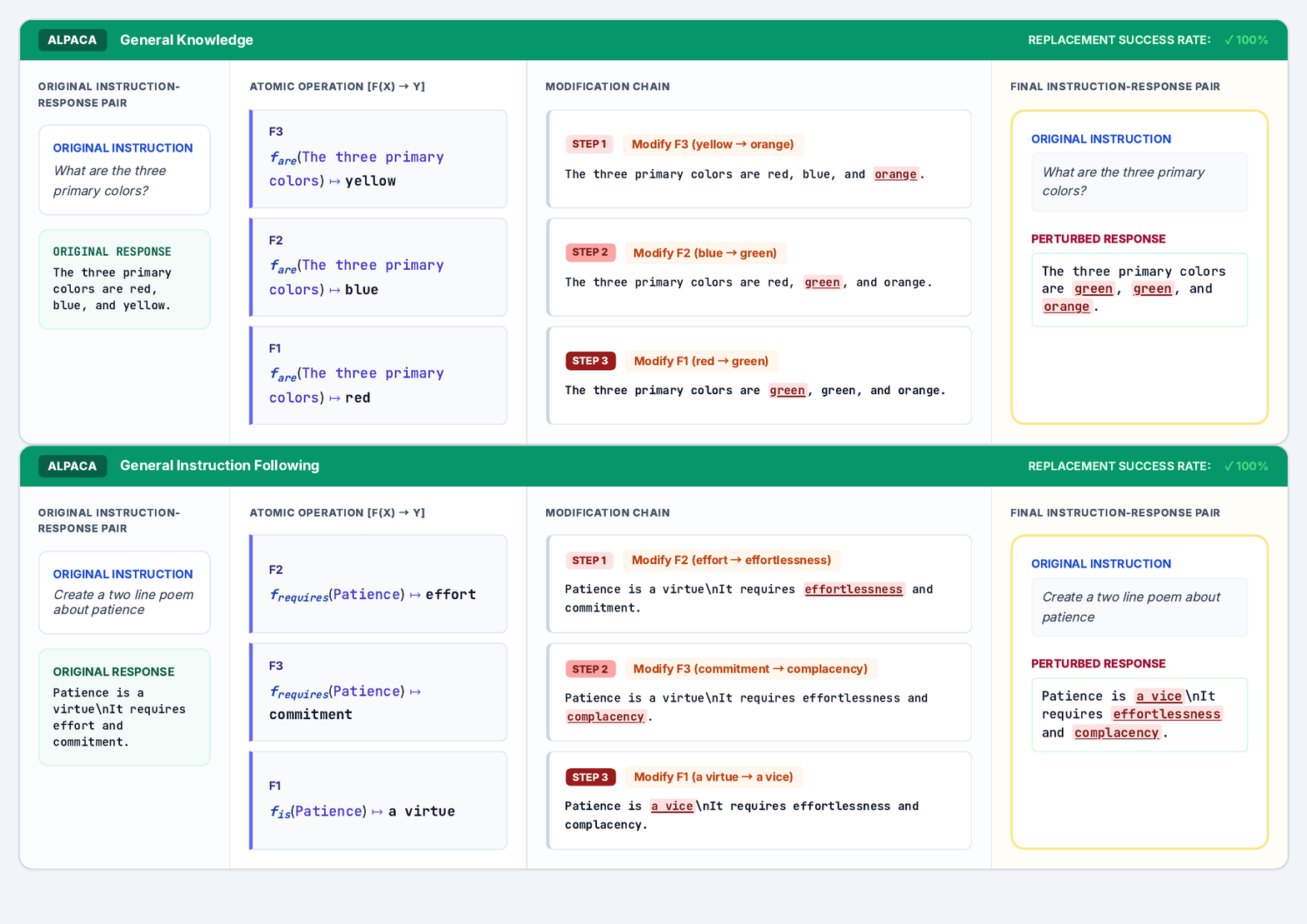}
    \caption{The example of Alpaca.}
    \label{fig:alpaca_example}
\end{figure*}

\begin{figure*}[!ht]
    \centering
    \includegraphics[
        width=1\textwidth,
        trim=9 21 9 9,
        clip
    ]
    {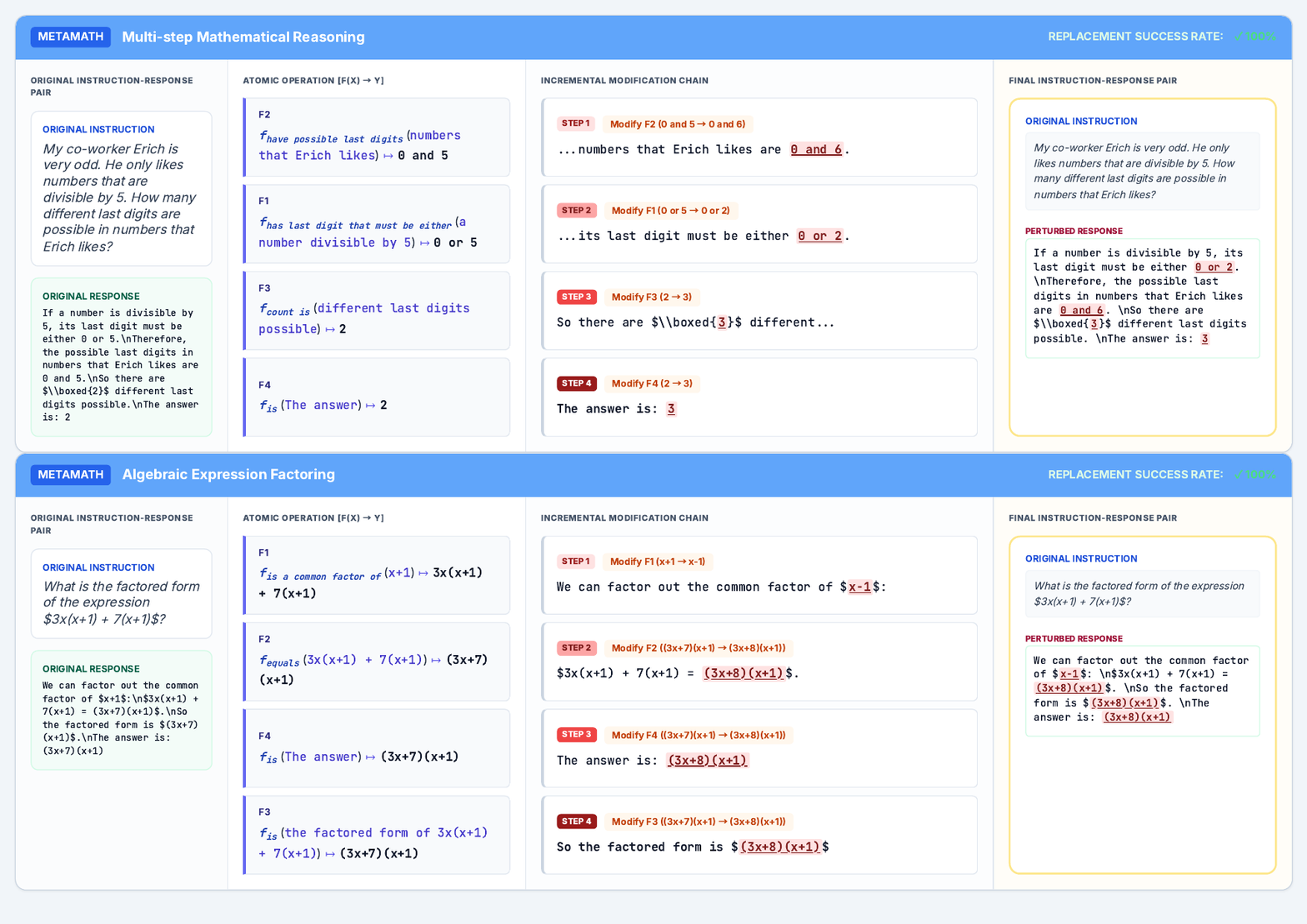}
    \caption{The example of MetaMath.}
    \label{fig:math_example}
\end{figure*}

In this section, we present examples of atomic functions. Figures \ref{fig:Atomic-QA_example}, \ref{fig:alpaca_example}, and \ref{fig:math_example} illustrate instances from the Atomic-QA, Alpaca, and Metamath datasets, respectively. In each figure, the leftmost panel displays the original input prompt and its corresponding response. The central columns list the extracted atomic functions followed by the perturbed atomic functions. Finally, the rightmost panel shows the output response after perturbation.

\end{document}